\documentclass[12pt,letterpaper]{article}%
\usepackage{setspace}
\usepackage{amssymb}
\usepackage{amsmath}
\usepackage{amsfonts}
\usepackage{graphicx}
\usepackage{color}%
\newtheorem{theorem}{Theorem}

\newtheorem{corollary}[theorem]{Corollary}

\newtheorem{lemma}[theorem]{Lemma}

\newenvironment{proof}[1][Proof]{\noindent\textbf{#1.} }{\ \rule{0.5em}{0.5em}}
\ifx\pdfoutput\relax\let\pdfoutput=\undefined\fi
\newcount\msipdfoutput
\ifx\pdfoutput\undefined\else
\ifcase\pdfoutput\else
\ifx\paperwidth\undefined\else
\ifdim\paperheight=0pt\relax\else\pdfpageheight\paperheight\fi
\ifdim\paperwidth=0pt\relax\else\pdfpagewidth\paperwidth\fi
\fi\fi\fi
\begin{document}

\title{{\LARGE Discrete Bayesian Networks: The Exact Posterior Marginal Distributions
}}
\author{Do Le (Paul) Minh\\Department of ISDS, California State University, Fullerton\\CA\ 92831, USA\\dminh@fullerton.edu}
\date{\today}
\maketitle

Abstract: In a Bayesian network, we wish to evaluate the marginal probability
of a query variable,\ which may be conditioned on the observed values of some
evidence variables. Here we first present our \textquotedblleft border
algorithm,\textquotedblright\ which converts a BN into a directed chain. For
the polytrees, we then present in details, with some modifications and within
the border algorithm framework, the \textquotedblleft revised polytree
algorithm\textquotedblright\ by Peot \& Shachter (1991). Finally, we present
our \textquotedblleft parentless polytree method,\textquotedblright\ which,
coupled with the border algorithm, converts any Bayesian network into a
polytree, rendering the complexity of our inferences independent of the size
of network, and linear with the number of its evidence and query variables.
All quantities in this paper have probabilistic interpretations.\bigskip

\textbf{Keywords:} Bayesian networks; Exact inference; Border algorithm;
Revised polytree algorithm; Parentless polytree method\bigskip

\section{The Bayesian Networks\ (BNs)}

\begin{figure}
[ptb]
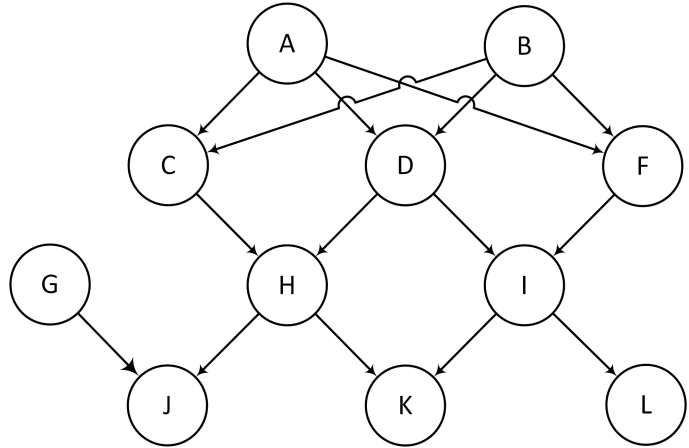

\begin{center}
\includegraphics[
natheight=2.323700in,
natwidth=3.544100in,
height=2.3237in,
width=3.5441in
] 1%
\caption{The Bayesian Network $\mathbb{A}$}%
\label{F1}%
\end{center}
\end{figure}

Consider a directed graph $\mathbb{G}$ defined over a set of $\ell$\ nodes
$\mathcal{V}=\left\{  V_{1},V_{2},...,V_{\ell}\right\}  $, in which each node
represents a variable. (We denote both a variable and its corresponding node
by the same notation, and use the two terms interchangeably.) The pairs of
nodes $\left(  V_{i},V_{j}\right)  $ may be connected by either the directed
edge $V_{i}\rightarrow V_{j}$ or $V_{j}\rightarrow V_{i}$, but not both. It is
not necessary that all pairs be connected in this manner. In this paper, we
will first use the graph in Figure \ref{F1} as an example.%

For node $V\in\mathcal{V}$, we call

\begin{enumerate}
\item the nodes sending the directed edges to $V$ the \textquotedblleft
parents\textquotedblright\ of $V$. We denote the set of the parents\ of $V$ by
$\mathcal{H}_{V}.$ In Figure \ref{F1}, $\mathcal{H}_{H}=\left\{  C,D\right\}
$. A node is said to be a \textquotedblleft root\textquotedblright\ if it has
no parents. (For example, nodes $A$, $B$ and $G$.)

\item the nodes receiving the directed edges from $V$ the \textquotedblleft
children\textquotedblright\ of $V$. We denote the set of the children of $V$
by $\mathcal{L}_{V}$. In Figure \ref{F1}, $\mathcal{L}_{D}=\left\{
H,I\right\}  $. A node is said to be a \textquotedblleft
leaf\textquotedblright\ if it has no children. (For example, nodes $J$, $K$
and $L$.) We also call the parents and children of $V$ its \textquotedblleft
neighbors.\textquotedblright

\item the parents of the children of $V$, except $V$, the \textquotedblleft
co-parents\textquotedblright\ of $V$. We denote the set of the co-parents\ of
$V$ by $\mathcal{K}_{V}=\left\{  \cup_{\eta\in\mathcal{L}_{V}}%
\text{$\mathcal{H}$}_{\eta}\right\}  \backslash V$. (We denote by
$\mathcal{X}\backslash\mathcal{Y}$ the set $\left\{  X:X\in\mathcal{X}%
,X\notin\mathcal{Y}\right\}  $. $\mathcal{X}\backslash\mathcal{Y}=\varnothing$
iff $\mathcal{X}\subseteq\mathcal{Y}$.) In our example, $\mathcal{K}%
_{D}=\left\{  C,F\right\}  $.
\end{enumerate}

The set of edges connecting nodes $V_{i}$ and $V_{j}$ either directly or via
other nodes $V_{k}$, ..., $V_{m}$ in the form of $V_{i}\rightarrow
V_{k}\rightarrow...\rightarrow V_{m}\rightarrow V_{j}$ is called a
\textquotedblleft\ directed path\textquotedblright\ from $V_{i}$ to $V_{j}$.
We restrict ourselves to the \textquotedblleft directed acyclic
graph\textquotedblright\ (DAG) in which there is no directed path that starts
and ends at the same node. If there is a directed path\ from $V_{i}$ to
$V_{j}$, we say $V_{i}$ is an \textquotedblleft ancestor\textquotedblright\ of
$V_{j}$ and $V_{j}$ a \textquotedblleft descendant\textquotedblright\ of
$V_{i}$. Let $\mathcal{N}_{V}$ and $\mathcal{M}_{V}$ be the set of all
ancestors\ and descendants of $V$, respectively. In Figure \ref{F1} ,
$\mathcal{N}_{I}=\left\{  A,B,D,F\right\}  $, $\mathcal{M}_{C}=\left\{
H,J,K\right\}  $.

The \textquotedblleft Markovian assumption\textquotedblright\ of a DAG is that
every variable is conditionally independent of its non-descendants given its
parents. Attached to each node $V\in\mathcal{V}$ is a conditional probability
distribution $\Pr\left\{  V|\mathcal{H}_{V}\right\}  $. If a node has no
parent, its distribution is unconditional. We assume in this paper that all
$V\in\mathcal{V}$ are discrete, and all conditional probability distributions
are in the form of the conditional probability tables (CPTs), taking strictly
positive values. We assume that the \textquotedblleft size\textquotedblright%
\ of $\Pr\left\{  V|\mathcal{H}_{V}\right\}  $ (that is, the number of
possible values of $V$ and $\mathcal{H}_{V}$)\ is finite for all
$V\in\mathcal{V}$.

A \textquotedblleft Bayesian network\textquotedblright\ (BN) is a pair
$\left(  \mathbb{G},\Theta\right)  $, where $\mathbb{G}$ is a DAG over a set
of variables $\mathcal{V}=\left\{  V_{1},V_{2},...,V_{\ell}\right\}  $ (called
the \textquotedblleft network structure\textquotedblright)\ and $\Theta$ a set
of all CPTs (called the \textquotedblleft network
parametrization\textquotedblright). We will refer to the DAG in Figure
\ref{F1} and its parametrization the Bayesian network $\mathbb{A}$, or the BN
$\mathbb{A}$.

It has been shown that the dependence constraints imposed by $\mathbb{G}$ and
the numeric constraints imposed by $\Theta$ result in the unique joint
probability distribution,%
\begin{equation}
\Pr\left\{  \mathcal{V}\right\}  =\Pr\left\{  V_{1},V_{2},...,V_{\ell
}\right\}  =\prod_{V\in\mathcal{V}}\Pr\left\{  V|\text{$\mathcal{H}$}%
_{V}\right\}  . \label{1}%
\end{equation}
This equation is known as the \textquotedblleft chain rule for Bayesian
networks\textquotedblright\ (Pearl, 1987, Equation 3). In our example,
\begin{align*}
&  \Pr\left\{  A=a,B=b,C=c,...,L=\ell\right\} \\
&  =\Pr\left\{  A=a\right\}  \Pr\left\{  B=b\right\}  \Pr\left\{
C=c|A=a,B=b\right\}  ...\Pr\left\{  L=\ell|I=i\right\}  .
\end{align*}

\subsection{The Marginal Distribution}

We wish to evaluate the marginal probability $\Pr\left\{  Q\right\}  $, in
which $Q\in\mathcal{V}$ is known as a \textquotedblleft query
variable.\textquotedblright\ This probability may be conditioned on the fact
that some other variables in $\mathcal{V}$ are observed to take certain values.

Suppose $f$ is a function defined over a set of variables $\mathcal{X}%
\subseteq\mathcal{V}$. We say the \textquotedblleft scope\textquotedblright%
\ of $f$ is $\mathcal{X}$. We list out the scope if necessary, such as
$f\left(  \mathcal{X}\right)  $; if not, we simply write $f\left(
\cdot\right)  $.

In this paper, suppose $\mathcal{X}=\left\{  \mathcal{Y},\mathcal{Z}\right\}
\subseteq\mathcal{V}$ where $\mathcal{Y}\cap\mathcal{Z}=\varnothing$ and
$\mathcal{Y=}\left\{  Y_{1},...,Y_{n}\right\}  $. We express $\Pr\left\{
\mathcal{X}\right\}  $ as $\Pr\left\{  \mathcal{Y},\mathcal{Z}\right\}  $.
Given $\Pr\left\{  \mathcal{Y},\mathcal{Z}\right\}  $, \textquotedblleft
summing out\textquotedblright\ (or \textquotedblleft
eliminating\textquotedblright) $\mathcal{Y}$ from\ $\Pr\left\{  \mathcal{Y}%
,\mathcal{Z}\right\}  $ means obtaining $\Pr\left\{  \mathcal{Z}\right\}  $ as
follows: For every fixed $\mathcal{Z}=z$,%
\begin{align*}
&  \sum_{\mathcal{Y}}\Pr\left\{  z,\mathcal{Y}\right\} \\
&  =\sum_{Y_{1}}...\sum_{Y_{n-1}}\left(  \sum_{Y_{n}}\Pr\left\{  z,Y_{1}%
=y_{1},...,Y_{n-1}=y_{n-1},Y_{n}=y_{n}\right\}  \right) \\
&  =\sum_{Y_{1}}...\left(  \sum_{Y_{n-1}}\Pr\left\{  z,Y_{1}=y_{1}%
,...,Y_{n-1}=y_{n-1}\right\}  \right)  =\Pr\left\{  z\right\}  .
\end{align*}
We write,
\begin{equation}
\sum_{\mathcal{Y}}\Pr\left\{  \mathcal{Z},\mathcal{Y}\right\}  =\Pr\left\{
\mathcal{Z}\right\}  . \label{10}%
\end{equation}

One way to evaluate the marginal probability $\Pr\left\{  V_{j}\right\}  $ is
to use Equation (\ref{1}) to calculate the joint probability $\Pr\left\{
V_{1},...,V_{\ell}\right\}  $, then sum out\ all variables in $\left\{
V_{1},...,V_{j-1},V_{j+1},...,V_{\ell}\right\}  $. This brute-force method is
known to be NP-hard; that is, there is often an exponential relationship
between the number of variables $\ell$ and the complexity of computations
(Cooper, 1990). Thus it may be infeasible for large networks.

There have been many attempts in the literature to find the most efficient
methods to calculate $\Pr\left\{  Q\right\}  $. They can be divided into two
broad categories: the approximate and the exact methods. One example of the
approximate methods is using Gibbs samplings to generate \textquotedblleft
variates\textquotedblright\ (or \textquotedblleft
instantiations\textquotedblright) for $\mathcal{V}$, then using statistical
techniques to find an estimate for $\Pr\left\{  Q\right\}  $. (See Pearl,
1987.) In this paper, we present a method to compute $\Pr\left\{  Q\right\}  $
exactly, apart from precision or rounding errors.

Guo \& Hsu (2002) did a survey of the exact algorithms for the BNs, including
the two most well-known ones, namely the variable eliminations (Zhang \&
Poole, 1996; Dechter, 1999) and the clique-tree propagations (Lauritzen \&
Spiegelhalter, 1988; Lepar \& Shenoy, 1999). Other methods reviewed were the
message propagations in polytrees (Kim \& Pearl, 1983; Pearl 1986a, 1986b),
loop cutset conditioning (Pearl, 1986b; D\'{\i}ez, 1996), arc reversal/node
reduction (Shachter, 1990), symbolic probabilistic inference (Shachter
\textit{et al.}, 1990) and differential approach (Darwiche, 2003). We also
want to mention the more recent LAZY propagation algorithm (Madsen \& Jensen, 1999).

In this paper, we first present the \emph{border algorithm}.\ Like the
clique-tree propagation, instead of obtaining the joint probability
$\Pr\left\{  V_{1},...,V_{\ell}\right\}  $, the border algorithm breaks a
Bayesian network into smaller parts and calculate the marginal probabilities
of these parts, avoiding the exponential blow-ups associated with large
networks. In the next section, we first show how a BN can be so divided, in
such a way that its independency structure can be exploited. In Section 3, we
explain how to calculate the marginal probability of each part when there is
no observed evidence. In Section 4, we show how to calculate them, conditional
on some observed evidences.

In Section 5, we focus on a special kind of BN called the \textquotedblleft
polytrees,\textquotedblright\ and\ present in details, with some modifications
and within the border algorithm framework, the \textquotedblleft revised
polytree algorithm\textquotedblright\ by Peot \& Shachter (1991).

In Section 6, we present our \emph{parentless polytree method}, which, coupled
with the border algorithm, can convert any BN\ into a polytree. This part is
static,\ in that they need to be done only once, off-line, prior to any
dialogue with a user. Then we show the dynamic,\ on-line part of our method,
in which the conditional marginal probabilities can be calculated whenever new
evidences are entered or queries posed.

Finally, our discussions and summary are presented in Section 7.

\section{Partitioning a DAG}

In this section, we will show how a BN can be partitioned into smaller parts.

\subsection{The Set Relationships}

Consider a non-empty set of nodes $\mathcal{X}\subseteq\mathcal{V}$.\ We also call

\begin{enumerate}
\item $\mathcal{H}_{\mathcal{X}}=\left\{  \cup_{V\in\mathcal{X}}%
\mathcal{H}_{V}\right\}  \backslash\mathcal{X}$ the \textquotedblleft
parent\textquotedblright\ of $\mathcal{X}$. If $\mathcal{H}_{\mathcal{X}%
}=\varnothing$, we say $\mathcal{X}$ is \textquotedblleft
parentless\textquotedblright\ (or \textquotedblleft
ancestral\textquotedblright). For the BN $\mathbb{A}$, $\mathcal{H}_{\left\{
A,H\right\}  }=\left\{  C,D\right\}  $.

\item $\mathcal{L}_{\mathcal{X}}=\left\{  \cup_{V\in\mathcal{X}}%
\mathcal{L}_{V}\right\}  \backslash\left\{  \mathcal{X},\mathcal{H}%
_{\mathcal{X}}\right\}  $ the \textquotedblleft child\textquotedblright\ of
$\mathcal{X}$. If $\mathcal{L}_{\mathcal{X}}=\varnothing$, we say
$\mathcal{X}$ is \textquotedblleft childless.\textquotedblright\ For the BN
$\mathbb{A}$, $\mathcal{L}_{\left\{  A,H\right\}  }=\left\{  J,K\right\}  $.
(Although $D$ is a child of $A$, it is also a parent of $H$; so it is a member
of $\mathcal{H}_{\left\{  A,H\right\}  }$, not of $\mathcal{L}_{\left\{
A,H\right\}  }$.)

\item $\mathcal{K}_{\mathcal{X}}=\left\{  \cup_{V\in\mathcal{X}}{\mathcal{K}%
}_{V}\right\}  \backslash\left\{  \mathcal{X},\mathcal{H}_{\mathcal{X}%
},{\mathcal{L}}_{\mathcal{X}}\right\}  $ the \textquotedblleft
co-parent\textquotedblright\ of $\mathcal{X}$. If a child of $V\in\mathcal{X}$
is also in $\mathcal{X}$, then all its parents are in $\left\{  \mathcal{X}%
,\mathcal{H}_{\mathcal{X}}\right\}  $. Thus we are only concerned with the
children of $V$ in $\mathcal{L}_{\mathcal{X}}$. $\mathcal{K}_{\mathcal{X}}$
therefore can also be defined as $\left\{  \cup_{V\in{\mathcal{L}%
}_{\mathcal{X}}}{\mathcal{H}}_{V}\right\}  \backslash\left\{  \mathcal{X}%
,{\mathcal{L}}_{\mathcal{X}}\right\}  $. For BN $\mathbb{A}$, $\mathcal{K}%
_{\left\{  A,H\right\}  }=\left\{  B,G,I\right\}  $. If $\mathcal{K}%
_{\mathcal{X}}=\varnothing$, we say $\mathcal{X}$ is \textquotedblleft
co-parentless.\textquotedblright
\end{enumerate}

\subsection{\label{Rules}The Growing Parentless Set}

Consider a parentless set $\mathcal{P}\subseteq\mathcal{V}$. It is
\textquotedblleft growing\textquotedblright\ when it \textquotedblleft
recruits\textquotedblright\ new members. There are simple algorithms in the
literature that allow $\mathcal{P}$ to recruit a member in a \textquotedblleft
topological order\textquotedblright\ (that is, after all its parents), so that
it is always parentless. (For example, Koller \& Friedman, 2009, p.~1146.) Let
us call $\mathcal{D}=\mathcal{V}\backslash\mathcal{P}$ the \textquotedblleft
bottom part\textquotedblright\ of the BN. We present here an algorithm that
not only constructs a growing parentless $\mathcal{P}$, but also divides
$\mathcal{P}$ into two parts: $\mathcal{P}=\left\{  \mathcal{A},\mathcal{B}%
\right\}  $, where $\mathcal{A}$ is called the \textquotedblleft top
part,\textquotedblright\ and $\mathcal{B}$ the \textquotedblleft
border\textquotedblright\ that separates $\mathcal{A}$\ from $\mathcal{D}$. It
will become clear later why we wish to keep the size of border $\mathcal{B}$
(that is, the number of possible values of $\mathcal{B}$ and $\mathcal{H}%
_{\mathcal{B}}$) as small as possible.

We call the members of $\mathcal{A}$, $\mathcal{B}$ and $\mathcal{D}$ the
\textquotedblleft top variables,\textquotedblright\ the \textquotedblleft
border variables,\textquotedblright\ and the \textquotedblleft bottom
variables\textquotedblright, respectively.

For the initial top part $\mathcal{A}$, we start with $\mathcal{A}%
=\varnothing$. We use a co-parentless set of roots as the initial border
$\mathcal{B}$. There is at least one such set of roots in a BN. (Suppose a set
of roots has a non-root co-parent. Then if we trace through the ancestors of
this co-parent, we must encounter another set of roots. Again, if this set is
not co-parentless, we trace up further. Eventually, we must see a
co-parentless set of roots in a finite BN.) In our example, none of the roots
is co-parentless, but the set $\left\{  A,B\right\}  $ is.

All bottom\ variables will eventually join $\mathcal{P}$. However, we do not
choose which bottom variable to join next. This method does not give us
control over the membership of the top part $\mathcal{A}$. Instead, we first
decide which variable in $\mathcal{B}$ is to be \textquotedblleft
promoted\textquotedblright\ to $\mathcal{A}$. The promotion of a variable
$B\in\mathcal{B}$ may leave a \textquotedblleft hole\textquotedblright\ in the
border $\mathcal{B}$; thus $\mathcal{B}$ no longer separates $\mathcal{A}%
$\ from $\mathcal{D}$. This necessitates recruiting some bottom variables into
$\mathcal{B}$ to fill that hole, allowing $\mathcal{P}$ to grow. We call the
set of the bottom variables that are recruited into border $\mathcal{B}$ upon
the promotion of node $B$ the \textquotedblleft cohort\textquotedblright\ of
$B$ and denote it by $\mathcal{C}$. To fill the hole, $\mathcal{C}$ must
include at least the part of the \textquotedblleft Markov
blanket\textquotedblright\ of $B$ (that is, all its children and co-parents)
in $\mathcal{D}$. $\mathcal{C}$ may be empty, or may be more than what we need
for $\mathcal{B}$ to separate $\mathcal{A}$ and $\mathcal{D}$.

\begin{figure}
[ptb]
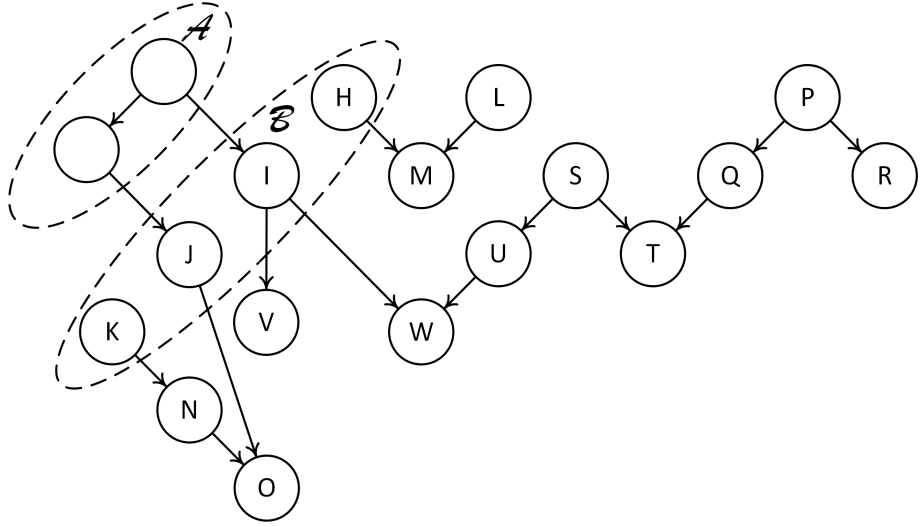

\begin{center}
\includegraphics[
natheight=2.763700in,
natwidth=4.768200in,
height=2.7637in,
width=4.7682in
]%
2%
\caption{A Bayesian Network}%
\label{F2}%
\end{center}
\end{figure}

For $\mathcal{P}$ to remain parentless, it is necessary that $\mathcal{H}%
_{\mathcal{C}}\subseteq\mathcal{P}$. Because all members of $\mathcal{C}$ are
separated from $\mathcal{A}$ by $\mathcal{B}$, $\mathcal{C}$ cannot have any
parent in $\mathcal{A}$. So we only need $\mathcal{H}_{\mathcal{C}}%
\subseteq\mathcal{B}$. Below are the many ways by which we can choose the next
variable $B\in\mathcal{B}$ to promote to $\mathcal{A}$, approximately in order
of preference to keep the size of $\mathcal{B}$ small:

\begin{enumerate}
\item $B$ has no bottom children, hence no bottom co-parent. Then
$\mathcal{C}=\varnothing$.

\item $B$ has bottom children, but no bottom co-parent. Then $\mathcal{C}%
=\mathcal{L}_{B}\cap\mathcal{D}$. (This is why we start $\mathcal{B}$ with a
co-parentless set of roots.)

\item $B$ has bottom co-parents, which are roots or have no bottom parents.
Then $\mathcal{C}=\left\{  \mathcal{L}_{B}\cup\mathcal{K}_{B}\right\}
\cap\mathcal{D}$. In Figure \ref{F2}, variable $J$ (having co-parent $N$, with
$\mathcal{H}_{N}=K\notin\mathcal{D}$) can be promoted with cohort $\left\{
N,O\right\}  $. Variable $H$ can also be promoted with cohort $\left\{
M,L\right\}  $, because its co-parent $L$ is a root.

\item $B$ is a fictitious variable $\varnothing$, the cohort of which is a
bottom variable having all parents in $\mathcal{P}$. In Figure \ref{F2}, we
can recruit variable $V$ (resulting in new border $\left\{  K,J,I,H,V\right\}
$).

\item $B$ is a fictitious variable $\varnothing$, the cohort of which is a
bottom root. In Figure \ref{F2}, we can recruit root $P$ (resulting in new
border $\left\{  K,J,I,H,P\right\}  $) or root $S$ (resulting in new border
$\left\{  K,J,I,H,S\right\}  $).

\item $B$ is any variable in $\mathcal{B}$, the cohort of which includes not
only its bottom children, but also all their bottom ancestors (hence the
bottom co-parents of $B$ and perhaps some roots). In Figure \ref{F2}, $I$ can
be promoted with cohort $\left\{  V,W,U,S\right\}  $.

\item $B$ is a fictitious variable $\varnothing$, the cohort of which includes
any bottom variable, together with all its bottom ancestors. Unless it is
necessary, the worst (but legal) strategy is to bring all bottom variables
simultaneously into $\mathcal{B}$.

\item $B$ is a fictitious variable $\varnothing$, the cohort of which is the
whole parentless set $\mathcal{P}^{\ast}\subseteq\mathcal{D}$. This is
equivalent to \textquotedblleft merging\textquotedblright\ $\mathcal{P}$ and
$\mathcal{P}^{\ast}$. In Figure \ref{F2}, we can merge $\mathcal{P=}\left\{
\mathcal{A},\mathcal{B}\right\}  $ with the parentless $\mathcal{P}^{\ast
}=\left\{  P,Q,R,S,T,U\right\}  $.

If $\mathcal{P}^{\ast}$ has already been divided into the top part
$\mathcal{A}^{\ast}$ and the border $\mathcal{B}^{\ast}$, then $\mathcal{A}%
^{\ast}$ can be merged with $\mathcal{A}$ and $\mathcal{B}^{\ast}$ with
$\mathcal{B}$. (This is equivalent to simultaneously promoting\ all members of
$\mathcal{A}^{\ast}$ after merging $\mathcal{P}$ and $\mathcal{P}^{\ast}$.) In
Figure \ref{F2}, if $\mathcal{B}^{\ast}=\left\{  U,T\right\}  $, then the new
border is $\left\{  K,J,I,H,U,T\right\}  $.
\end{enumerate}

%

We continue to use the notations such as $\mathcal{P}$, $\mathcal{A}$,
$\mathcal{B}$ and $\mathcal{D}$. However, we also define the initial top part
as $\mathcal{A}_{0}=\varnothing$, and denote the initial border, comprising of
a co-parentless set of roots, by $\mathcal{B}_{0}$.

At \textquotedblleft time\textquotedblright\ $j\geq1$,\ the variable promoted
to $\mathcal{A}_{j-1}$ (which may be a fictitious variable $\varnothing$) is
re-named as $V_{j}$; the resulting top part becomes $\mathcal{A}_{j}$. Thus,
for all $j\geq1$,%
\begin{equation}
\mathcal{A}_{j}=\left\{  \mathcal{A}_{j-1},V_{j}\right\}  =\left\{
V_{1},...,V_{j}\right\}  . \label{45}%
\end{equation}

Let $\mathcal{C}_{0}=\mathcal{B}_{0}$. For all $j\geq1$, the
cohort\ associated with $V_{j}$ is re-named as $\mathcal{C}_{j}$. After
promoting $V_{j}\in\mathcal{B}_{j-1}$ and recruiting $\mathcal{C}_{j}$, the
resulting border is, for all $j\geq1$,
\begin{equation}
\mathcal{B}_{j}=\left\{  \mathcal{B}_{j-1}\backslash V_{j},\mathcal{C}%
_{j}\right\}  , \label{41}%
\end{equation}
with $\mathcal{H}_{\mathcal{C}_{j}}\subseteq\mathcal{B}_{j-1}$.

Let $\mathcal{P}_{0}=\mathcal{B}_{0}$. The parentless set $\mathcal{P}$\ grows
cohort-by-cohort as, for all $j\geq1$,%
\begin{equation}
\mathcal{P}_{j}=\left\{  \mathcal{A}_{j},\mathcal{B}_{j}\right\}  =\left\{
\mathcal{A}_{j-1},V_{j}\right\}  \cup\left\{  \mathcal{B}_{j-1}\backslash
V_{j},\mathcal{C}_{j}\right\}  =\left\{  \mathcal{P}_{j-1},\mathcal{C}%
_{j}\right\}  =\cup_{k=0}^{j}\mathcal{C}_{k}. \label{42}%
\end{equation}
Eventually, all variables in $\mathcal{V}$ will join $\mathcal{P}$. Let
$\gamma$ be the time that this happens. We call $\mathcal{B}_{\gamma}$ the
\textquotedblleft last\textquotedblright\ border. Then $\mathcal{V}$ is
partitioned into disjoint sets as $\mathcal{V}=\mathcal{P}_{\gamma}=\cup
_{k=0}^{\gamma}\mathcal{C}_{k}$.

Let the bottom part at time $j$ ($0\leq j\leq\gamma$) be
\begin{equation}
\mathcal{D}_{j}=\mathcal{V}\backslash\mathcal{P}_{j}=\mathcal{P}_{\gamma
}\backslash\mathcal{P}_{j}=\cup_{k=j+1}^{\gamma}\mathcal{C}_{k}=\left\{
\mathcal{C}_{j+1},\mathcal{D}_{j+1}\right\}  . \label{44}%
\end{equation}

The above promotion rules do not result in a unique promotion order; and we do
not attempt to optimize here, so that the maximum size of all borders
$\mathcal{B}_{i}$ ($i=1,2,...,\gamma$) is as small as possible. We can
heuristically search among all members of $\mathcal{B}_{j}$ to identify the
node whose promotion leads to the smallest next border $\mathcal{B}_{j+1}$,
but this does not guarantee a global minimum. The above order of preference
may help.

We show the results obtained by one particular promotion order for the BN
$\mathbb{A}$ in Table 1. The last column shows the rule we use to promote
$V_{i}$. The function $\Phi\left(  \mathcal{C}_{i}\right)  $ will be
introduced later.

\begin{table}[h]
\begin{center}
$%
\begin{array}
[c]{|c|c|c|c|c|c|}\hline
\text{time }i & V_{i} & \mathcal{B}_{i}\backslash\mathcal{C}_{i} &
\mathcal{C}_{i} & \Phi\left(  C_{i}\right)  & \text{Rule}\\\hline
0 & \varnothing & \varnothing & A,B & \Pr\left\{  A\right\}  \Pr\left\{
B\right\}  & \\\hline
1 & A & B & C,D,F & \Pr\left\{  C|A,B\right\}  \Pr\left\{  D|A,B\right\}
\Pr\left\{  F|A,B\right\}  & 2\\\hline
2 & B & C,D,F & \varnothing & 1 & 1\\\hline
3 & C & D,F & H & \Pr\left\{  H|C,D\right\}  & 2\\\hline
4 & D & F,H & I & \Pr\left\{  I|D,F\right\}  & 2\\\hline
5 & F & H,I & \varnothing & 1 & 1\\\hline
6 & H & I & J,K,G & \Pr\left\{  J|G,H\right\}  \Pr\left\{  K|H,I\right\}
\Pr\left\{  G\right\}  & 3\\\hline
7 & G & I,J,K & \varnothing & 1 & 1\\\hline
\gamma=8 & I & J,K & L & \Pr\left\{  L|I\right\}  & 2\\\hline
\end{array}
$
\end{center}
\caption{The borders obtained for the BN $\mathbb{A}$ from one particular
promotion order }%
\end{table}

To keep $\left\{  \mathcal{P}_{i},i=1,2,...,\gamma\right\}  $ parentless, some
borders $\mathcal{B}_{i}$ may have more members than what required to separate
$\mathcal{A}_{i}$ and $\mathcal{D}_{i}$. Rule 1 is useful in this case, to
reduce the membership of $\mathcal{B}_{i}$ to its minimum. For example, it was
used to reduce $\mathcal{B}_{4}=\left\{  F,H,I\right\}  $ to $\mathcal{B}%
_{5}=\left\{  H,I\right\}  $.

\begin{figure}
[ptb]
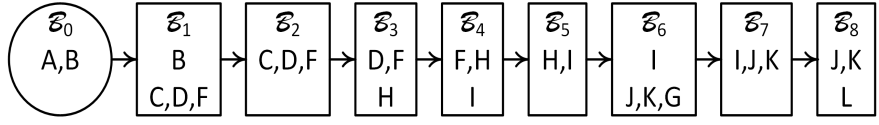

\begin{center}
\includegraphics[
natheight=0.699200in,
natwidth=4.569900in,
height=0.6992in,
width=4.5699in
]%
3%
\caption{A border chain for the BN $\mathbb{A}$}%
\label{F5}%
\end{center}
\end{figure}

Here we construct a directed chain of possibly overlapping borders $\left\{
\mathcal{B}_{i},i=1,2,...,\gamma\right\}  $, called the \textquotedblleft
border chain.\textquotedblright\ A border chain is Markovian, in the sense
that the knowledge of $\mathcal{B}_{j}$ is sufficient for the study of
$\mathcal{B}_{j+1}$. Figure \ref{F5} shows the corresponding border chain for
the BN $\mathbb{A}$.%

\section{Inferences without Evidences}

In this section, we explain how the \textquotedblleft prior marginal
probability\textquotedblright\ $\Pr\left\{  \mathcal{B}_{i}\right\}  $ can be
calculated, assuming that no variable is observed taking any value.

\subsection{The Parentless Set Probabilities}

We first present the following important lemma, which is based on a simple
observation that \emph{in a BN, a parentless set of nodes and its
parametrization is a Bayesian sub-network}:

\begin{lemma}
\label{P0}If $\mathcal{P}\subseteq\mathcal{V}$ is parentless, then%
\[
\Pr\left\{  \mathcal{P}\right\}  =\prod_{V\in\mathcal{P}}\Pr\left\{
V|{\mathcal{H}}_{V}\right\}  .
\]

\end{lemma}

\begin{proof}
The lemma follows from Equation (\ref{1})
\end{proof}

For the BN $\mathbb{A}$, as $\left\{  A,B,D\right\}  $ is parentless,%
\[
\Pr\left\{  A,B,D\right\}  =\Pr\left\{  A\right\}  \Pr\left\{  B\right\}
\Pr\left\{  D|A,B\right\}  .
\]

We do not use $\left\{  A,B,D\right\}  $ however, because $D$ alone does not
separate $\left\{  A,B\right\}  $ from the rest of the network. So for a
general BN, we start with the parentless $\mathcal{P}_{0}=\mathcal{B}%
_{0}=\mathcal{C}_{0}$ and define%
\begin{equation}
\Phi\left(  \mathcal{C}_{0}\right)  =\Pr\left\{  \mathcal{P}_{0}\right\}
=\Pr\left\{  \mathcal{B}_{0}\right\}  =\Pr\left\{  \mathcal{C}_{0}\right\}
=\prod_{V\in\mathcal{B}_{0}}\Pr\left\{  V\right\}  . \label{8}%
\end{equation}

Recall that, when $V_{j}$ is promoted, it brings a cohort $\mathcal{C}_{j}$
into $\mathcal{B}_{j-1}$. By the Markovian assumption, we do not need the
whole $\Pr\left\{  \mathcal{C}_{j}|\mathcal{P}_{j-1}\right\}  $, but only
$\Pr\left\{  \mathcal{C}_{j}|\mathcal{H}_{\mathcal{C}_{j}}\right\}  $ with
$\mathcal{H}_{\mathcal{C}_{j}}\subseteq\mathcal{B}_{j-1}\subseteq
\mathcal{P}_{j-1}$. For all $0\leq j\leq\gamma$, let us denote the
\textquotedblleft cohort probability tables\textquotedblright\ $\Pr\left\{
\mathcal{C}_{j}|\mathcal{H}_{\mathcal{C}_{j}}\right\}  $ by $\Phi\left(
\mathcal{C}_{j}\right)  $. If $\mathcal{C}_{j}=\varnothing$, we set
$\Phi\left(  \mathcal{C}_{j}\right)  =1$.\ Column 5 of Table 1 shows the
cohort probability tables for the BN $\mathbb{A}$.

\begin{theorem}
\label{P}For all $0\leq j\leq\gamma$,%
\[
\Pr\left\{  \mathcal{P}_{j}\right\}  =\Phi\left(  \mathcal{C}_{j}\right)
\Pr\left\{  \mathcal{P}_{j-1}\right\}  =\prod_{k=0}^{j}\Phi\left(
\mathcal{C}_{k}\right)  .
\]

\end{theorem}

\begin{proof}
From Equation (\ref{42}),%
\begin{align*}
\Pr\left\{  \mathcal{P}_{j}\right\}   &  =\Pr\left\{  \mathcal{P}%
_{j-1},\mathcal{C}_{j}\right\}  =\Pr\left\{  \mathcal{C}_{j}|\mathcal{P}%
_{j-1}\right\}  \Pr\left\{  \mathcal{P}_{j-1}\right\} \\
&  =\Pr\left\{  \mathcal{C}_{j}|\mathcal{H}_{\mathcal{C}_{j}}\right\}
\Pr\left\{  \mathcal{P}_{j-1}\right\}  =\Phi\left(  \mathcal{C}_{j}\right)
\Pr\left\{  \mathcal{P}_{j-1}\right\}  .
\end{align*}
The theorem follows because $\Pr\left\{  \mathcal{P}_{0}\right\}  =\Phi\left(
\mathcal{C}_{0}\right)  $.
\end{proof}

Theorem \ref{P} can be used to obtain the joint probability of $\mathcal{P}%
_{j}$ when $j$ is small. For example,%
\[
\Pr\left\{  \mathcal{P}_{1}\right\}  =\Pr\left\{  A,B,C,D,F\right\}
=\Phi\left(  \mathcal{C}_{0}\right)  \Phi\left(  \mathcal{C}_{1}\right)  .
\]
However, as $\mathcal{P}$ grows, eventually we return to Equation (\ref{1}):
$\Pr\left\{  \mathcal{V}\right\}  =\Pr\left\{  \mathcal{P}_{\gamma}\right\}
=\prod_{k=0}^{\gamma}\Phi\left(  \mathcal{C}_{k}\right)  $, which is what we
did not want to use in the first place.

Fortunately, as we will see in the next section, what we have here is a very
\textquotedblleft cruel\textquotedblright\ parentless set of nodes that, after
promoting and extracting information from a member, immediately
\textquotedblleft eliminates\textquotedblright\ that member!

\subsection{The Border Probabilities}

We now show how $\Pr\left\{  \mathcal{B}_{j}\right\}  $ can be recursively
calculated from $\Pr\left\{  \mathcal{B}_{j-1}\right\}  $:

\begin{theorem}
\label{FW}For all $1\leq j\leq\gamma$,%
\[
\Pr\left\{  \mathcal{B}_{j}\right\}  =\sum_{V_{j}}\Phi\left(  \mathcal{C}%
_{j}\right)  \Pr\left\{  \mathcal{B}_{j-1}\right\}  .
\]

\end{theorem}

\begin{proof}
For all $\mathcal{P}_{j}$ ($1\leq j\leq\gamma$), our strategy is not to
eliminate all members of $\mathcal{A}_{j}$ at the same time, in the form of%
\[
\Pr\left\{  \mathcal{B}_{j}\right\}  =\sum_{\mathcal{A}_{j}}\Pr\left\{
\mathcal{A}_{j},\mathcal{B}_{j}\right\}  =\sum_{\mathcal{A}_{j}}\Pr\left\{
\mathcal{P}_{j}\right\}  =\sum_{\mathcal{A}_{j}}\Phi\left(  \mathcal{C}%
_{j}\right)  \Pr\left\{  \mathcal{P}_{j-1}\right\}  .
\]
Rather, we eliminate the variables in $\mathcal{A}_{j}$ one-by-one: After
variable $V_{j}$ is promoted into $\mathcal{A}_{j}=\left\{  \mathcal{A}%
_{j-1},V_{j}\right\}  $, it is immediately eliminated. In other words, because
the scope of $\Phi\left(  \mathcal{C}_{j}\right)  $ (which is $\left\{
\mathcal{C}_{j},\mathcal{H}_{\mathcal{C}_{j}}\right\}  \subseteq\left\{
\mathcal{C}_{j},\mathcal{B}_{j-1}\right\}  $) does not include any member of
$\mathcal{A}_{j-1}$, as far as the summing out of $\mathcal{A}_{j-1}$ is
concerned, $\Phi\left(  \mathcal{C}_{j}\right)  $ can be treated as a
constant:%
\[
\Pr\left\{  \mathcal{B}_{j}\right\}  =\sum_{\left\{  \mathcal{A}_{j-1}%
,V_{j}\right\}  }\Phi\left(  \mathcal{C}_{j}\right)  \Pr\left\{
\mathcal{P}_{j-1}\right\}  =\sum_{V_{j}}\Phi\left(  \mathcal{C}_{j}\right)
\left(  \sum_{\mathcal{A}_{j-1}}\Pr\left\{  \mathcal{A}_{j-1},\mathcal{B}%
_{j-1}\right\}  \right)  ,
\]
hence the theorem.
\end{proof}

There must be one value of $\tau$ ($0\leq\tau\leq\gamma$) such that
$\Pr\left\{  \mathcal{B}_{\tau}\right\}  $ can be calculated. At least, from
Equation (\ref{8}), we know $\Pr\left\{  \mathcal{B}_{0}\right\}  $. Starting
with $\Pr\left\{  \mathcal{B}_{\tau}\right\}  $, we can calculate $\Pr\left\{
\mathcal{B}_{j}\right\}  $ for all $\tau<j\leq\gamma$ recursively by the above theorem.

We call our algorithm the \textquotedblleft border algorithm\textquotedblright%
\ because it breaks the large joint probability $\Pr\left\{  \mathcal{V}%
\right\}  $ down into many smaller border probabilities $\Pr\left\{
\mathcal{B}_{j}\right\}  $, thus avoiding the exponential blow-ups associated
with large networks. That is why we want the size of the largest border to be
as small as possible.

We now show how the marginal probabilities can be obtained given some evidences.

\section{Inferences with Evidences}

For a variable $V\in\mathcal{V}$, let $\operatorname*{Va}\left(  V\right)  $
be the set of possible values of $V$ such that $\Pr\left\{  V=v|{\mathcal{H}%
}_{V}\right\}  >0$. A variable $E$ is said to be an \textquotedblleft evidence
variable\textquotedblright\ if it is observed taking value only in a subset
$\operatorname*{Va}^{e}\left(  E\right)  \subset\operatorname*{Va}\left(
E\right)  $. Variable $V$ is non-evidential if $\operatorname*{Va}^{e}\left(
V\right)  =\operatorname*{Va}\left(  V\right)  $. For example, suppose
$\operatorname*{Va}\left(  X\right)  =\left\{  1,2,3\right\}  $. If $X$ is
observed not taking value 3, then it is evidential with $\operatorname*{Va}%
^{e}\left(  X\right)  =\left\{  1,2\right\}  $. Let $\mathcal{E}$ be the set
of all evidence variables.

Consider set $\mathcal{Y}=\left\{  Y_{1},...,Y_{n}\right\}  \subseteq
\mathcal{V}$. We denote the event that $\mathcal{Y}$ occurs by%
\[
\left[  \mathcal{Y}\right]  =\left\{  \mathcal{Y}\in\operatorname*{Va}%
\nolimits^{e}\left(  Y_{1}\right)  \times...\times\operatorname*{Va}%
\nolimits^{e}\left(  Y_{n}\right)  \right\}  .
\]
If $\mathcal{Y\cap E=\varnothing}$, $\left[  \mathcal{Y}\right]  $ is a sure
event. Thus $\left[  \mathcal{Y}\right]  =\left[  \mathcal{Y\cap E}\right]  $.

One of the most important tasks in analyzing a BN is to calculate $\Pr\left\{
Q|\left[  \mathcal{E}\right]  \right\}  $, which is known as the
\textquotedblleft posterior marginal distribution\textquotedblright\ (or the
\textquotedblleft conditional marginal distribution\textquotedblright) of a
\textquotedblleft query variable\textquotedblright\ $Q$.\ 

For the rest of this paper, we will show how we can first calculate the joint
distribution table $\Pr\left\{  Q,\left[  \mathcal{E}\right]  \right\}  $ for
all possible values of $Q$. This allows us to calculate $\Pr\left\{  \left[
\mathcal{E}\right]  \right\}  =\sum_{Q}\Pr\left\{  Q,\left[  \mathcal{E}%
\right]  \right\}  $ and then%
\[
\Pr\left\{  Q|\left[  \mathcal{E}\right]  \right\}  =\frac{\Pr\left\{
Q,\left[  \mathcal{E}\right]  \right\}  }{\sum_{Q}\Pr\left\{  Q,\left[
\mathcal{E}\right]  \right\}  }=\frac{\Pr\left\{  Q,\left[  \mathcal{E}%
\right]  \right\}  }{\Pr\left\{  \left[  \mathcal{E}\right]  \right\}  }.
\]

\subsection{The Evidence Indicator Columns}

Consider a table having $t$ rows, each row corresponding to an instantiation
of a set of variables $\mathcal{X}\subseteq\mathcal{V}$. If an evidence
variable $E$ is in $\mathcal{X}$, we define an \textquotedblleft evidence
indicator\ column\textquotedblright\ $\boldsymbol{I}_{E}$ having size $t$,
such that it takes value $1$\ if $E\in\operatorname*{Va}^{e}\left(  E\right)
$, and $0$ otherwise. For a non-evidence variable $V\in\mathcal{X}$, we also
define the column $\boldsymbol{I}_{V}$ having size $t$, all members of which
are $1$.

We will use the following notation for a set of nodes $\mathcal{X}%
=\mathcal{\mathcal{Y}}\cup\mathcal{\mathcal{Z}}\subseteq\mathcal{V}$:%
\[
\boldsymbol{I}_{\mathcal{X}}=\prod_{V\in\mathcal{X}}\boldsymbol{I}%
_{V}=\boldsymbol{I}_{\mathcal{\mathcal{Y}}\cup\mathcal{\mathcal{Z}}%
}=\boldsymbol{I}_{\mathcal{Y}}\boldsymbol{I}_{\mathcal{Z}}.
\]
Multiplying a table having scope $\mathcal{X}$ with column $\boldsymbol{I}%
_{\mathcal{X}}$ is equivalent to zeroing out the rows inconsistent with the
evidences in $\mathcal{X}$.

For the CPT $\Pr\left\{  V|{\mathcal{H}}_{V}\right\}  $, we define its
\textquotedblleft reduced\ CPT\textquotedblright\ as:%
\begin{equation}
\Pr\nolimits_{r}\left\{  V|{\mathcal{H}}_{V}\right\}  =\Pr\left\{
V|{\mathcal{H}}_{V}\right\}  \boldsymbol{I}_{V\cup{\mathcal{H}}_{V}}.
\label{27}%
\end{equation}

Previously we defined the cohort probability tables $\Phi\left(
\mathcal{C}_{j}\right)  =\Pr\left\{  \mathcal{C}_{j}|\mathcal{H}%
_{\mathcal{C}_{j}}\right\}  $ for all $0\leq j\leq\gamma$. We now define the
\textquotedblleft reduced cohort probability tables\textquotedblright\ as%
\[
\phi\left(  \mathcal{C}_{j}\right)  =\Phi\left(  \mathcal{C}_{j}\right)
\boldsymbol{I}_{\mathcal{C}_{j}\cup\mathcal{H}_{\mathcal{C}_{j}}}.
\]

The following lemma is the evidential version of Equation (\ref{10}):

\begin{lemma}
\label{E2}Given $\mathcal{Y},\mathcal{Z\subseteq V}$, $\mathcal{Y}%
\cap\mathcal{Z}=\varnothing$, then with scope $\mathcal{Z}$,%
\[
\Pr\left\{  \mathcal{Z},\left[  \mathcal{Y}\right]  \right\}  =\sum
_{\mathcal{Y}}\Pr\left\{  \mathcal{Z},\mathcal{Y}\right\}  \boldsymbol{I}%
_{\mathcal{Y}}.
\]

\end{lemma}

\begin{proof}
If $\mathcal{Y\cap E=\varnothing}$, we have Equation (\ref{10}) because
$\boldsymbol{I}_{\mathcal{Y}}=1$ and $\Pr\left\{  \mathcal{Z},\left[
\mathcal{Y}\right]  \right\}  =\Pr\left\{  \mathcal{Z}\right\}  $. Suppose
$\mathcal{Y}=\left\{  Y_{1},...,Y_{n}\right\}  $ is observed taking value in
$\operatorname*{Va}\nolimits^{e}\left(  Y_{1}\right)  \times...\times
\operatorname*{Va}\nolimits^{e}\left(  Y_{n}\right)  $. For every fixed
$\mathcal{Z}=z$, summing out $\mathcal{Y}$ yields:%
\begin{align*}
&  \sum_{\mathcal{Y}}\Pr\left\{  z,\mathcal{Y}\right\}  \boldsymbol{I}%
_{\mathcal{Y}}\\
&  =\sum_{Y_{1}}...\sum_{Y_{n-1}}\left(  \sum_{Y_{n}}\Pr\left\{  z,Y_{1}%
=y_{1},...,Y_{n-1}=y_{n-1},Y_{n}=y_{n}\right\}  \prod_{i=1}^{n}\boldsymbol{I}%
_{Y_{i}}\right) \\
&  =\sum_{Y_{1}}...\left(  \sum_{Y_{n-1}}\Pr\left\{  z,Y_{1}=y_{1}%
,...,Y_{n-1}=y_{n-1},Y_{n}\in\operatorname*{Va}\nolimits^{e}\left(
Y_{n}\right)  \right\}  \prod_{i=1}^{n-1}\boldsymbol{I}_{Y_{i}}\right) \\
&  =\sum_{Y_{1}}\Pr\left\{  z,Y_{1}=y_{1},Y_{2}\in\operatorname*{Va}%
\nolimits^{e}\left(  Y_{2}\right)  ,...,Y_{n}\in\operatorname*{Va}%
\nolimits^{e}\left(  Y_{n}\right)  \right\}  \boldsymbol{I}_{Y_{1}}%
=\Pr\left\{  z,\left[  \mathcal{Y}\right]  \right\}  .
\end{align*}

\end{proof}

We are now ready to obtain the necessary information for the calculations of
$\Pr\left\{  Q,\left[  \mathcal{E}\right]  \right\}  $.

\subsection{The Downward Pass for the Top Evidences}

We first consider the \textquotedblleft top evidences\textquotedblright%
\ within the top part $\mathcal{A}_{j}$, and define the following notation:
For all $0\leq j\leq\gamma$, by Lemma \ref{E2},
\begin{align}
\Pi\left(  \mathcal{B}_{j}\right)   &  =\Pr\left\{  \mathcal{B}_{j},\left[
\mathcal{A}_{j}\right]  \right\}  \boldsymbol{I}_{\mathcal{B}_{j}}\nonumber\\
&  =\sum_{\mathcal{A}_{j}}\Pr\left\{  \mathcal{B}_{j},\mathcal{A}_{j}\right\}
\boldsymbol{I}_{\mathcal{A}_{j}\cup\mathcal{B}_{j}}=\sum_{\mathcal{A}_{j}}%
\Pr\left\{  \mathcal{P}_{j}\right\}  \boldsymbol{I}_{\mathcal{P}_{j}}\text{.}
\label{49}%
\end{align}

The following theorem is the evidential version of Theorem \ref{FW}:

\begin{theorem}
\label{FE} For all $1\leq j\leq\gamma$,
\[
\Pi\left(  \mathcal{B}_{j}\right)  =\sum_{V_{j}}\phi\left(  \mathcal{C}%
_{j}\right)  \Pi\left(  \mathcal{B}_{j-1}\right)  .
\]

\end{theorem}

\begin{proof}
Because ${\mathcal{H}}_{{\mathcal{C}}_{j}}\subseteq{\mathcal{P}}_{j-1}$, we
have ${\mathcal{P}}_{j}=C_{j}\cup{\mathcal{P}}_{j-1}=C_{j}\cup{\mathcal{H}%
}_{{\mathcal{C}}_{j}}\cup{\mathcal{P}}_{j-1}$. From Definition (\ref{49}) and
Theorem \ref{P},%
\begin{align*}
\Pi\left(  \mathcal{B}_{j}\right)   &  =\sum_{\mathcal{A}_{j}}\Pr\left\{
\mathcal{P}_{j}\right\}  \boldsymbol{I}_{\mathcal{P}_{j}}=\sum_{\mathcal{A}%
_{j}}\Phi\left(  \mathcal{C}_{j}\right)  \boldsymbol{I}_{\mathcal{C}_{j}%
\cup\mathcal{H}_{\mathcal{C}_{j}}}\Pr\left\{  \mathcal{P}_{j-1}\right\}
\boldsymbol{I}_{\mathcal{P}_{j-1}}\\
&  =\sum_{\mathcal{A}_{j}}\phi\left(  \mathcal{C}_{j}\right)  \Pr\left\{
\mathcal{P}_{j-1}\right\}  \boldsymbol{I}_{\mathcal{P}_{j-1}}.
\end{align*}
From Equation (\ref{45}), and as the scope of $\phi\left(  \mathcal{C}%
_{j}\right)  $ (which is $\left\{  \mathcal{C}_{j},\mathcal{H}_{\mathcal{C}%
_{j}}\right\}  $) is not in $\mathcal{A}_{j-1}$,%
\[
\Pi\left(  \mathcal{B}_{j}\right)  =\sum_{V_{j}}\phi\left(  \mathcal{C}%
_{j}\right)  \sum_{\mathcal{A}_{j-1}}\Pr\left\{  \mathcal{P}_{j-1}\right\}
\boldsymbol{I}_{\mathcal{P}_{j-1}}=\sum_{V_{j}}\phi\left(  \mathcal{C}%
_{j}\right)  \Pi\left(  \mathcal{B}_{j-1}\right)  .
\]

\end{proof}

There must be one value of $\nu$ ($0\leq\nu\leq\gamma$) such that $\Pi\left(
\mathcal{B}_{\nu}\right)  $ can be calculated. Let $\alpha$ be the first time
an evidence variable is recruited into $\mathcal{P}$. For all $0\leq j<\alpha
$, $\mathcal{A}_{j}$ has no evidence and $\boldsymbol{I}_{\mathcal{B}_{j}}=1$;
thus $\Pi\left(  \mathcal{B}_{j}\right)  =\Pr\left\{  \mathcal{B}_{j}\right\}
$. $\mathcal{A}_{\alpha}$ also has no evidence and thus,
\begin{equation}
\Pi\left(  \mathcal{B}_{\alpha}\right)  =\Pr\left\{  \mathcal{B}_{\alpha
}\right\}  \boldsymbol{I}_{\mathcal{B}_{\alpha}}. \label{47}%
\end{equation}
Starting with $\Pi\left(  \mathcal{B}_{\nu}\right)  $, we can calculate
$\Pi\left(  \mathcal{B}_{j}\right)  $ recursively for all $\nu<j\leq\gamma$ by
the above theorem.

For the BN $\mathbb{A}$, assume $\mathcal{E}=\left\{  H=h,K=k\right\}  $.
Since $H\in\mathcal{C}_{3}$, $\alpha=3$. Thus
\begin{align*}
\Pi\left(  \mathcal{B}_{0}\right)   &  =\Pr\left\{  \mathcal{B}_{0}\right\}
=\Pr\left\{  A,B\right\}  ;\text{ }\\
\Pi\left(  \mathcal{B}_{1}\right)   &  =\Pr\left\{  \mathcal{B}_{1}\right\}
=\Pr\left\{  B,C,D,F\right\}  ;\text{ }\\
\Pi\left(  \mathcal{B}_{2}\right)   &  =\Pr\left\{  \mathcal{B}_{2}\right\}
=\Pr\left\{  C,D,F\right\}  ;\\
\Pi\left(  \mathcal{B}_{3}\right)   &  =\Pr\left\{  D,F,H\right\}
\boldsymbol{I}_{H}=\Pr\left\{  D,F,h\right\}  .
\end{align*}

\begin{enumerate}
\item Border $\mathcal{B}_{4}=\left\{  F,H,I\right\}  $ has $V_{4}=D$ and
$\mathcal{C}_{4}=I$:%
\begin{align*}
\Pi\left(  \mathcal{B}_{4}\right)   &  =\sum_{D}\phi\left(  \mathcal{C}%
_{4}\right)  \Pi\left(  \mathcal{B}_{3}\right)  =\sum_{D}\Pr\left\{
I|D,F\right\}  \Pr\left\{  D,F,h\right\} \\
&  =\sum_{D}\Pr\left\{  D,F,I,h\right\}  =\Pr\left\{  F,I,h\right\}  .
\end{align*}

\item Border $\mathcal{B}_{5}=\left\{  H,I\right\}  $ has $V_{5}=F$ and
$\mathcal{C}_{5}=\varnothing$:%
\[
\Pi\left(  \mathcal{B}_{5}\right)  =\sum_{F}\phi\left(  \mathcal{C}%
_{5}\right)  \Pi\left(  \mathcal{B}_{4}\right)  =\sum_{F}\Pr\left\{
F,I,h\right\}  =\Pr\left\{  I,h\right\}  .
\]

\item Border $\mathcal{B}_{6}=\left\{  I,J,K,G\right\}  $ has $V_{6}=H=h$ and
$\mathcal{C}_{6}=\left\{  J,K,G\right\}  $:%
\begin{align*}
\Pi\left(  \mathcal{B}_{6}\right)   &  =\phi\left(  \mathcal{C}_{6}\right)
\Pi\left(  \mathcal{B}_{5}\right) \\
&  =\Pr\left\{  J|G,h\right\}  \Pr\left\{  k|I,h\right\}  \Pr\left\{
G\right\}  \Pr\left\{  I,h\right\}  =\Pr\left\{  I,J,G,k,h\right\}  .
\end{align*}

\item Border $\mathcal{B}_{7}=\left\{  I,J,K\right\}  $ has $V_{7}=G$ and
$\mathcal{C}_{7}=\varnothing$:%
\[
\Pi\left(  \mathcal{B}_{7}\right)  =\sum_{G}\phi\left(  \mathcal{C}%
_{7}\right)  \Pi\left(  \mathcal{B}_{6}\right)  =\sum_{G}\Pr\left\{
I,J,G,k,h\right\}  =\Pr\left\{  I,J,k,h\right\}  .
\]

\item Border $\mathcal{B}_{8}=\left\{  J,K,L\right\}  $ has $V_{8}=I$ and
$\mathcal{C}_{8}=L$:%
\[
\Pi\left(  \mathcal{B}_{8}\right)  =\sum_{I}\phi\left(  \mathcal{C}%
_{8}\right)  \Pi\left(  \mathcal{B}_{7}\right)  =\sum_{I}\Pr\left\{
L|I\right\}  \Pr\left\{  I,J,k,h\right\}  =\Pr\left\{  J,L,k,h\right\}  .
\]

\end{enumerate}

\subsection{The Upward Pass for the Bottom Evidences}

Moving downward border-by-border from $\mathcal{B}_{0}$ to $\mathcal{B}_{j}$,
we can only collect information about the top evidences inside $\mathcal{P}%
_{j}$. To collect information about the bottom evidences inside $\mathcal{D}%
_{j}$, we move upward\ from the last border $\mathcal{B}_{\gamma}$ to
$\mathcal{B}_{j}$.

In the downward passes, we make use of the parentless property of
$\mathcal{P}_{j}$; in the upward passes we need the fact that the border
$\mathcal{B}_{j}$ separates $\mathcal{A}_{j}$ and $\mathcal{D}_{j}%
=\mathcal{V}\backslash\mathcal{P}_{j}$. Thus, to study $\mathcal{D}_{j}$, we
do not need the information of the whole $\mathcal{P}_{j}$, but only of
$\mathcal{B}_{j}$.

We first present the following lemma:

\begin{lemma}
\label{P3}For all $1\leq j\leq\gamma$,%
\[
\Pr\left\{  \mathcal{D}_{j-1}|\mathcal{B}_{j-1}\right\}  =\prod_{k=j}^{\gamma
}\Phi\left(  \mathcal{C}_{k}\right)  =\Phi\left(  \mathcal{C}_{j}\right)
\Pr\left\{  \mathcal{D}_{j}|\mathcal{B}_{j}\right\}  .
\]

\end{lemma}

\begin{proof}
For all $1\leq j\leq\gamma$, as both $\mathcal{V}=\mathcal{P}_{\gamma
}=\left\{  \mathcal{D}_{j-1},\mathcal{P}_{j-1}\right\}  $ and $\mathcal{P}%
_{j-1}$ are parentless, from Theorem \ref{P},%
\begin{align*}
\Pr\left\{  \mathcal{P}_{\gamma}\right\}   &  =\prod_{k=0}^{\gamma}\Phi\left(
\mathcal{C}_{k}\right)  =\Pr\left\{  \mathcal{D}_{j-1},\mathcal{P}%
_{j-1}\right\}  =\Pr\left\{  \mathcal{D}_{j-1}|\mathcal{P}_{j-1}\right\}
\Pr\left\{  \mathcal{P}_{j-1}\right\} \\
&  =\Pr\left\{  \mathcal{D}_{j-1}|\mathcal{B}_{j-1}\right\}  \prod_{k=0}%
^{j-1}\Phi\left(  \mathcal{C}_{k}\right)  .
\end{align*}

Assuming all $\Pr\left\{  V|{\mathcal{H}}_{V}\right\}  >0$,%
\[
\Pr\left\{  \mathcal{D}_{j-1}|\mathcal{B}_{j-1}\right\}  =\prod_{k=j}^{\gamma
}\Phi\left(  \mathcal{C}_{k}\right)  =\Phi\left(  \mathcal{C}_{j}\right)
\prod_{k=j+1}^{\gamma}\Phi\left(  \mathcal{C}_{k}\right)  =\Phi\left(
\mathcal{C}_{j}\right)  \Pr\left\{  \mathcal{D}_{j}|\mathcal{B}_{j}\right\}
.
\]

\end{proof}

We define the following notation: For all $0\leq j\leq\gamma-1$, by Lemma
\ref{E2},%
\begin{equation}
\Lambda\left(  \mathcal{B}_{j}\right)  =\Pr\left\{  \left[  \mathcal{D}%
_{j}\right]  |\mathcal{B}_{j}\right\}  \boldsymbol{I}_{\mathcal{B}_{j}}%
=\sum_{\mathcal{D}_{j}}\Pr\left\{  \mathcal{D}_{j}|\mathcal{B}_{j}\right\}
\boldsymbol{I}_{\mathcal{D}_{j}\cup\mathcal{B}_{j}}. \label{55}%
\end{equation}
Since $\mathcal{D}_{\gamma}=\varnothing$, we also define $\Lambda\left(
\mathcal{B}_{\gamma}\right)  =\boldsymbol{I}_{\mathcal{B}_{\gamma}}$.

Although we write $\Lambda\left(  \mathcal{B}_{j}\right)  $, the scope of
$\Lambda\left(  \mathcal{B}_{j}\right)  $ may not be the whole $\mathcal{B}%
_{j}$, because $\mathcal{B}_{j}$ may have more variables than the minimal set
needed to separate $\mathcal{A}_{j}$ and $\mathcal{D}_{j}$. For example, in
the BN $\mathbb{A}$, while $\mathcal{B}_{4}=\left\{  F,H,I\right\}  $, we only
need $\left\{  H,I\right\}  $ for the study of $\mathcal{D}_{4}=\left\{
G,J,K,L\right\}  $.

\begin{theorem}
\label{B}For all $1\leq j\leq\gamma$,%
\[
\Lambda\left(  \mathcal{B}_{j-1}\right)  =\sum_{\mathcal{C}_{j}}\phi\left(
\mathcal{C}_{j}\right)  \Lambda\left(  \mathcal{B}_{j}\right)  .
\]

\end{theorem}

\begin{proof}
From Equation (\ref{44}), $\mathcal{D}_{j-1}=\left\{  \mathcal{C}%
_{j},\mathcal{D}_{j}\right\}  $. From Equation (\ref{41}), $\mathcal{B}%
_{j-1}\cup\mathcal{C}_{j}=V_{j}\cup\mathcal{B}_{j}$. Also, if $V_{j}%
\neq\varnothing$, then its cohort $\mathcal{C}_{j}$ must include all its
bottom children, or $V_{j}\in\mathcal{H}_{\mathcal{C}_{j}}\subseteq
\mathcal{B}_{j-1}$. Thus,%
\begin{align*}
\mathcal{B}_{j-1}\cup\mathcal{D}_{j-1}  &  =\mathcal{B}_{j-1}\cup\left\{
\mathcal{C}_{j}\cup\mathcal{H}_{\mathcal{C}_{j}}\right\}  \cup\left\{
\mathcal{C}_{j}\cup\mathcal{D}_{j}\right\} \\
&  =V_{j}\cup\mathcal{B}_{j}\cup\mathcal{H}_{\mathcal{C}_{j}}\cup\left\{
\mathcal{C}_{j}\cup\mathcal{D}_{j}\right\}  =\mathcal{B}_{j}\cup
\mathcal{H}_{\mathcal{C}_{j}}\cup\mathcal{C}_{j}\cup\mathcal{D}_{j}.
\end{align*}

From Lemma \ref{P3},%
\[
\Lambda\left(  \mathcal{B}_{j-1}\right)  =\sum_{\mathcal{D}_{j-1}}\Pr\left\{
\mathcal{D}_{j-1}|\mathcal{B}_{j-1}\right\}  \boldsymbol{I}_{\mathcal{B}%
_{j-1}\cup\mathcal{D}_{j-1}}=\sum_{\left\{  \mathcal{D}_{j},\mathcal{C}%
_{j}\right\}  }\Phi\left(  \mathcal{C}_{j}\right)  \boldsymbol{I}%
_{\mathcal{C}_{j}\cup\mathcal{H}_{\mathcal{C}_{j}}}\Pr\left\{  \mathcal{D}%
_{j}|\mathcal{B}_{j}\right\}  \boldsymbol{I}_{\mathcal{D}_{j}\cup
\mathcal{B}_{j}}.
\]
Because the scope of $\phi\left(  \mathcal{C}_{j}\right)  $ (which is
$\left\{  \mathcal{C}_{j},\mathcal{H}_{\mathcal{C}_{j}}\right\}
\subseteq\left\{  \mathcal{C}_{j},\mathcal{P}_{j-1}\right\}  =\mathcal{P}_{j}%
$) is not in $\mathcal{D}_{j}$, from Equation (\ref{44}),
\[
\Lambda\left(  \mathcal{B}_{j-1}\right)  =\sum_{\mathcal{C}_{j}}\phi\left(
\mathcal{C}_{j}\right)  \sum_{\mathcal{D}_{j}}\Pr\left\{  \mathcal{D}%
_{j}|\mathcal{B}_{j}\right\}  \boldsymbol{I}_{\mathcal{D}_{j}\cup
\mathcal{B}_{j}}=\sum_{\mathcal{C}_{j}}\phi\left(  \mathcal{C}_{j}\right)
\Lambda\left(  \mathcal{B}_{j}\right)  .
\]

\end{proof}

Suppose there is a value of $\omega$ ($1<\omega\leq\gamma$) such that
$\Lambda\left(  \mathcal{B}_{\omega}\right)  $ can be calculated. Especially,
let $\beta$ be the last time an evidence variable is recruited into
$\mathcal{P}$. Then for all $\beta\leq j\leq\gamma$, $\mathcal{D}_{j}$ has no
evidence. Hence
\begin{equation}
\Lambda\left(  \mathcal{B}_{j}\right)  =\boldsymbol{I}_{\mathcal{B}_{j}}\text{
for all }\beta\leq j\leq\gamma. \label{32}%
\end{equation}
Starting with $\Lambda\left(  \mathcal{B}_{\omega}\right)  $, we can calculate
$\Lambda\left(  \mathcal{B}_{j}\right)  $ for all $0\leq j<\omega$ recursively
by the above lemma.

For the BN $\mathbb{A}$, with $\mathcal{E}=\left\{  H=h,K=k\right\}  $. Thus
$\beta=6$ and $\Lambda\left(  \mathcal{B}_{8}\right)  =\Lambda\left(
\mathcal{B}_{7}\right)  =\Lambda\left(  \mathcal{B}_{6}\right)
=\boldsymbol{I}_{K}$.

\begin{enumerate}
\item Because $\mathcal{B}_{6}$ has cohort $\mathcal{C}_{6}=\left\{
J,K,G\right\}  $:%
\begin{align*}
\Lambda\left(  \mathcal{B}_{5}\right)   &  =\sum_{\mathcal{C}_{6}}\phi\left(
\mathcal{C}_{6}\right)  \Lambda\left(  \mathcal{B}_{6}\right) \\
&  =\Pr\left\{  k|h,I\right\}  \sum_{G}\left(  \Pr\left\{  G\right\}  \sum
_{J}\Pr\left\{  J|G,h\right\}  \right)  \boldsymbol{I}_{K}=\Pr\left\{
k|I,h\right\}  .
\end{align*}

\item Because $\mathcal{B}_{5}$ has cohort $\mathcal{C}_{5}=\varnothing$:%
\[
\Lambda\left(  \mathcal{B}_{4}\right)  =\Lambda\left(  \mathcal{B}_{5}\right)
=\Pr\left\{  k|I,h\right\}  .
\]

\item Because $\mathcal{B}_{4}$ has cohort $\mathcal{C}_{4}=I$:%
\begin{align*}
\Lambda\left(  \mathcal{B}_{3}\right)   &  =\sum_{\mathcal{C}_{4}}\phi\left(
\mathcal{C}_{4}\right)  \Lambda\left(  \mathcal{B}_{4}\right)  =\sum_{I}%
\Pr\left\{  I|D,F\right\}  \Pr\left\{  k|I,h\right\} \\
&  =\sum_{I}\Pr\left\{  k,I|D,F,h\right\}  =\Pr\left\{  k|D,F,h\right\}  .
\end{align*}

\item Because $\mathcal{B}_{3}$ has cohort $\mathcal{C}_{3}=H=h$:%
\[
\Lambda\left(  \mathcal{B}_{2}\right)  =\phi\left(  \mathcal{C}_{3}\right)
\Lambda\left(  \mathcal{B}_{3}\right)  =\Pr\left\{  h|C,D\right\}  \Pr\left\{
k|D,F,h\right\}  =\Pr\left\{  h,k|C,D,F\right\}  .
\]

\item Because $\mathcal{B}_{2}$ has cohort $\mathcal{C}_{2}=\varnothing$:%
\[
\Lambda\left(  \mathcal{B}_{1}\right)  =\Lambda\left(  \mathcal{B}_{2}\right)
=\Pr\left\{  h,k|C,D,F\right\}  .
\]

\item Because $\mathcal{B}_{1}$ has cohort $\mathcal{C}_{1}=\left\{
C,D,F\right\}  $:%
\begin{align*}
\Lambda\left(  \mathcal{B}_{0}\right)   &  =\sum_{\mathcal{C}_{1}}\phi\left(
\mathcal{C}_{1}\right)  \Lambda\left(  \mathcal{B}_{1}\right) \\
&  =\sum_{\left\{  C,D,F\right\}  }\Pr\left\{  C|A,B\right\}  \Pr\left\{
D|A,B\right\}  \Pr\left\{  F|A,B\right\}  \Pr\left\{  h,k|C,D,F\right\} \\
&  =\Pr\left\{  h,k|A,B\right\}  .
\end{align*}

\end{enumerate}

\subsection{The Posterior Marginal Distributions}

Combining the downward and upward passes yields:

\begin{theorem}
\label{FB}For all $0\leq j\leq\gamma$,%
\[
\Pr\left\{  \mathcal{B}_{j},\left[  \mathcal{E}\backslash\mathcal{B}%
_{j}\right]  \right\}  \boldsymbol{I}_{\mathcal{B}_{j}}=\Pi\left(
\mathcal{B}_{j}\right)  \Lambda\left(  \mathcal{B}_{j}\right)  .
\]

\end{theorem}

\begin{proof}
By Lemma \ref{E2}, because the event $\left[  \mathcal{E}\backslash
\mathcal{B}_{j}\right]  $ is the same as the event $\left[  \mathcal{V}%
\backslash\mathcal{B}_{j}\right]  =\left[  \mathcal{A}_{j}\cup\mathcal{D}%
_{j}\right]  $,%
\begin{align*}
\Pr\left\{  \mathcal{B}_{j},\left[  \mathcal{E}\backslash\mathcal{B}%
_{j}\right]  \right\}  \boldsymbol{I}_{\mathcal{B}_{j}}  &  =\Pr\left\{
\mathcal{B}_{j},\left[  \mathcal{A}_{j}\cup\mathcal{D}_{j}\right]  \right\}
\boldsymbol{I}_{\mathcal{B}_{j}}=\sum_{\left\{  \mathcal{A}_{j},\mathcal{D}%
_{j}\right\}  }\Pr\left\{  \mathcal{B}_{j},\mathcal{A}_{j},\mathcal{D}%
_{j}\right\}  \boldsymbol{I}_{\mathcal{B}_{j}\cup\mathcal{A}_{j}%
\cup\mathcal{D}_{j}}\\
&  =\sum_{\left\{  \mathcal{A}_{j},\mathcal{D}_{j}\right\}  }\Pr\left\{
\mathcal{A}_{j},\mathcal{B}_{j}\right\}  \Pr\left\{  \mathcal{D}%
_{j}|\mathcal{B}_{j}\right\}  \boldsymbol{I}_{\mathcal{B}_{j}\cup
\mathcal{A}_{j}\cup\mathcal{D}_{j}}.
\end{align*}
As $\mathcal{D}_{j}\cap\left\{  \mathcal{A}_{j},\mathcal{B}_{j}\right\}
=\varnothing$,
\begin{align*}
\Pr\left\{  \mathcal{B}_{j},\left[  \mathcal{E}\backslash\mathcal{B}%
_{j}\right]  \right\}  \boldsymbol{I}_{\mathcal{B}_{j}}  &  =\sum
_{\mathcal{A}_{j}}\Pr\left\{  \mathcal{A}_{j},\mathcal{B}_{j}\right\}
\boldsymbol{I}_{\mathcal{A}_{j}\cup\mathcal{B}_{j}}\left(  \sum_{\mathcal{D}%
_{j}}\Pr\left\{  \mathcal{D}_{j}|\mathcal{B}_{j}\right\}  \boldsymbol{I}%
_{\mathcal{D}_{j}\cup\mathcal{B}_{j}}\right) \\
&  =\sum_{\mathcal{A}_{j}}\Pr\left\{  \mathcal{A}_{j},\mathcal{B}_{j}\right\}
\boldsymbol{I}_{\mathcal{A}_{j}\cup\mathcal{B}_{j}}\Lambda\left(
\mathcal{B}_{j}\right)  .
\end{align*}
As $\mathcal{B}_{j}\cap\mathcal{A}_{j}=\varnothing$,%
\[
\Pr\left\{  \mathcal{B}_{j},\left[  \mathcal{E}\backslash\mathcal{B}%
_{j}\right]  \right\}  \boldsymbol{I}_{\mathcal{B}_{j}}=\Lambda\left(
\mathcal{B}_{j}\right)  \sum_{\mathcal{A}_{j}}\Pr\left\{  \mathcal{A}%
_{j},\mathcal{B}_{j}\right\}  \boldsymbol{I}_{\mathcal{A}_{j}\cup
\mathcal{B}_{j}}=\Pi\left(  \mathcal{B}_{j}\right)  \Lambda\left(
\mathcal{B}_{j}\right)  .
\]

\end{proof}

\begin{corollary}
\label{FF}For node $Q\in\mathcal{B}_{j}$ where $0\leq j\leq\gamma$,%
\[
\Pr\left\{  Q,\left[  \mathcal{E}\backslash Q\right]  \right\}  \boldsymbol{I}%
_{Q}=\sum_{\mathcal{B}_{j}\backslash Q}\Pi\left(  \mathcal{B}_{j}\right)
\Lambda\left(  \mathcal{B}_{j}\right)  .
\]
If $Q\notin\mathcal{E},$%
\[
\Pr\left\{  Q,\left[  \mathcal{E}\right]  \right\}  =\sum_{\mathcal{B}%
_{j}\backslash Q}\Pi\left(  \mathcal{B}_{j}\right)  \Lambda\left(
\mathcal{B}_{j}\right)  .
\]

\end{corollary}

\begin{proof}
For node $Q\in\mathcal{B}_{j}$,%
\begin{align*}
\Pr\left\{  Q,\left[  \mathcal{E}\backslash Q\right]  \right\}  \boldsymbol{I}%
_{Q}  &  =\sum_{\mathcal{E}\backslash Q}\Pr\left\{  \mathcal{E}\right\}
\boldsymbol{I}_{\mathcal{E}}=\sum_{\mathcal{B}_{j}\backslash Q}\boldsymbol{I}%
_{\mathcal{B}_{j}}\sum_{\mathcal{E}\backslash\mathcal{B}_{j}}\Pr\left\{
\mathcal{B}_{j},\mathcal{E}\backslash\mathcal{B}_{j}\right\}  \boldsymbol{I}%
_{\mathcal{E}\backslash\mathcal{B}_{j}}\\
&  =\sum_{\mathcal{B}_{j}\backslash Q}\boldsymbol{I}_{\mathcal{B}_{j}}%
\Pr\left\{  \mathcal{B}_{j},\left[  \mathcal{E}\backslash\mathcal{B}%
_{j}\right]  \right\}  =\sum_{\mathcal{B}_{j}\backslash Q}\Pi\left(
\mathcal{B}_{j}\right)  \Lambda\left(  \mathcal{B}_{j}\right)  .
\end{align*}

\end{proof}

Recall that $\beta$ is the last time an evidence is recruited into
$\mathcal{P}$, if we are looking for the \textquotedblleft
post-evidence\textquotedblright\ $\Pr\left\{  Q,\left[  \mathcal{E}\backslash
Q\right]  \right\}  \boldsymbol{I}_{Q}$ where $Q\in\mathcal{B}_{j}$ and
$\beta\leq j\leq\gamma$, then due to Equation (\ref{32}), we can find them by
the downward pass alone as $\sum_{\mathcal{B}_{j}\backslash Q}\Pi\left(
\mathcal{B}_{j}\right)  \boldsymbol{I}_{\mathcal{B}_{j}}=\sum_{\mathcal{B}%
_{j}\backslash Q}\Pi\left(  \mathcal{B}_{j}\right)  $.

For the BN $\mathbb{A}$ with $\mathcal{E}=\left\{  H=h,K=k\right\}  $, by the
downward pass alone we already have%
\begin{align*}
\Pr\left\{  \mathcal{B}_{8},\left[  \mathcal{E}\backslash\mathcal{B}%
_{8}\right]  \right\}  \boldsymbol{I}_{\mathcal{B}_{8}}  &  =\Pi\left(
\mathcal{B}_{8}\right)  =\Pr\left\{  J,L,k,h\right\}  ;\\
\Pr\left\{  \mathcal{B}_{7},\left[  \mathcal{E}\backslash\mathcal{B}%
_{7}\right]  \right\}  \boldsymbol{I}_{\mathcal{B}_{7}}  &  =\Pi\left(
\mathcal{B}_{7}\right)  =\Pr\left\{  I,J,k,h\right\}  ;\\
\Pr\left\{  \mathcal{B}_{6},\left[  \mathcal{E}\backslash\mathcal{B}%
_{6}\right]  \right\}  \boldsymbol{I}_{\mathcal{B}_{6}}  &  =\Pi\left(
\mathcal{B}_{6}\right)  =\Pr\left\{  I,J,G,k,h\right\}  .
\end{align*}
Now with Theorem \ref{FB},

\begin{enumerate}
\item $\Pi\left(  \mathcal{B}_{5}\right)  \Lambda\left(  \mathcal{B}%
_{5}\right)  =\Pr\left\{  I,h\right\}  \Pr\left\{  k|I,h\right\}  =\Pr\left\{
I,h,k\right\}  .$

\item $\Pi\left(  \mathcal{B}_{4}\right)  \Lambda\left(  \mathcal{B}%
_{4}\right)  =\Pr\left\{  F,I,h\right\}  \Pr\left\{  k|I,h\right\}
=\Pr\left\{  F,I,h,k\right\}  .$

\item $\Pi\left(  \mathcal{B}_{3}\right)  \Lambda\left(  \mathcal{B}%
_{3}\right)  =\Pr\left\{  D,F,h\right\}  \Pr\left\{  k|D,F,h\right\}
=\Pr\left\{  D,F,h,k\right\}  .$

\item $\Pi\left(  \mathcal{B}_{2}\right)  \Lambda\left(  \mathcal{B}%
_{2}\right)  =\Pr\left\{  C,D,F\right\}  \Pr\left\{  h,k|C,D,F\right\}
=\Pr\left\{  C,D,F,h,k\right\}  .$

\item $\Pi\left(  \mathcal{B}_{1}\right)  \Lambda\left(  \mathcal{B}%
_{1}\right)  =\Pr\left\{  B,C,D,F\right\}  \Pr\left\{  h,k|C,D,F\right\}
=\Pr\left\{  B,C,D,F,h,k\right\}  .$

\item $\Pi\left(  \mathcal{B}_{0}\right)  \Lambda\left(  \mathcal{B}%
_{0}\right)  =\Pr\left\{  A,B\right\}  \Pr\left\{  h,k|A,B\right\}
=\Pr\left\{  A,B,h,k\right\}  .$
\end{enumerate}

A variable $V\in\mathcal{V}$ may appear in more than one borders. For example,
variable $I$ appears in $\mathcal{B}_{4}$, $\mathcal{B}_{5}$, $\mathcal{B}%
_{6}$ and $\mathcal{B}_{7}$. We obtain the same result regardless which of
these borders we choose to marginalize.

Border algorithm is applicable to all BNs. In the next section, we study a
special kind of BNs called the polytrees,\ and\ present in details, with some
modifications and within the border algorithm framework, the \textquotedblleft
revised polytree algorithm\textquotedblright\ by Peot \& Shachter (1991). This
is an important section, because we will show later that, with the help of the
border algorithm, any BN can be modified to become a polytree.

\section{The Revised Polytree Algorithm}

A polytree is a BN which is \textquotedblleft singly
connected;\textquotedblright\ that is, there is only one \emph{undirected}
path connecting any two nodes. (From now on, \textquotedblleft
path\textquotedblright\ means \textquotedblleft undirected
path.\textquotedblright) The BN $\mathbb{A}$\ in Figure \ref{F1} is not a
polytree because there are 2 paths from $A$ to $H$, namely $A-C-H$ and $A-D-H$.

In other words, while we assume all BNs are acyclic (that is, they have no
directed cycles), a polytree also does not have any undirected cycle (or
\textquotedblleft loop\textquotedblright). The BN $\mathbb{A}$\ has loop
$A-C-H-D-A$.

\begin{figure}
[ptb]
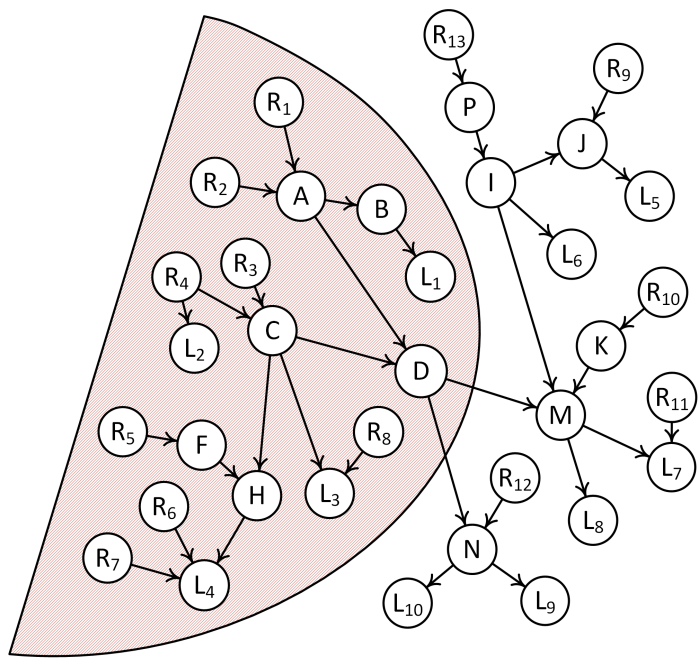

\begin{center}
\includegraphics[
natheight=3.400200in,
natwidth=3.610500in,
height=3.4002in,
width=3.6105in
]%
4%
\caption{ The Polytree $\mathbb{B}$}%
\label{F6}%
\end{center}
\end{figure}

As an illustration, we will use the polytree\ as shown in Figure \ref{F6},
which we will refer to as the Polytree $\mathbb{B}$.%

\subsection{To Find the Path Connecting Two Nodes}

Here we present a method to identify the unique path connecting any two nodes
in a polytree.

We strategically designate some nodes as \textquotedblleft
hubs,\textquotedblright\ and pre-load the unique path connecting each pair of
hubs, excluding the hubs themselves. The bigger the network, the more hubs we
need. For the Polytree $\mathbb{B}$, let us pick nodes $J$\ and $H$ as hubs,
connected by path $I-M-D-C$.

For each node, we also pre-load the path from it to its nearest hub. For node
$P$ the path is $P-I-J$; for node $A$ the path is $A-D-C-H$.

To find the path from node $X$ to node $Y$:

\begin{enumerate}
\item Form the (possibly cyclic)\ path from $X$ to the hub nearest to $X$,
then to the hub nearest to $Y$, then to $Y$. For nodes $P$ and $A$, this path
is $\left(  P-I-J\right)  -\left(  I-M-D-C\right)  -\left(  H-C-D-A\right)  $.

\item Replace the largest loop around each hub with its furthest node. With
the above path, replace loop $I-J-I$ with node $I$, and loop $D-C-H-C-D$ with
node $D$, resulting in path $P-I-M-D-A$ connecting nodes $P$ and $A$.
\end{enumerate}

\subsection{The Decompositions by Nodes}

Node $V$ in a polytree decomposes the polytree into two parts:

\begin{enumerate}
\item $\mathcal{P}_{V}$, the parentless set including $V$\ and all the nodes
that are connected to $V$ \textquotedblleft from above,\textquotedblright\ via
its parents $\mathcal{H}_{V}$. (In Figure \ref{F6}, $\mathcal{P}_{D}$ is in
the shaded region.) $\mathcal{P}_{V}$ has border $V$ and the top part
$\mathcal{A}_{V}=\mathcal{P}_{V}\backslash V$. Consistent with Definition
(\ref{49}), we define%
\begin{equation}
\Pi\left(  V\right)  =\Pr\left\{  V,\left[  \mathcal{A}_{V}\right]  \right\}
\boldsymbol{I}_{V}=\sum_{\mathcal{A}_{V}}\Pr\left\{  \mathcal{P}_{V}\right\}
\boldsymbol{I}_{\mathcal{P}_{V}}. \label{14}%
\end{equation}

\item $\mathcal{D}_{V}$, the bottom set of $\mathcal{P}_{V}$, in which all
nodes are connected to $V$ \textquotedblleft from below,\textquotedblright%
\ via its children $\mathcal{L}_{V}$. Consistent with Definition (\ref{55}),
we define%
\begin{equation}
\Lambda\left(  V\right)  =\Pr\left\{  \left[  \mathcal{D}_{V}\right]
|V\right\}  \boldsymbol{I}_{V}=\sum_{\mathcal{D}_{V}}\Pr\left\{
\mathcal{D}_{V}|V\right\}  \boldsymbol{I}_{V\cup\mathcal{D}_{V}}. \label{13}%
\end{equation}

\end{enumerate}

\subsection{The Decompositions by Edges}

So far, we focused on the nodes in a BN. Let us now consider a typical edge
$X\rightarrow Y$. While node $X$\ decomposes the polytree into $\mathcal{P}%
_{X}$ and $\mathcal{D}_{X}$, edge $X\rightarrow Y$ also decomposes it into two parts:

\begin{enumerate}
\item $\mathcal{T}_{X\rightarrow Y}$, the parentless set of nodes on the
parent side of edge $X\rightarrow Y$, having border $X$. Not only does
$\mathcal{T}_{X\rightarrow Y}$ include $\mathcal{P}_{X}$, but also all the
nodes that connect to $X$ from below, except those via $Y$. In the Polytree
$\mathbb{B}$, in addition to $\mathcal{P}_{D}$, $\mathcal{T}_{D\rightarrow M}$
also includes $\left\{  N,R_{12},L_{9},L_{10}\right\}  $.

Consistent with Definition (\ref{49}), we define the \emph{downward
message}\ about the evidences in $\mathcal{T}_{X\rightarrow Y}$ that node $X$
can send to its child $Y$ as%
\begin{equation}
\Pi_{Y}\left(  X\right)  =\Pr\left\{  X,\left[  \mathcal{T}_{X\rightarrow
Y}\backslash X\right]  \right\}  \boldsymbol{I}_{X}=\sum_{\mathcal{T}%
_{X\rightarrow Y}\backslash X}\Pr\left\{  \mathcal{T}_{X\rightarrow
Y}\right\}  \boldsymbol{I}_{\mathcal{T}_{X\rightarrow Y}}. \label{78}%
\end{equation}

\item The bottom set $\mathcal{U}_{X\rightarrow Y}=\mathcal{V}\backslash
\mathcal{T}_{X\rightarrow Y}$, on the child side of edge $X\rightarrow Y$,
separated from $\mathcal{T}_{X\rightarrow Y}\backslash X$ by $X$. Not only
does $\mathcal{U}_{X\rightarrow Y}$ include $\left\{  Y,\mathcal{D}%
_{Y}\right\}  $, but also all the nodes that connect to $Y$ from above, except
those via $X$. Hence,%
\begin{equation}
\mathcal{U}_{X\rightarrow Y}=\left\{  Y,\mathcal{D}_{Y},\cup_{V\in
\mathcal{H}_{Y}\backslash X}\mathcal{T}_{V\rightarrow Y}\right\}  . \label{79}%
\end{equation}
On the other hand,%
\begin{equation}
\mathcal{T}_{X\rightarrow Y}=\left\{  \mathcal{P}_{X},\cup_{V\in
\mathcal{L}_{X}\backslash Y}\mathcal{U}_{X\rightarrow V}\right\}  . \label{80}%
\end{equation}

Consistent with Definition (\ref{55}), we define the \emph{upward
message}\ about the evidences in $\mathcal{U}_{X\rightarrow Y}$ that node $Y$
can send to its parent $X$ as%
\begin{equation}
\Lambda_{Y}\left(  X\right)  =\Pr\left\{  \left[  \mathcal{U}_{X\rightarrow
Y}\right]  |X\right\}  \boldsymbol{I}_{X}=\sum_{\mathcal{U}_{X\rightarrow Y}%
}\Pr\left\{  \mathcal{U}_{X\rightarrow Y}|X\right\}  \boldsymbol{I}%
_{X\cup\mathcal{U}_{X\rightarrow Y}}. \label{77}%
\end{equation}

\end{enumerate}

We will often use the following two properties related to edge $X\rightarrow
Y$ in a polytree:

\begin{enumerate}
\item Because two distinct parents $Z$ and $T$ of $X$ have $X$ as a common
child, the two parentless sets $\mathcal{T}_{Z\rightarrow X}$ and
$\mathcal{T}_{T\rightarrow X}$ must be disjoint and independent (otherwise,
there are two paths from their common member to $Y$, via the two parents).
Thus,%
\[
\Pr\left\{  \cup_{V\in\mathcal{H}_{X}}\mathcal{T}_{V\rightarrow X}\right\}
=\prod\limits_{V\in\mathcal{H}_{X}}\Pr\left\{  \mathcal{T}_{V\rightarrow
X}\right\}  .
\]

\item Because two distinct children $Y$ and $W$ of $X$ have $X$ as a common
parent, the two sets $\mathcal{U}_{X\rightarrow Y}$ and $\mathcal{U}%
_{X\rightarrow W}$ must be disjoint and independent given $X$ (otherwise,
there are two paths from $X$ to their common member, via the two children).
Thus,%
\[
\Pr\left\{  \cup_{V\in\mathcal{L}_{X}}\mathcal{U}_{X\rightarrow V}|X\right\}
=\prod\limits_{V\in\mathcal{L}_{X}}\Pr\left\{  \mathcal{U}_{X\rightarrow
V}|X\right\}  .
\]

\end{enumerate}

\subsection{The Message Propagations in the Polytrees}

We now present the lemmas about the relationships among $\Pi\left(  X\right)
$, $\Lambda\left(  X\right)  $, $\Pi_{Y}\left(  X\right)  $ and $\Lambda
_{Y}\left(  X\right)  $, which are known in the literature, but are now proven
within the border algorithm framework:

\begin{lemma}
\label{P7}For edge $X\rightarrow Y$ in a polytree,%
\[
\Pi_{Y}\left(  X\right)  =\Pi\left(  X\right)  \prod\limits_{V\in
\mathcal{L}_{X}\backslash Y}\Lambda_{V}\left(  X\right)  .
\]

\end{lemma}

\begin{proof}
Consider the parentless $\mathcal{P}_{X}$, having border $X$. As in Equation
(\ref{80}), recruiting the childless $\cup_{V\in\mathcal{L}_{X}\backslash
Y}\mathcal{U}_{X\rightarrow V}$ without promotion results in the parentless
$\mathcal{T}_{X\rightarrow Y}$ with border $X\cup_{V\in\mathcal{L}%
_{X}\backslash Y}\mathcal{U}_{X\rightarrow V}$. From Theorem \ref{FE},%
\begin{align*}
\Pi\left(  X\cup_{V\in\mathcal{L}_{X}\backslash Y}\mathcal{U}_{X\rightarrow
V}\right)   &  =\Pi\left(  X\right)  \Pr\left\{  \cup_{V\in\mathcal{L}%
_{X}\backslash Y}\mathcal{U}_{X\rightarrow V}|X\right\}  \boldsymbol{I}%
_{X\cup_{V\in\mathcal{L}_{X}\backslash Y}\mathcal{U}_{X\rightarrow V}}\\
&  =\Pi\left(  X\right)  \prod\limits_{V\in\mathcal{L}_{X}\backslash Y}%
\Pr\left\{  \mathcal{U}_{X\rightarrow V}|X\right\}  \boldsymbol{I}%
_{X\cup\mathcal{U}_{X\rightarrow V}}.
\end{align*}

Now use Rule 1 to promote the childless $\cup_{V\in\mathcal{L}_{X}\backslash
Y}\mathcal{U}_{X\rightarrow V}$ without cohort, resulting in the parentless
$\mathcal{T}_{X\rightarrow Y}$ having border $X$. From Theorem \ref{FE},%
\begin{align*}
\Pi_{Y}\left(  X\right)   &  =\sum_{\cup_{V\in\mathcal{L}_{X}\backslash
Y}\mathcal{U}_{X\rightarrow V}}\Pi\left(  X\cup_{V\in\mathcal{L}_{X}\backslash
Y}\mathcal{U}_{X\rightarrow V}\right) \\
&  =\sum_{\cup_{V\in\mathcal{L}_{X}\backslash Y}\mathcal{U}_{X\rightarrow V}%
}\Pi\left(  X\right)  \prod\limits_{V\in\mathcal{L}_{X}\backslash Y}%
\Pr\left\{  \mathcal{U}_{X\rightarrow V}|X\right\}  \boldsymbol{I}%
_{X\cup\mathcal{U}_{X\rightarrow V}}.
\end{align*}

As the scope of $\Pi\left(  X\right)  $ (which is $X$) is not in $\cup
_{V\in\mathcal{L}_{X}\backslash Y}\mathcal{U}_{X\rightarrow V}$,%
\[
\Pi_{Y}\left(  X\right)  =\Pi\left(  X\right)  \prod\limits_{V\in
\mathcal{L}_{X}\backslash Y}\left(  \sum_{\mathcal{U}_{X\rightarrow V}}%
\Pr\left\{  \mathcal{U}_{X\rightarrow V}|X\right\}  \boldsymbol{I}%
_{X\cup\mathcal{U}_{X\rightarrow V}}\right)  ,
\]
which is the lemma by Definition (\ref{77}). This is similar to Equation
(4.45) in Pearl, 1988.
\end{proof}

\begin{lemma}
\label{P5}For edge $X\rightarrow Y$ in a polytree,%
\[
\Lambda_{Y}\left(  X\right)  =\sum_{Y}\Lambda\left(  Y\right)  \sum
_{\mathcal{H}_{Y}\backslash X}\Pr\nolimits_{r}\left\{  Y|\mathcal{H}%
_{Y}\right\}  \prod\limits_{V\in\mathcal{H}_{Y}\backslash X}\Pi_{Y}\left(
V\right)  .
\]

\end{lemma}

\begin{proof}
Consider the parentless $\mathcal{T}_{X\rightarrow Y}$, having border $X$ and
the bottom set $\mathcal{U}_{X\rightarrow Y}$. As shown in Equation
(\ref{79}), recruiting $Y\cup_{V\in\mathcal{H}_{Y}\backslash X}\mathcal{T}%
_{V\rightarrow Y}$ without promotion results in the parentless $\mathcal{P}%
_{Y}$ with bottom part $\mathcal{D}_{Y}$. The reduced cohort table is%
\begin{align*}
&  \Pr\left\{  Y\cup_{V\in\mathcal{H}_{Y}\backslash X}\mathcal{T}%
_{V\rightarrow Y}|X\right\}  \boldsymbol{I}_{X\cup Y\cup_{V\in\mathcal{H}%
_{Y}\backslash X}\mathcal{T}_{V\rightarrow Y}}\\
&  =\Pr\left\{  Y|\cup_{V\in\mathcal{H}_{Y}\backslash X}\mathcal{T}%
_{V\rightarrow Y},X\right\}  \Pr\left\{  \cup_{V\in\mathcal{H}_{Y}\backslash
X}\mathcal{T}_{V\rightarrow Y}|X\right\}  \boldsymbol{I}_{X\cup Y\cup
_{V\in\mathcal{H}_{Y}\backslash X}\mathcal{T}_{V\rightarrow Y}}%
\end{align*}

Because $\mathcal{H}_{Y}\subseteq X\cup_{V\in\mathcal{H}_{Y}\backslash
X}\mathcal{T}_{V\rightarrow Y}$, the reduced cohort table becomes%
\begin{align*}
&  \Pr\left\{  Y|\mathcal{H}_{Y}\right\}  \boldsymbol{I}_{Y\cup\mathcal{H}%
_{Y}}\Pr\left\{  \cup_{V\in\mathcal{H}_{Y}\backslash X}\mathcal{T}%
_{V\rightarrow Y}\right\}  \boldsymbol{I}_{\cup_{V\in\mathcal{H}_{Y}\backslash
X}\mathcal{T}_{V\rightarrow Y}}\\
&  =\Pr\nolimits_{r}\left\{  Y|\mathcal{H}_{Y}\right\}  \prod\limits_{V\in
\mathcal{H}_{Y}\backslash X}\Pr\left\{  \mathcal{T}_{V\rightarrow Y}\right\}
\boldsymbol{I}_{\mathcal{T}_{V\rightarrow Y}}.
\end{align*}
From Theorem \ref{B}, as $Y\notin\cup_{V\in\mathcal{H}_{Y}\backslash
X}\mathcal{T}_{V\rightarrow Y}$,%
\begin{align*}
\Lambda_{Y}\left(  X\right)   &  =\sum_{Y\cup_{V\in\mathcal{H}_{Y}\backslash
X}\mathcal{T}_{V\rightarrow Y}}\left(  \Pr\nolimits_{r}\left\{  Y|\mathcal{H}%
_{Y}\right\}  \prod\limits_{V\in\mathcal{H}_{Y}\backslash X}\Pr\left\{
\mathcal{T}_{V\rightarrow Y}\right\}  \boldsymbol{I}_{\mathcal{T}%
_{V\rightarrow Y}}\right)  \Lambda\left(  Y\right) \\
&  =\sum_{Y}\Lambda\left(  Y\right)  \sum_{\mathcal{H}_{Y}\backslash X}%
\Pr\nolimits_{r}\left\{  Y|\mathcal{H}_{Y}\right\}  \prod\limits_{V\in
\mathcal{H}_{Y}\backslash X}\left(  \sum_{\mathcal{T}_{V\rightarrow
Y}\backslash V}\Pr\left\{  \mathcal{T}_{V\rightarrow Y}\right\}
\boldsymbol{I}_{\mathcal{T}_{V\rightarrow Y}}\right)  ,
\end{align*}
which is the lemma by Definition (\ref{78}). This is similar to Equation
(4.44) in Pearl (1988).
\end{proof}

The values of $\Pi\left(  \cdot\right)  $ and $\Lambda\left(  \cdot\right)  $
can be calculated from the messages as in the following lemmas:

\begin{lemma}
\label{P1}For node $X$ in a polytree,
\[
\Pi\left(  X\right)  =\sum_{\mathcal{H}_{X}}\Pr\nolimits_{r}\left\{
X|\mathcal{H}_{X}\right\}  \prod\limits_{V\in\mathcal{H}_{X}}\Pi_{X}\left(
V\right)  .
\]

\end{lemma}

\begin{proof}
Consider the parentless $\cup_{V\in\mathcal{H}_{X}}\mathcal{T}_{V\rightarrow
X}$ having border $\mathcal{H}_{X}$ and the top part $\cup_{V\in
\mathcal{H}_{X}}\left\{  \mathcal{T}_{V\rightarrow X}\backslash V\right\}  $.
By Definition (\ref{49}),
\begin{align*}
\Pi\left(  \mathcal{H}_{X}\right)   &  =\sum_{\cup_{V\in\mathcal{H}_{X}%
}\left\{  \mathcal{T}_{V\rightarrow X}\backslash V\right\}  }\Pr\left\{
\cup_{V\in\mathcal{H}_{X}}\mathcal{T}_{V\rightarrow X}\right\}  \boldsymbol{I}%
_{\cup_{V\in\mathcal{H}_{X}}\mathcal{T}_{V\rightarrow X}}\\
&  =\prod\limits_{V\in\mathcal{H}_{X}}\left(  \sum_{\mathcal{T}_{V\rightarrow
X}\backslash V}\Pr\left\{  \mathcal{T}_{V\rightarrow X}\right\}
\boldsymbol{I}_{\mathcal{T}_{V\rightarrow X}}\right)  =\prod\limits_{V\in
\mathcal{H}_{X}}\Pi_{X}\left(  V\right)  .
\end{align*}
Now recruit $X$, resulting in the parentless $\mathcal{P}_{X}=X\cup
_{V\in\mathcal{H}_{X}}\mathcal{T}_{V\rightarrow X}$. Then use Rule 1 to
promote $\mathcal{H}_{X}$ without cohort, leaving $X$ as the border of
$\mathcal{P}_{X}$. From Theorem \ref{FE},%
\[
\Pi\left(  X\right)  =\sum_{\mathcal{H}_{X}}\Pr\nolimits_{r}\left\{
X|\mathcal{H}_{X}\right\}  \Pi\left(  \mathcal{H}_{X}\right)  =\sum
_{\mathcal{H}_{X}}\Pr\nolimits_{r}\left\{  X|\mathcal{H}_{X}\right\}
\prod\limits_{V\in\mathcal{H}_{X}}\Pi_{X}\left(  V\right)  ,
\]
hence the lemma, which is similar to Equation (4.38) in Pearl (1988).
\end{proof}

\begin{lemma}
\label{P2}For node $X$ in a polytree,%
\[
\Lambda\left(  X\right)  =\prod\limits_{V\in\mathcal{L}_{X}}\Lambda_{V}\left(
X\right)  .
\]

\end{lemma}

\begin{proof}
Consider the parentless $\mathcal{P}_{X}$, having border $X$. Recruit the rest
of the network $\mathcal{D}_{X}=\cup_{V\in\mathcal{L}_{X}}\mathcal{U}%
_{X\rightarrow V}$ without promotion, resulting in the bottom part
$\varnothing$ with $\Lambda\left(  \varnothing\right)  =1$. From Theorem
\ref{B},
\begin{align*}
\Lambda\left(  X\right)   &  =\sum_{\cup_{V\in\mathcal{L}_{X}}\mathcal{U}%
_{X\rightarrow V}}\Pr\left\{  \cup_{V\in\mathcal{L}_{X}}\mathcal{U}%
_{X\rightarrow V}|X\right\}  \boldsymbol{I}_{X\cup_{V\in\mathcal{L}_{X}%
}\mathcal{U}_{X\rightarrow V}}\Lambda\left(  \varnothing\right) \\
&  =\prod\limits_{V\in\mathcal{L}_{X}}\sum_{\mathcal{U}_{X\rightarrow V}}%
\Pr\left\{  \mathcal{U}_{X\rightarrow V}|X\right\}  \boldsymbol{I}%
_{X\cup\mathcal{U}_{X\rightarrow V}},
\end{align*}
hence the lemma by Definition (\ref{77}), which is similar to Equation (4.35)
in Pearl (1988).
\end{proof}

Our dealing with the evidences here is slightly different to that in Peot \&
Shachter (1991, pp.~308-309). While we attach the indicator column
$\boldsymbol{I}_{V}$ to all $\Pi\left(  V\right)  $, $\Pi_{Y}\left(  V\right)
$, $\Lambda\left(  V\right)  $ and $\Lambda_{Y}\left(  V\right)  $, they call
it the \textquotedblleft local evidence\textquotedblright\ $\Lambda_{V}\left(
V\right)  $ and attach it to $\Lambda\left(  V\right)  $ only.

Combining Lemmas \ref{P7} and \ref{P1} yields:

\begin{theorem}
\label{P20}If node $X$ has received the messages from all members of
$\mathcal{H}_{X}\cup\left\{  \mathcal{L}_{X}\backslash Y\right\}  $, then it
can send a downward message to its child $Y$ as%
\[
\Pi_{Y}\left(  X\right)  =\left(  \prod\limits_{V\in\mathcal{L}_{X}\backslash
Y}\Lambda_{V}\left(  X\right)  \right)  \sum_{\mathcal{H}_{X}}\left(
\Pr\nolimits_{r}\left\{  X|\mathcal{H}_{X}\right\}  \prod\limits_{V\in
\mathcal{H}_{X}}\Pi_{X}\left(  V\right)  \right)  .
\]

\end{theorem}

Combining Lemmas \ref{P5} and \ref{P2} yields:

\begin{theorem}
\label{P21}If node $X$ has received the messages from all members of $\left\{
\mathcal{H}_{X}\backslash H\right\}  \cup\mathcal{L}_{X}$, then it can send an
upward message to is parent $H$ as
\[
\Lambda_{X}\left(  H\right)  =\sum_{X}\left(  \prod\limits_{V\in
\mathcal{L}_{X}}\Lambda_{V}\left(  X\right)  \right)  \sum_{\mathcal{H}%
_{X}\backslash H}\left(  \Pr\nolimits_{r}\left\{  X|\mathcal{H}_{X}\right\}
\prod\limits_{V\in\mathcal{H}_{X}\backslash H}\Pi_{X}\left(  V\right)
\right)  .
\]

\end{theorem}

\subsection{The Collection Phase}

In the first phase of the revised polytree algorithm, which is known as the
\textquotedblleft collection phase,\textquotedblright\ a randomly selected
node $P$ is designated as a \textquotedblleft pivot,\textquotedblright\ and
the messages are passed (or \textquotedblleft propagated\textquotedblright) to
$P$.

\subsubsection{The Evidential Cores}

The \textquotedblleft evidential core\textquotedblright\ (or EC) of a
polytree\ is the smallest sub-polytree which contains all the evidences. In
other words, it comprises of the evidence set $\mathcal{E}$ and all the nodes
and edges on the path connecting every pair of the evidence nodes.
Corresponding to a particular evidence set, the EC is a unique.

\begin{figure}
[ptb]
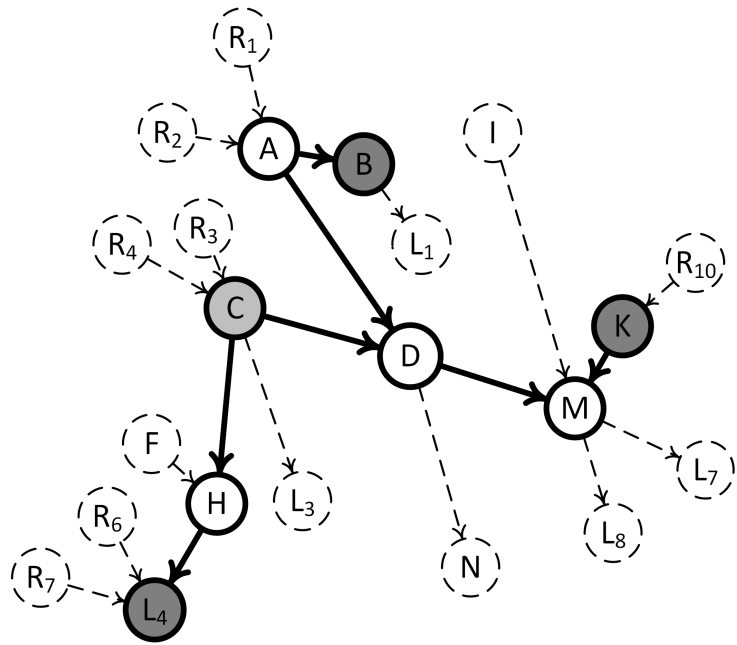

\begin{center}
\includegraphics[
natheight=3.350400in,
natwidth=3.802400in,
height=3.3504in,
width=3.8024in
]%
5%
\caption{The EC $\mathbb{B}$ in the Polytree $\mathbb{B}$, corresponding to
$\mathcal{E}=\left\{  {B,C,K,L}_{{4}}\right\}  $}%
\label{F7}%
\end{center}
\end{figure}

Figure \ref{F7} shows the EC in the Polytree $\mathbb{B}$, corresponding to
$\mathcal{E}=\left\{  B,C,K,L_{4}\right\}  $, not including the nodes or edges
in dash, such as node $L_{1}$. We call it the EC $\mathbb{B}$.%

A node in a polytree is said to be \textquotedblleft inside\textquotedblright%
\ if it is in the EC; otherwise it is \textquotedblleft
outside.\textquotedblright\ Likewise, an edge $X\rightarrow Y$ is
\textquotedblleft inside\textquotedblright\ if both $X$ and $Y$ are inside;
otherwise it is \textquotedblleft outside.\textquotedblright\ A path is
\textquotedblleft inside\textquotedblright\ if all its component edges are.

It is important to observe that, for an outside edge $X\rightarrow Y$, the
whole EC can only be either on its parent side $\mathcal{T}_{X\rightarrow Y}$
, or its child side $\mathcal{U}_{X\rightarrow Y}$, not both. Otherwise, being
connected, the EC must include edge $X\rightarrow Y$ on the path connecting
the evidences on both sides.

\subsubsection{The Boundary Conditions}

We want to use Theorems \ref{P20} and \ref{P21} for calculating the messages
along the inside edges only. However, this requires the knowledge of the
boundary conditions; that is, of the messages from the outside neighbors to
the inside nodes, such as from those in dash in Figure \ref{F7}. So we need
the following theorem:

\begin{theorem}
\label{P10}Consider an inside node $X$ of an EC.

\noindent(a) If $V\in\mathcal{H}_{X}$ is outside, then $\Pi_{X}\left(
V\right)  =\Pr\left\{  V\right\}  $.

\noindent(b) If $V\in\mathcal{L}_{X}$ is outside, then $\Lambda_{V}\left(
X\right)  =\boldsymbol{I}_{X}$.
\end{theorem}

\begin{proof}
(a) Because the inside $X$ is on the child side of edge $V\rightarrow X$, so
is the whole EC, and thus $\mathcal{T}_{V\rightarrow X}$ has no evidence, or
$\boldsymbol{I}_{\mathcal{T}_{V\rightarrow X}}=1$. Hence, by Definition
(\ref{78}), $\Pi_{X}\left(  V\right)  =\Pr\left\{  V\right\}  .$

(b) If $V\in\mathcal{L}_{X}$ is outside, then because the inside $X$ is on the
parent side of edge $X\rightarrow V$, so is the whole EC; hence $\mathcal{U}%
_{X\rightarrow V}$ has no evidence. By Definition (\ref{77}), $\Lambda
_{V}\left(  X\right)  =\boldsymbol{I}_{X}$.
\end{proof}

So, assuming that all prior probabilities $\Pr\left\{  V\right\}  $ are
pre-loaded, Theorems \ref{P20} and \ref{P21} can be used without the need to
calculate the messages from the outside neighbors.

\subsubsection{The Message Initializations}

Like all polytrees, an EC has its own roots and leaves. For example, the sets
of roots and leaves of the EC $\mathbb{B}$ are $\left\{  K,A,C\right\}  $ and
$\left\{  B,L_{4},M\right\}  $, respectively.

By definition of the EC, every path in it must end with an evidence node,
although there are evidence nodes not at the end of a path (such as nodes $C$
in the EC $\mathbb{B}$). Also, an evidence node at the end of a path can only
have one inside child or parent, otherwise the path does not end with it. If
it has one child, then it is a root of the EC (such as nodes $K$\ in the EC
$\mathbb{B}$); if one parent, it is a leaf (such as nodes $B$\ and $L_{4}$ in
the EC $\mathbb{B}$).

We initialize the message propagations in an EC\ as follows:

\begin{theorem}
\label{P19}Consider an EC.

\noindent(a) For an evidence root $R$ having one inside child $C$, $\Pi
_{C}\left(  R\right)  =\Pi\left(  R\right)  =\Pr\left\{  R\right\}
\boldsymbol{I}_{R}$.

\noindent(b) For an evidence leaf $L$, $\Lambda\left(  L\right)
=\boldsymbol{I}_{L}.$
\end{theorem}

\begin{proof}
(a) Because $\mathcal{A}_{R}=\mathcal{P}_{R}\backslash R$ is non-evidential,
$\Pi\left(  R\right)  =\Pr\left\{  R\right\}  \boldsymbol{I}_{R}$ by
Definition (\ref{14}). Also, because $R$ has only one inside child $C$,
$\Lambda_{V}\left(  R\right)  =\boldsymbol{I}_{R}$ for all $V\in
\mathcal{L}_{R}\backslash C$. Thus $\Pi_{C}\left(  R\right)  =\Pi\left(
R\right)  $ from Lemma \ref{P7}.

(b) For an evidence leaf $L$, $\mathcal{D}_{L}$ has no evidence and thus
$\Lambda\left(  L\right)  =\boldsymbol{I}_{L}$ by Definition (\ref{13}).
\end{proof}

\subsubsection{To the Pivot}

In the collection phase, starting with an evidence root or leaf in the EC, we
send the messages from all the inside nodes to a randomly selected inside
pivot node $P$. At any time, if $P$ has not received a message along an edge
to or from it, we trace along the possibly many inside paths leading to this
edge, and must see either an inside node that is ready to do so, or an
evidence node at the end of an inside path that can send its message to $P$ by
Theorem \ref{P19}.

We may need both Theorems \ref{P20} and \ref{P21} to send the messages along a
path, because the directions may change (for example, along path $B\leftarrow$
$A\rightarrow D$). In the EC $\mathbb{B}$ in Figure \ref{F7}, if we use node
$D$ as the pivot, then the collection phase includes the messages sent along
paths $B\leftarrow A\rightarrow D$, $L_{4}\leftarrow H\leftarrow C\rightarrow
D$ and $K\rightarrow$ $M\leftarrow D$.

\subsection{The Distribution Phase}

In the second phase of the revised polytree algorithm, which is known as the
\textquotedblleft distribution phase,\textquotedblright\ the messages are
passed from the pivot node $P$ to a query variable.

\subsubsection{The Informed Nodes}

Once a node $Q$ has received the messages from all its neighbors, we say it is
\textquotedblleft informed.\textquotedblright\ At the end of the collection
phase, the pivot node is the first informed node.

The posterior marginal probability of an informed node can now be obtained:
With the messages from all its parents, we can use Lemma \ref{P1} to calculate
$\Pi\left(  Q\right)  $; with the messages from all its children, we can use
Lemma \ref{P2} to calculate $\Lambda\left(  Q\right)  $. $\Pr\left\{
Q,\left[  \mathcal{E}\backslash Q\right]  \right\}  \boldsymbol{I}_{Q}$ then
can be calculated by Theorem \ref{FB} as $\Pi\left(  Q\right)  \Lambda\left(
Q\right)  $. Alternatively, we can use the following theorem:

\begin{theorem}
\label{P23}With edge $Q\rightarrow Y$,
\[
\Pr\left\{  Q,\left[  \mathcal{E}\backslash Q\right]  \right\}  \boldsymbol{I}%
_{Q}=\Pi_{Y}\left(  Q\right)  \Lambda_{Y}\left(  Q\right)  .
\]

\end{theorem}

\begin{proof}
By Theorem \ref{FB}, Lemmas \ref{P2} and \ref{P7},%
\begin{align*}
\Pr\left\{  Q,\left[  \mathcal{E}\backslash Q\right]  \right\}  \boldsymbol{I}%
_{Q}  &  =\Pi\left(  Q\right)  \Lambda\left(  Q\right)  =\Pi\left(  Q\right)
\prod\limits_{V\in\mathcal{L}_{Q}}\Lambda_{V}\left(  Q\right) \\
&  =\left(  \Pi\left(  Q\right)  \prod\limits_{V\in\mathcal{L}_{Q}\backslash
Y}\Lambda_{V}\left(  Q\right)  \right)  \Lambda_{Y}\left(  Q\right)  =\Pi
_{Y}\left(  Q\right)  \Lambda_{Y}\left(  Q\right)  .
\end{align*}

\end{proof}

\subsubsection{The Inside Query Variables}

Let $p$ be the number of paths connecting an inside node $V$ with the rest of
the network (that is, the number of its neighbors). If it is not the pivot,
then only one of these paths leads to the pivot node $P$. For us to use
Theorem \ref{P20} or \ref{P21} to send a message from $V$ to $P$ in the
collection phase, $V$ must have received the messages along all $p-1$ paths,
except the one leading to $P$. Now, in the distribution phase, once it
receives a message from the informed $P$, it becomes informed.

Let $\mathcal{J}$ be the set of all the informed nodes, which we call the
\textquotedblleft informed set.\textquotedblright\ When we sent the messages
from an informed node $V$ to an uninformed node $Q$, not only $Q$, but also
all the nodes along the path from $V$ to $Q$ become informed. In the EC
$\mathbb{B}$ in Figure \ref{F7}, the message propagations from the informed
node $D$ to node $H$ also make node $C$ informed. Thus, \textquotedblleft
spreading out\textquotedblright\ from a single pivot node, the informed set
forms a connected sub-polytree, in that there is a single path connecting any
two informed nodes and all nodes along that path are informed.

\subsubsection{The Outside Query Variables}

Starting from an inside pivot node, the informed set $\mathcal{J}$ does not
have to cover the EC and can spread beyond it.

Consider now an outside uninformed\ node $Q$. The paths starting from $Q$\ to
all nodes in $\mathcal{J}$ must go through a unique \textquotedblleft
gate\textquotedblright\ $T\in\mathcal{J}$; otherwise, because $\mathcal{J}$ is
connected, there are more than one paths connecting $Q$ with an informed node
via the different gates. Thus $Q$ only needs the messages sent from $T$ to it
(in this direction) to become informed, as this has all the evidential
information. All other messages sent to it along other paths are from the
outside neighbors. In this manner, the informed set $\mathcal{J}$ spreads to
$Q$.

For the Polytree $\mathbb{B}$ in Figure 5, suppose $\mathcal{J}=\left\{
D,C,H\right\}  $ and the outside node $R_{8}$\ is a query variable. The gate
from $R_{8}$ to $\mathcal{J}$ is $C$ and the messages along path $C\leftarrow
L_{3}\rightarrow R_{8}$ are all that are needed to make $R_{8}$\ informed.

Peot \& Shachter's revised polytree algorithm has been very much neglected in
the literature because not many practical BNs have this form. While they
further suggested that their algorithm be applied to a general BN via
\textquotedblleft cutset conditioning,\textquotedblright\ we continue this
paper by presenting a novel method to convert any BN into a polytree.

\section{The Border Polytrees (BPs)}

In Section 2, we showed that the border algorithm \textquotedblleft
stretches\textquotedblright\ a BN into a border chain, which is a special form
of the polytrees, in which each node has at most one parent and one child. For
example, we stretched the BN $\mathbb{A}$ in Figure \ref{F1} into a border
chain in Figure \ref{F5}. We will now show how to convert any BN into a polytree.

\subsection{Stage I: The Macro-node Polytrees}

Our method involves two stages: In Stage I, we partition the BN to form a polytree.

A set of nodes is said to be \textquotedblleft combinable\textquotedblright%
\ into a \textquotedblleft macro-node\textquotedblright\ if the network
remains acyclic after the combination of its members. For example, we cannot
combine nodes $A$ and $H$ in the BN $\mathbb{A}$, as this results in the
directed loop $\left\{  A,H\right\}  \rightarrow D\rightarrow\left\{
A,H\right\}  $. According to Chang \& Fung's (1989) Node Aggregation Theorem,
if all directed paths connecting any two nodes in a set do not contain a node
outside it, then this set is combinable. Hence, set $\left\{  A,C,D,H\right\}
$ in the BN $\mathbb{A}$ is.

If two loops in a DAG share a common node, they belong to the same
\textquotedblleft independent loop set\textquotedblright\ (ILS). If we convert
all ILSs in a DAG into polytrees, then the DAG itself becomes a polytree.
Because each ILS is combinable, we can combine the nodes in each for this
purpose. However this may yield some unnecessarily large macro-nodes. Chang \&
Fung suggest two methods of converting an ILS into a polytree; one is a
heuristic search through the space of its \textquotedblleft feasible
partitions\textquotedblright\ for the optimal node combinations, the other is
what they call the \textquotedblleft graph-directed
algorithm.\textquotedblright\ Here, we propose an algorithm\ which is somewhat
along the line of Ng and Levitt's method for incrementally extending what they
called a \textquotedblleft layered\textquotedblright\ polytree (1994).

\subsubsection{The Isolated Loops}

In a DAG, let us isolate one of its loops (that is, delete all nodes outside
it) and call it an \textquotedblleft isolated loop.\textquotedblright\ Some
non-root and non-leaf nodes of the DAG may now become the roots or leaves of
the isolated loop. On the other hand, some roots or leaves of the former may
no longer be in the latter.

\begin{figure}
[ptb]
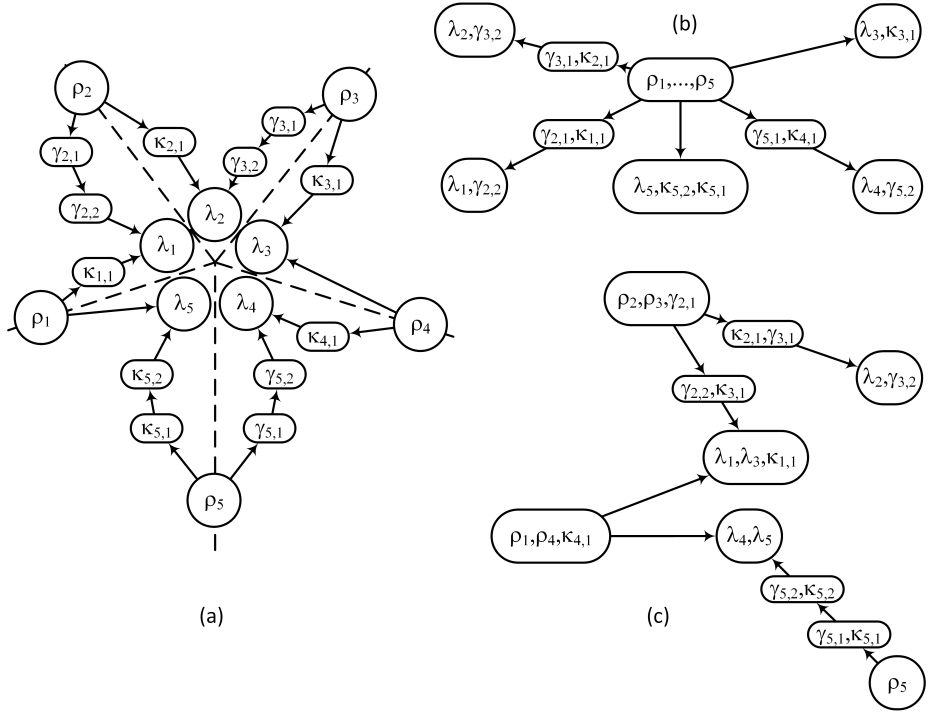

\begin{center}
\includegraphics[
natheight=3.744300in,
natwidth=4.950000in,
height=3.7443in,
width=4.95in
]%
6%
\caption{An isolated loop with five roots and five leaves}%
\label{F12}%
\end{center}
\end{figure}

Let us label the $r$ roots in a loop consecutively as $\rho_{1}$, ...,
$\rho_{r}$, with $\rho_{r+1}=\rho_{1}$. (See Figure \ref{F12}(a).)%

\begin{lemma}
An isolated loop has the same number of roots and leaves, and can be organized
into a star shape.
\end{lemma}

\begin{proof}
Two neighboring roots cannot be in a parent-child relationship, thus must be
connected from below via a leaf.

If there are $r=2$ roots, then we need two leaves to have a loop connecting
all four. However, if $r>2$ and there are more than one leaves in the direct
path connecting any two consecutive roots, we have a loop connecting all four,
in addition to the loop connecting all $r$ roots, which violates our isolated
loop assumption. Thus with $r$ roots, we have $r$ leaves. Figure \ref{F12}(a)
shows a typical loop which is organized into a star shape.
\end{proof}

Let $\lambda_{i}$ be the single leaf between $\rho_{i}$ and $\rho_{i+1}$. Let
$\kappa_{i,k}$ ($k=1,...,n_{i}$) be the $n_{i}$ nodes in the directed path
from $\rho_{i}$ to $\lambda_{i}$ and $\gamma_{i,k}$ ($k=1,...,m_{i}$) be the
$m_{i}$ nodes in the directed path from $\rho_{i+1}$ to $\lambda_{i}$.

There are more than one ways we can combine the nodes in an isolated loop in
order to \textquotedblleft open\textquotedblright\ it; that is, to make it a
polytree. For example, we can combine all its roots after some appropriate
node combinations to make path $\rho_{i}\rightarrow...\rightarrow\lambda_{i}$
and path $\rho_{i+1}\rightarrow...\rightarrow\lambda_{i}$ having the same
\textquotedblleft length;\textquotedblright\ that is, have the same number of
macro-nodes. For example, to have the polytree in Figure \ref{F12}(b), we form
path $\rho_{2}\rightarrow\gamma_{2,1}\rightarrow\left\{  \lambda_{1}%
,\gamma_{2,2}\right\}  $, so that it has the same length as path $\rho
_{1}\rightarrow\kappa_{1,1}\rightarrow\lambda_{1}$, before combining them into
path $\left\{  \rho_{1},\rho_{2}\right\}  \rightarrow\left\{  \gamma
_{2,1},\kappa_{1,1}\right\}  \rightarrow\left\{  \lambda_{1},\gamma
_{2,2}\right\}  $. Similarly, we can combine all its leaves.

Another novel method is to \textquotedblleft fold\textquotedblright\ the loop
along one of its axes, bringing together the pairs of opposite paths in order
to combine them, creating an acyclic semi-star. The star must be made
\textquotedblleft symmetrical\textquotedblright\ around the chosen axis first,
in that each pair of opposite paths must have the same length. The loop in
Figure \ref{F12}(a) is not symmetrical around axis $\lambda_{2}-\rho_{5}$. To
make it symmetrical, among other combinations, we may form path $\left\{
\rho_{2},\gamma_{2,1}\right\}  \rightarrow\gamma_{2,2}\rightarrow\lambda_{1}$,
so that it can be combined with path $\rho_{3}\rightarrow\kappa_{3,1}%
\rightarrow\lambda_{3}$. The polytree in Figure \ref{F12}(c) is one possible
result of folding the loop in Figure \ref{F12}(a) along its axis $\lambda
_{2}-\rho_{5}$.

So it seems that we can covert a DAG into a polytree by open all loops in
isolation, one-by-one in this manner. Unfortunately, this procedure generally
is not correct, as this may make other loops in the same ILS cyclic. For
example, if there is an additional directed path $\rho_{2}\rightarrow
\eta\rightarrow\rho_{3}$ in Figure \ref{F12}(a), then combining $\rho_{2}$ and
$\rho_{3}$ as in Figure \ref{F12}(c) violates the Node Aggregation Theorem.
Furthermore, these cyclic loops may not be easily identifiable.

We now suggest a correct and systematic way to identify and open all loops in
a DAG.

\subsubsection{The Parentless Polytree Method}

In our \textquotedblleft parentless polytree method,\textquotedblright\ we
construct a growing parentless polytree from a DAG as follows: We start with a
root in the DAG and execute the following steps:

\begin{enumerate}
\item While there is an un-recruited node

\begin{enumerate}
\item Recruit node $\tau$ in a topological order (that is, after all its
parents $\left\{  \pi_{1},...,\pi_{p}\right\}  $), together with edge
$\tau\leftarrow\pi_{1}$

\item While there is an un-recruited edge $\tau\leftarrow\pi_{i}$

\begin{enumerate}
\item recruit edge $\tau\leftarrow\pi_{i}$

\item isolate and open any resulted loop
\end{enumerate}

\item End while

\item Go to Step 1
\end{enumerate}

\item End while
\end{enumerate}

The growing polytree must be kept parentless, so that no directed path can
return to it, and thus no directed cycle via a node outside the polytree can
occur as a result of any node combinations inside the polytree.

When we recruit node $\tau$ having $p$ parents into the polytree $\mathcal{P}%
$, we bring with it $p$ edges $\tau\leftarrow\pi_{i}$ ($i=1,...,p$), where
each pair $\left\{  \pi_{i},\pi_{j}\right\}  $ is connected by at most one
path. (We have no path if $\pi_{i}$ and $\pi_{j}$ are in two unconnected
subsets of $\mathcal{P}$.) Recruiting $\tau$ with all its parents creates at
most $C_{2}^{p}$\ loops.

\textquotedblleft Recruiting edge $\tau\leftarrow\pi_{i}$\textquotedblright%
\ means adding it in the polytree. Following Ng and Levitt (1994), not only do
we recruit the nodes one-at-a-time, we also recruit the edges one-at-a-time.
If $\tau$ has only one parent, then with edge $\tau\leftarrow\pi_{1}$ we still
have a polytree. Otherwise, after recruiting edge $\tau\leftarrow\pi_{2}$, we
have at most one loop $\tau\leftarrow\pi_{1}-...-\pi_{2}\rightarrow\tau$.
After opening this loop, we again have a polytree. Then edge $\tau
\leftarrow\pi_{3}$ yields at most another loop via $\pi_{3}$.... In this way,
we only have to open at most $p-1$ easily identifiable loops.

If two loops created by $\tau$ have different pairs of parents $\left\{
\pi_{i},\pi_{j}\right\}  $ and $\left\{  \pi_{k},\pi_{m}\right\}  $, then
opening one cannot make the other cyclic. So let us consider two loops having
parents $\left\{  \pi_{i},\pi_{j}\right\}  $ and $\left\{  \pi_{j},\pi
_{k}\right\}  $ and recruit edges $\tau\leftarrow\pi_{i}$ and $\tau
\leftarrow\pi_{j}$ first, resulting in a loop having path $\pi_{i}%
\rightarrow\tau\leftarrow\pi_{j}$. (If $\mathcal{P}$\ is not parentless, we
may have a loop with path $\pi_{i}\rightarrow\tau\rightarrow\psi$ instead.) If
we open this loop without combining $\tau$ with any node, then when we recruit
edge $\tau\leftarrow\pi_{k}$ later, the resulting loop is acyclic, because
path $\pi_{j}\rightarrow\tau\leftarrow\pi_{k}$ is still in it. It is feasible
not to combine leaf $\tau$ in the folding method, as we do not need the
macro-node $\left\{  \tau,\pi_{i}\right\}  $ to reduce the length of path
$\tau\leftarrow\pi_{i}\leftarrow...\leftarrow\rho_{k}$ and we can choose the
symmetrical axis going through $\tau$, so that it is not combined with any
other leaf.

But suppose we wish to combine leaf $\tau$ with node $\eta$ in the loop with
$\left\{  \pi_{i},\pi_{j}\right\}  $, and only to find out later when
recruiting edge $\tau\leftarrow\pi_{k}$ that loop $\left\{  \eta,\tau\right\}
\rightarrow...\rightarrow\pi_{k}\rightarrow\left\{  \eta,\tau\right\}  $ is
cyclic because of the directed path $\eta\rightarrow...\rightarrow\pi
_{k}\rightarrow\tau$. In this case, we \textquotedblleft
open\textquotedblright\ this loop by expand the macro-node $\left\{  \eta
,\tau\right\}  $ to $\left\{  \eta,...,\pi_{k},\tau\right\}  $. If there is
any loop in this macro-node, it has the form $\tau\leftarrow...-\eta
\rightarrow...\rightarrow\pi_{k}\rightarrow\tau$ and therefore is not cyclic.
After recruiting all $p$ edges, we have a polytree (without any cycle, hence
is acyclic), and can return to Step 1.

After recruiting all nodes and their edges in a DAG in this manner, we will
have partitioned the DAG into what we call a \textquotedblleft macro-node
polytree.\textquotedblright%

\begin{figure}
[ptb]
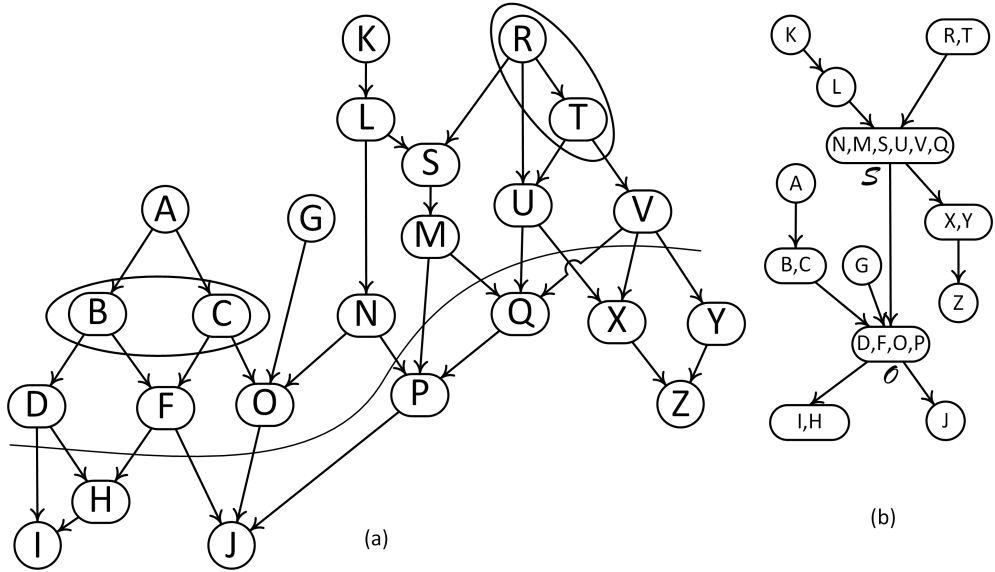

\begin{center}
\includegraphics[
natheight=3.070000in,
natwidth=5.180600in,
height=3.07in,
width=5.1806in
]%
7%
\caption{The Bayesian Network $\mathbb{C}$, or BN $\mathbb{C}$}%
\label{F8}%
\end{center}
\end{figure}

For illustration, let us use the BN in Figure \ref{F8}, which we will refer to
as the Bayesian network $\mathbb{C}$, or BN $\mathbb{C}$. Assume that, after
forming the macro-nodes $\left\{  R,T\right\}  $ and $\left\{  B,C\right\}  $,
we arrive at the parentless polytree above the curved line in Figure \ref{F8}(a).

\begin{enumerate}
\item Recruiting node $X$ results in loop $X-U-\left\{  R,T\right\}  -V-X$,
which can be opened by forming the macro-node $\left\{  U,V\right\}  .$

\item Recruiting node $Y$ and then node $Z$ results in loop $Z-Y-\left\{
U,V\right\}  -X-Z$, which can be opened by forming the macro-node $\left\{
X,Y\right\}  .$

\item Recruiting node $Q$ results in loop $Q-\left\{  U,V\right\}  -\left\{
R,T\right\}  -S-M-Q$, which can be opened by forming the macro-node $\left\{
M,S,U,V\right\}  .$

\item Recruiting edges $P\leftarrow\left\{  M,S,U,P\right\}  $ and
$P\leftarrow Q$ results in loop $P-Q-\left\{  M,S,U,V\right\}  -P$, which can
be opened by forming the macro-node $\left\{  M,S,U,V,Q\right\}  .$

\item Recruiting edge $P\leftarrow N$ results in loop $P-N-L-\left\{
M,S,U,V,Q\right\}  -P$, which can be opened by forming the macro-node
$\left\{  N,M,S,U,V,Q\right\}  .$

\item Recruiting edges $J\leftarrow P$ and $J\leftarrow O$ results in loop
$J-P-\left\{  N,M,S,U,V,Q\right\}  -O-J$, which can be opened by forming the
macro-node $\left\{  O,P\right\}  .$

\item Recruiting edge $J\leftarrow F$ results in loop $J-\left\{  O,P\right\}
-\left\{  B,C\right\}  -F-J$, which can be opened by forming the macro-node
$\left\{  F,O,P\right\}  .$

\item Recruiting node $H$ results in loop $H-\left\{  F,O,P\right\}  -\left\{
B,C\right\}  -D-H$, which can be opened by forming the macro-node $\left\{
D,F,O,P\right\}  .$

\item Recruiting node $I$ results in loop $I-H-\left\{  D,F,O,P\right\}  -I$,
which can be opened by forming the macro-node $\left\{  I,H\right\}  .$
\end{enumerate}

We now can apply the revised polytree algorithm to the final macro-node
polytree corresponding to the BN $\mathbb{C}$, as shown in Figure \ref{F8}(b).
However, the size of some CPTs can be large, such as
\begin{equation}
\Pr\left\{  D,F,O,P|N,M,S,U,V,Q,G,B,C\right\}  =\Pr\left\{  \mathcal{O}%
|\mathcal{S},G,B,C\right\}  , \label{60}%
\end{equation}
where $\mathcal{O}=\left\{  D,F,O,P\right\}  $ and $\mathcal{S}=\left\{
N,M,S,U,V,Q\right\}  $. This is why we do not stop after Stage I, but continue
to Stage II.

\subsection{Stage II: The Border Polytrees}

In Stage II, we use the border algorithm, which explores the independence
relationships between the individual nodes within the macro-nodes, to stretch
each macro-node into a border chain, if it is not already in this form.

\begin{figure}
[ptb]
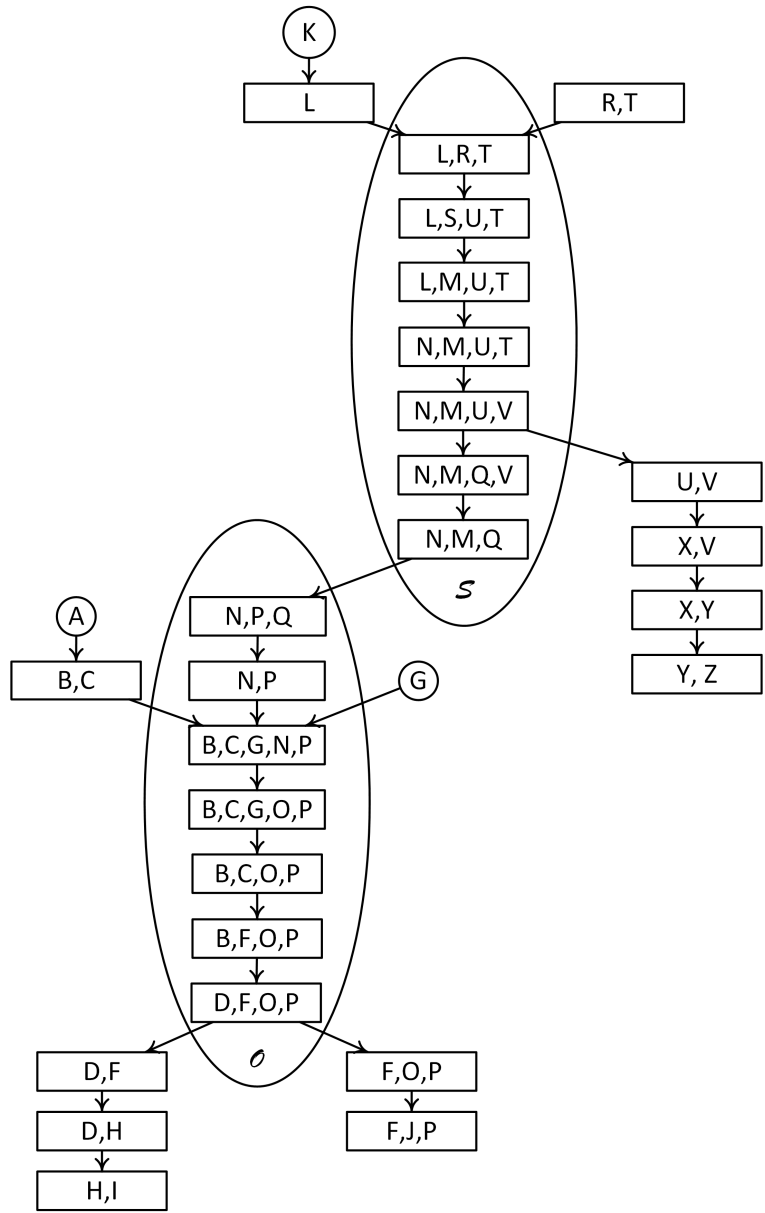

\begin{center}
\includegraphics[
natheight=6.372400in,
natwidth=3.938000in,
height=6.3724in,
width=3.938in
]%
8%
\caption{BP $\mathbb{C}$, a border polytree for the BN $\mathbb{C}$}%
\label{F10}%
\end{center}
\end{figure}

A result obtained for the BN $\mathbb{C}$ after Stage II is shown in Figure
\ref{F10}.%

Inside a macro-node, we follow the eight rules in \S 2.2 for promoting a node
from a border and recruiting its cohort, so that each border separates the top
part of the macro-node with its bottom part. However, because not all
macro-notes are parentless, we need to introduce the following additional rules:

\begin{enumerate}
\item[9.] We cannot stretch a macro-node before its parents. In the BN
$\mathbb{C}$, we must start with the macro-node $\left\{  K\right\}  $,
$\left\{  R,T\right\}  $, $\left\{  G\right\}  $\ or $\left\{  A\right\}  $.

\item[10.] Macro-node $\Gamma$\ cannot recruit any member of its macro-node
child $\Delta$. In the macro-node $\mathcal{S}$ of the BN $\mathbb{C}$, border
$\left\{  N,M,Q\right\}  $ is obtained from border $\left\{  N,M,Q,V\right\}
$ by promoting $V$ without recruiting its child $Y$ in macro-node $\left\{
X,Y\right\}  $.

\item[11.] If macro-node $\Delta$\ has parents in macro-node $\Gamma$, the
entire set $\Gamma\cap\mathcal{H}_{\Delta}$ must be together in at least one
border in $\Gamma$. In the BN $\mathbb{C}$, because $\mathcal{S\cap
H}_{\left\{  X,Y\right\}  }=\left\{  U,V\right\}  $ we form border $\left\{
N,M,U,V\right\}  \subseteq\mathcal{S}$, because $\mathcal{S\cap H}%
_{\mathcal{O}}=\left\{  N,M,Q\right\}  $ we form border $\left\{
N,M,Q\right\}  \subseteq\mathcal{S}$. Also, because $\mathcal{O\cap
H}_{\left\{  H,I\right\}  }=\left\{  D,F\right\}  $ and $\mathcal{O\cap H}%
_{J}=\left\{  F,O,P\right\}  $, we form border $\left\{  D,F,O,P\right\}
\subseteq\mathcal{O}$.

\item[12.] Suppose we wish to promote node $B\in\mathcal{B}_{i-1}%
\subseteq\Delta$. By Rule 10, we do not recruit $\mathcal{L}_{B}%
\nsubseteq\Delta$. However, we need to recruit all un-recruited members of
$\mathcal{L}_{B}\cap\Delta$ (that is, $\mathcal{F}_{B}=\mathcal{L}_{B}%
\cap\mathcal{D}_{\mathcal{B}_{i-1}}\cap\Delta$) and all members of
$\mathcal{H}_{\mathcal{F}_{B}}\subseteq\mathcal{V}$ (that is, the co-parents
of $B$), even if they are not in $\Delta$.

Assume there are $r$ macro-nodes $\Gamma_{k}\neq\Delta$ ($k=1,...,r$) such
that $\Gamma_{k}\cap\mathcal{H}_{\mathcal{F}_{B}}\neq\varnothing$. By Rules 9
and 11, we must have already constructed in $\Gamma_{k}$ a border
$\mathcal{B}_{i-1,k}$ such that $\left\{  \Gamma_{k}\cap\mathcal{H}%
_{\mathcal{F}_{B}}\right\}  \subseteq\mathcal{B}_{i-1,k}^{\ast}=\left\{
\Gamma_{k}\cap\mathcal{H}_{\Delta}\right\}  \subseteq\mathcal{B}_{i-1,k}$. For
notational advantages, we also denote $\mathcal{B}_{i-1}\subseteq\Delta$ by
$\mathcal{B}_{i-1,0}$ or $\mathcal{B}_{i-1,0}^{\ast}$.

We break our procedure into two steps: (i) We first recruit to $\Delta$ all
$\Gamma_{k}\cap\mathcal{H}_{\Delta}$ ($k=1,...,r$) (which have not been
recruited) to form border
\begin{equation}
\mathcal{B}_{i}=\mathcal{B}_{i-1}\cup_{j=1}^{r}\left\{  \Gamma_{k}%
\cap\mathcal{H}_{\Delta}\right\}  =\cup_{j=0}^{r}\mathcal{B}_{i-1,j}^{\ast
}\supseteq\mathcal{H}_{\mathcal{F}_{B}}, \label{63}%
\end{equation}
where $\mathcal{B}_{i}$ has all $\mathcal{B}_{i-1,k}$ as parents. (ii)\ Then
we promote node $B$ from border $\mathcal{B}_{i}$\ with cohort $\mathcal{F}%
_{B}\subseteq\Delta$ by Rule 2.

Note that the entire set $\Gamma_{k}\cap\mathcal{H}_{\Delta}$ constructed by
Rule 11 must be recruited into $\Delta$ together. This is to ensure that there
is at most one edge connecting any two stretched macro-nodes. Otherwise, the
main benefit of the macro-node polytrees is destroyed.

In the BN $\mathbb{C}$, suppose we wish to promote node $N$ from border
$\left\{  N,P\right\}  \subset\mathcal{O}$. Because $\mathcal{F}%
_{N}=\mathcal{L}_{N}\cap\mathcal{D}_{\left\{  N,P\right\}  }\cap\mathcal{O}%
=O$, we need to recruit nodes $C\in\left\{  B,C\right\}  \cap\mathcal{H}_{O}$
and $G\in\left\{  G\right\}  \cap\mathcal{H}_{O}$. However, Rule 11 puts
$\left\{  B,C\right\}  \cap\mathcal{H}_{\mathcal{O}}=\left\{  B,C\right\}  $
in one border and instead of $C$ alone, we recruit the whole set $\left\{
B,C\right\}  $. We thus form border $\cup_{j=0}^{2}\mathcal{B}_{i-1,j}^{\ast}$
with $\mathcal{B}_{i-1,0}^{\ast}=\left\{  N,P\right\}  $, $\mathcal{B}%
_{i-1,1}^{\ast}=\left\{  B,C\right\}  $, $\mathcal{B}_{i-1,2}^{\ast}=\left\{
G\right\}  $ before promoting $N$\ with cohort $O$.

\item[13.] Rule 12 also applies when we start stretching a non-root macro-node
$\Delta$. We do not start with the nodes having parents in it, but with node
$V$ having parents only in other macro-nodes $\Gamma_{k}\neq\Delta$
($k=1,...,r$); that is, $\Delta\cap\mathcal{H}_{V}=\varnothing$, and
$\Gamma_{k}\cap\mathcal{H}_{V}\neq\varnothing$. In this case we first form
border $\mathcal{B}_{i}$ as in Equation (\ref{63}), with $\mathcal{B}%
_{i-1,0}=\mathcal{B}_{i-1,0}^{\ast}=\varnothing$ and with $\mathcal{B}%
_{i-1,k}^{\ast}$ such that $\left\{  \Gamma_{k}\cap\mathcal{H}_{V}\right\}
\subseteq\mathcal{B}_{i-1,k}^{\ast}=\left\{  \Gamma_{k}\cap\mathcal{H}%
_{\Delta}\right\}  \subseteq\mathcal{B}_{i-1,k}$. Then we can recruit $V$.

In the BN $\mathbb{C}$, to start stretching macro-node $\mathcal{S}$ with node
$S$ having parents in macro-nodes $\left\{  L\right\}  $ and $\left\{
R,T\right\}  $, we first form border $\cup_{j=1}^{2}\mathcal{B}_{i-1,j}^{\ast
}$ with $\mathcal{B}_{i-1,1}^{\ast}=\mathcal{H}_{\mathcal{S}}\cap\left\{
L\right\}  =L$ and $\mathcal{B}_{i-1,2}^{\ast}=\mathcal{H}_{\mathcal{S}}%
\cap\left\{  R,T\right\}  =\left\{  R,T\right\}  $. We then can promote $R$
with cohort $\left\{  S,U\right\}  $.

If $r=1$ and $\mathcal{B}_{i-1,1}^{\ast}=\mathcal{B}_{i-1,1}$, then there is
no need to repeat $\mathcal{B}_{i}=\mathcal{B}_{i-1,1}$. Suppose we wish to
start stretching macro-node $\mathcal{O}$ in the BN $\mathbb{C}$ by recruiting
node $P$. Because $P$ only has parents in $\mathcal{S}$, there is no need to
repeat border $\mathcal{B}_{i-1,1}^{\ast}=\mathcal{B}_{i-1,1}=\mathcal{S\cap
H}_{\mathcal{O}}=\left\{  N,M,Q\right\}  $. We simply promote $M$ to recruit
$P$.
\end{enumerate}

After stretching all macro-nodes, we obtain a \textquotedblleft border
polytree\textquotedblright\ or a BP, made up of the \textquotedblleft
borders.\textquotedblright\ The variables in each border are called its
\textquotedblleft component variables.\textquotedblright\ The BP in Figure
\ref{F10} is called the BP $\mathbb{C}$.

\begin{figure}
[ptb]
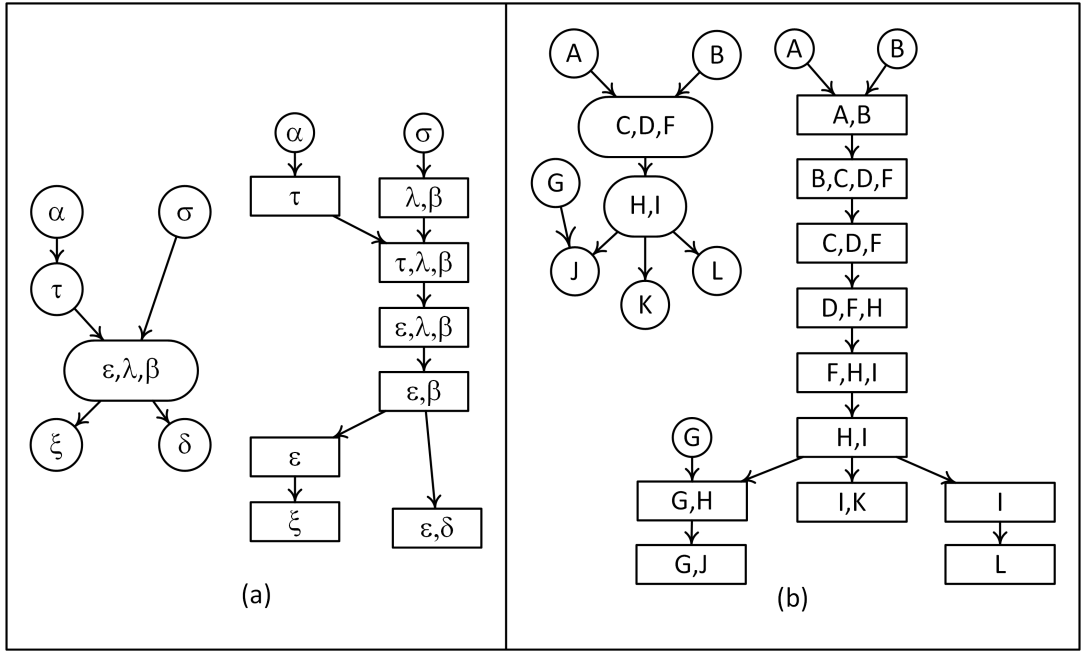

\begin{center}
\includegraphics[
natheight=3.395600in,
natwidth=5.637200in,
height=3.3956in,
width=5.6372in
]%
9%
\caption{The macro-node \& border polytrees for (a) the Dyspnoea example and
(b) the BN $\mathbb{A}$}%
\label{F11}%
\end{center}
\end{figure}

For comparison with a undirected junction tree of the well-known
\textquotedblleft Dyspnoea\textquotedblright\ example by Lauritzen \&
Spiegelhalter (1988, pp.~164 and 209), we include in Figure \ref{F11}(a) its
macro-node and border polytrees. Figure \ref{F11}(b) shows the macro-node and
border polytrees of the BN $\mathbb{A}$.%

Like a clique tree (or a junction tree), a BP has the \textquotedblleft
running intersection property;\textquotedblright\ that is, after a variable is
recruited into border $\mathcal{B}_{j}$, it stays in all future consecutive
borders $\mathcal{B}_{j+1}$, $\mathcal{B}_{j+2}$,... until its promotion; then
it will never be recruited again into any later borders. However, a BP is not
\textquotedblleft family preserving,\textquotedblright\ in that it is not
necessary that a variable must be with all its parents in one border. Most
importantly, a BP is directional, while a clique tree is not.

The borders in a BP generally have smaller CPTs than the macro-nodes in
macro-node polytree. In the BN $\mathbb{C}$, instead of the CPT as in Equation
(\ref{60}) involving thirteen variables, we now have $\Pr\left\{
D,F,O,P|B,F,O,P\right\}  $, with five variables $\left\{  D,F,O,P,B\right\}
$. Despite their smaller CPTs, their posterior marginal probabilities still
can be calculated by adapting the results in Section 4 as we will show next.

\subsection{The Message Propagations in a Border Polytree}

There are two types of non-root borders in a BP:

\begin{enumerate}
\item $\mathcal{B}_{i}=\left\{  \mathcal{B}_{i-1},\mathcal{C}_{i}\right\}
\backslash V_{i}$, obtained by following Rules 1-8 in \S 2.2, having
$\mathcal{B}_{i-1}$ as a single parent. (If $\mathcal{B}_{i}=X$ and
$\mathcal{B}_{i-1}=Y$, then $\mathcal{C}_{i}=X$, $V_{i}=Y$.)

\item $\mathcal{B}_{i}=\cup_{j=0}^{r}\mathcal{B}_{i-1,j}^{\ast}$, obtained by
Rule 12 or 13, with $\mathcal{H}_{\mathcal{B}_{i}}=\left\{  \mathcal{B}%
_{i-1,j},j=0,...,r\right\}  $.
\end{enumerate}

Returning to the proofs of Lemmas \ref{P7} and \ref{P2}, we see that they are
still valid for the BPs. However, we need to modify Lemmas \ref{P5} and
\ref{P1} according to the type of border in a BP.

\subsubsection{The Downward Propagations}

We first modify Lemma \ref{P1}:

\begin{lemma}
\label{P34}For border $\mathcal{B}_{i}=\left\{  \mathcal{B}_{i-1}%
,\mathcal{C}_{i}\right\}  \backslash V_{i}$,
\[
\Pi\left(  \mathcal{B}_{i}\right)  =\sum_{V_{i}}\phi\left(  \mathcal{C}%
_{i}\right)  \Pi_{\mathcal{B}_{i}}\left(  \mathcal{B}_{i-1}\right)  .
\]
If $\mathcal{B}_{i}$ is the only child of $\mathcal{B}_{i-1},$%
\[
\Pi\left(  \mathcal{B}_{i}\right)  =\sum_{V_{i}}\phi\left(  \mathcal{C}%
_{i}\right)  \Pi\left(  \mathcal{B}_{i-1}\right)  .
\]

\end{lemma}

\begin{proof}
Consider the parentless $\mathcal{T}_{\mathcal{B}_{i-1}\rightarrow
\mathcal{B}_{i}}$ with border $\mathcal{B}_{i-1}$. Promoting $V_{i}%
\in\mathcal{B}_{i-1}$ with cohort $\mathcal{C}_{i}$ yields the parentless
$\mathcal{P}_{\mathcal{B}_{i}}$ with border $\mathcal{B}_{i}$. This because,
with $\left\{  \mathcal{B}_{i-1}\backslash V_{i}\right\}  \subseteq
\mathcal{T}_{\mathcal{B}_{i-1}\rightarrow\mathcal{B}_{i}}$,
\[
\mathcal{T}_{\mathcal{B}_{i-1}\rightarrow\mathcal{B}_{i}}\cup\mathcal{C}%
_{i}=\mathcal{T}_{\mathcal{B}_{i-1}\rightarrow\mathcal{B}_{i}}\cup\left\{
\mathcal{B}_{i-1}\backslash V_{i}\right\}  \cup\mathcal{C}_{i}=\mathcal{T}%
_{\mathcal{B}_{i-1}\rightarrow\mathcal{B}_{i}}\cup\mathcal{B}_{i}%
=\mathcal{P}_{\mathcal{B}_{i}}\text{.}%
\]
The lemma follows from Theorem \ref{FE}.

If $\mathcal{B}_{i}$ is the only child of $\mathcal{B}_{i-1}$, by Lemma
\ref{P7}, $\Pi_{\mathcal{B}_{i}}\left(  \mathcal{B}_{i-1}\right)  =\Pi\left(
\mathcal{B}_{i-1}\right)  $.
\end{proof}

\begin{lemma}
\label{P35}For border $\mathcal{B}_{i}=\cup_{j=0}^{r}\mathcal{B}_{i-1,j}%
^{\ast}$,%
\[
\Pi\left(  \mathcal{B}_{i}\right)  =\prod\limits_{j=0}^{r}\sum_{\mathcal{B}%
_{i-1,j}\backslash\mathcal{B}_{i-1,j}^{\ast}}\Pi_{\mathcal{B}_{i}}\left(
\mathcal{B}_{i-1,j}\right)  .
\]

\end{lemma}

\begin{proof}
Consider the parentless $\overline{\mathcal{T}}_{i-1}=\cup_{j=0}%
^{r}\mathcal{T}_{\mathcal{B}_{i-1,j}\rightarrow\mathcal{B}_{i}}$ having border
$\overline{\mathcal{B}}_{i-1}=\cup_{j=0}^{r}\mathcal{B}_{i-1,j}$. Because all
$\mathcal{T}_{\mathcal{B}_{i-1,j}\rightarrow\mathcal{B}_{i}}$ are parentless
and independent, from Definitions (\ref{49}) and (\ref{78}),%
\begin{align*}
\Pi\left(  \overline{\mathcal{B}}_{i-1}\right)   &  =\Pr\left\{
\overline{\mathcal{B}}_{i-1},\left[  \overline{\mathcal{T}}_{i-1}%
\backslash\overline{\mathcal{B}}_{i-1}\right]  \right\}  \boldsymbol{I}%
_{\overline{\mathcal{B}}_{i-1}}\\
&  =\Pr\left\{  \cup_{j=0}^{r}\left\{  \mathcal{B}_{i-1,j},\left[
\mathcal{T}_{\mathcal{B}_{i-1,j}\rightarrow\mathcal{B}_{i}}\backslash
\mathcal{B}_{i-1,j}\right]  \right\}  \right\}  \boldsymbol{I}_{\cup_{j=0}%
^{r}\mathcal{B}_{i-1,j}}\\
&  =\prod\limits_{j=0}^{r}\Pr\left\{  \mathcal{B}_{i-1,j},\left[
\mathcal{T}_{\mathcal{B}_{i-1,j}\rightarrow\mathcal{B}_{i}}\backslash
\mathcal{B}_{i-1,j}\right]  \right\}  \boldsymbol{I}_{\mathcal{B}_{i-1,j}%
}=\prod\limits_{j=0}^{r}\Pi_{\mathcal{B}_{i}}\left(  \mathcal{B}%
_{i-1,j}\right)  .
\end{align*}
Because $\mathcal{B}_{i-1,j}\backslash\mathcal{B}_{i-1,j}^{\ast}$ does not
have children in the bottom set $\left\{  \mathcal{B}_{i},\mathcal{D}%
_{\mathcal{B}_{i}}\right\}  $, we use Rule 1 to promote $\cup_{j=0}%
^{r}\left\{  \mathcal{B}_{i-1,j}\backslash\mathcal{B}_{i-1,j}^{\ast}\right\}
$ without cohort to obtain the same parentless $\overline{\mathcal{T}}_{i-1}$
with border $\mathcal{B}_{i}=\cup_{j=0}^{r}\mathcal{B}_{i-1,j}^{\ast}$. From
Theorem \ref{FE},%
\[
\Pi\left(  \mathcal{B}_{i}\right)  =\sum_{\cup_{j=0}^{r}\left\{
\mathcal{B}_{i-1,j}\backslash\mathcal{B}_{i-1,j}^{\ast}\right\}  }\Pi\left(
\overline{\mathcal{B}}_{i-1}\right)  =\sum_{\cup_{j=0}^{r}\left\{
\mathcal{B}_{i-1,j}\backslash\mathcal{B}_{i-1,j}^{\ast}\right\}  }%
\prod\limits_{j=0}^{r}\Pi_{\mathcal{B}_{i}}\left(  \mathcal{B}_{i-1,j}\right)
,
\]
hence the lemma.
\end{proof}

Combining Lemma \ref{P7} with above lemmas yields the following BP version of
Theorem \ref{P20}.

\begin{theorem}
\label{P30}If border $\mathcal{B}_{i}$ has received the messages from all
members of $\mathcal{H}_{\mathcal{B}_{i}}\cup\left\{  \mathcal{L}%
_{\mathcal{B}_{i}}\backslash\mathcal{B}_{i+1,j}\right\}  $, then it can send a
downward message to its child $\mathcal{B}_{i+1,j}$ as%
\[
\Pi_{\mathcal{B}_{i+1,j}}\left(  \mathcal{B}_{i}\right)  =\Pi\left(
\mathcal{B}_{i}\right)  \left(  \prod\limits_{\mathcal{W}\in\mathcal{L}%
_{\mathcal{B}_{i}}\backslash\mathcal{B}_{i+1,j}}\Lambda_{\mathcal{W}}\left(
\mathcal{B}_{i}\right)  \right)  ,
\]
where $\Pi\left(  \mathcal{B}_{i}\right)  $ is calculated by Lemma \ref{P34}
or \ref{P35}.
\end{theorem}

\subsubsection{The Upward Propagations}

We now present the following BP versions of Lemma \ref{P5}:

\begin{lemma}
\label{P32}For border $\mathcal{B}_{i}=\left\{  \mathcal{B}_{i-1}%
,\mathcal{C}_{i}\right\}  \backslash V_{i}$,%
\[
\Lambda_{\mathcal{B}_{i}}\left(  \mathcal{B}_{i-1}\right)  =\sum
_{\mathcal{C}_{i}}\phi\left(  \mathcal{C}_{i}\right)  \Lambda\left(
\mathcal{B}_{i}\right)  .
\]
If $\mathcal{B}_{i}$ is the only child of $\mathcal{B}_{i-1},$%
\[
\Lambda\left(  \mathcal{B}_{i-1}\right)  =\sum_{\mathcal{C}_{i}}\phi\left(
\mathcal{C}_{i}\right)  \Lambda\left(  \mathcal{B}_{i}\right)  .
\]

\end{lemma}

\begin{proof}
Consider the parentless $\mathcal{T}_{\mathcal{B}_{i-1}\rightarrow
\mathcal{B}_{i}}$ with border $\mathcal{B}_{i-1}$. As shown in Lemma
\ref{P34}, promoting $V_{i}\in\mathcal{B}_{i-1}$ with cohort $\mathcal{C}_{i}$
yields the parentless $\mathcal{P}_{\mathcal{B}_{i}}$ with border
$\mathcal{B}_{i}$. The lemma follows from Theorem \ref{B}.

If $\mathcal{B}_{i}$ is the only child of $\mathcal{B}_{i-1}$, by Lemma
\ref{P2}, $\Lambda_{\mathcal{B}_{i}}\left(  \mathcal{B}_{i-1}\right)
=\Lambda\left(  \mathcal{B}_{i-1}\right)  $.
\end{proof}

\begin{lemma}
\label{P33}If border $\mathcal{B}_{i}=\cup_{j=0}^{r}\mathcal{B}_{i-1,j}^{\ast
}$, then for all $0\leq k\leq r$,%
\[
\Lambda_{\mathcal{B}_{i}}\left(  \mathcal{B}_{i-1,k}\right)  =\sum
_{\mathcal{H}_{\mathcal{B}_{i}}\backslash\mathcal{B}_{i-1,k}}\Lambda\left(
\mathcal{B}_{i}\right)  \prod\limits_{\mathcal{W}\in\mathcal{H}_{\mathcal{B}%
_{i}}\backslash\mathcal{B}_{i-1,k}}\Pi_{\mathcal{B}_{i}}\left(  \mathcal{W}%
\right)  .
\]

\end{lemma}

\begin{proof}
Consider the parentless $\mathcal{T}_{\mathcal{B}_{i-1,k}\rightarrow
\mathcal{B}_{i}}$ $\left(  0\leq k\leq r\right)  $ having border
$\mathcal{B}_{i-1,k}$. Let $\mathcal{I}_{i-1,k}=\mathcal{H}_{\mathcal{B}_{i}%
}\backslash\mathcal{B}_{i-1,k}$. We recruit $\cup_{\mathcal{W}\in
\mathcal{I}_{i-1,k}}\mathcal{T}_{\mathcal{W}\rightarrow\mathcal{B}_{i}}$
without promotion, resulting in the parentless $\overline{\mathcal{T}}%
_{i-1}=\cup_{j=0}^{r}\mathcal{T}_{\mathcal{B}_{i-1,j}\rightarrow
\mathcal{B}_{i}}$ having border $\overline{\mathcal{B}}_{i-1}=\cup_{j=0}%
^{r}\mathcal{B}_{i-1,j}$. From Theorem \ref{B},%
\begin{align*}
&  \Lambda_{\mathcal{B}_{i}}\left(  \mathcal{B}_{i-1,k}\right) \\
&  =\sum_{\cup_{\mathcal{W}\in\mathcal{I}_{i-1,k}}\mathcal{T}_{\mathcal{W}%
\rightarrow\mathcal{B}_{i}}}\left(  \prod\limits_{\mathcal{W}\in
\mathcal{I}_{i-1,k}}\Pr\left\{  \mathcal{T}_{\mathcal{W}\rightarrow
\mathcal{B}_{i}}\right\}  \boldsymbol{I}_{\mathcal{T}_{\mathcal{W}%
\rightarrow\mathcal{B}_{i}}}\right)  \Lambda\left(  \overline{\mathcal{B}%
}_{i-1}\right) \\
&  =\sum_{\mathcal{I}_{i-1,k}}\left\{  \sum_{\cup_{\mathcal{W}\in
\mathcal{I}_{i-1,k}}\left\{  \mathcal{T}_{\mathcal{W}\rightarrow
\mathcal{B}_{i}}\backslash\mathcal{W}\right\}  }\left(  \prod
\limits_{\mathcal{W}\in\mathcal{I}_{i-1,k}}\Pr\left\{  \mathcal{T}%
_{\mathcal{W}\rightarrow\mathcal{B}_{i}}\right\}  \boldsymbol{I}%
_{\mathcal{T}_{\mathcal{W}\rightarrow\mathcal{B}_{i}}}\right)  \Lambda\left(
\overline{\mathcal{B}}_{i-1}\right)  \right\}  .
\end{align*}
\newline Because $\overline{\mathcal{B}}_{i-1}\cap\left\{  \cup_{\mathcal{W}%
\in\mathcal{I}_{i-1,k}}\mathcal{T}_{\mathcal{W}\rightarrow\mathcal{B}_{i}%
}\backslash\mathcal{W}\right\}  =\varnothing$,
\begin{align*}
&  \Lambda_{\mathcal{B}_{i}}\left(  \mathcal{B}_{i-1,k}\right) \\
&  =\sum_{\mathcal{I}_{i-1,k}}\Lambda\left(  \overline{\mathcal{B}}%
_{i-1}\right)  \left\{  \sum_{\cup_{\mathcal{W}\in\mathcal{I}_{i-1,k}}\left\{
\mathcal{T}_{\mathcal{W}\rightarrow\mathcal{B}_{i}}\backslash\mathcal{W}%
\right\}  }\left(  \prod\limits_{\mathcal{W}\in\mathcal{I}_{i-1,k}}\Pr\left\{
\mathcal{T}_{\mathcal{W}\rightarrow\mathcal{B}_{i}}\right\}  \boldsymbol{I}%
_{\mathcal{T}_{\mathcal{W}\rightarrow\mathcal{B}_{i}}}\right)  \right\} \\
&  =\sum_{\mathcal{I}_{i-1,k}}\Lambda\left(  \overline{\mathcal{B}}%
_{i-1}\right)  \prod\limits_{\mathcal{W}\in\mathcal{I}_{i-1,k}}\left(
\sum_{\mathcal{T}_{\mathcal{W}\rightarrow\mathcal{B}_{i}}\backslash
\mathcal{W}}\Pr\left\{  \mathcal{T}_{\mathcal{W}\rightarrow\mathcal{B}_{i}%
}\right\}  \boldsymbol{I}_{\mathcal{T}_{\mathcal{W}\rightarrow\mathcal{B}_{i}%
}}\right)  .
\end{align*}

By Definition (\ref{78}),%
\[
\Lambda_{\mathcal{B}_{i}}\left(  \mathcal{B}_{i-1,k}\right)  =\sum
_{\mathcal{I}_{i-1,k}}\Lambda\left(  \overline{\mathcal{B}}_{i-1}\right)
\prod\limits_{\mathcal{W}\in\mathcal{I}_{i-1,k}}\Pi_{\mathcal{B}_{i}}\left(
\mathcal{W}\right)  .
\]

Now use Rule 1 to promote $\cup_{j=0}^{r}\left\{  \mathcal{B}_{i-1,j}%
\backslash\mathcal{B}_{i-1,j}^{\ast}\right\}  $ without cohort, leaving border
$\mathcal{B}_{i}=\cup_{j=0}^{r}\mathcal{B}_{i-1,j}^{\ast}$. From Theorem
\ref{B}, $\Lambda\left(  \overline{\mathcal{B}}_{i-1}\right)  =\Lambda\left(
\mathcal{B}_{i}\right)  $, hence the lemma.
\end{proof}

Lemma \ref{P33} can be carried out more efficiently as%
\begin{align*}
&  \Lambda_{\mathcal{B}_{i}}\left(  \mathcal{B}_{i-1,k}\right) \\
&  =\sum\limits_{\mathcal{B}_{i-1,r}}\Pi_{\mathcal{B}_{i}}^{\mathcal{B}%
_{i-1,r}}...\sum\limits_{\mathcal{B}_{i-1,k+1}}\Pi_{\mathcal{B}_{i}%
}^{\mathcal{B}_{i-1,k+1}}\sum\limits_{\mathcal{B}_{i-1,k-1}}\Pi_{\mathcal{B}%
_{i}}^{\mathcal{B}_{i-1,k-1}}...\sum\limits_{\mathcal{B}_{i-1,0}}%
\Pi_{\mathcal{B}_{i}}^{\mathcal{B}_{i-1,0}}\Lambda\left(  \mathcal{B}%
_{i}\right)  ,
\end{align*}
where $\Pi_{\mathcal{B}_{i}}^{\mathcal{B}_{i-1,j}}=\Pi_{\mathcal{B}_{i}%
}\left(  \mathcal{B}_{i-1,j}\right)  $ for all $0\leq j\leq r$.

Combining Lemma \ref{P2} with the above lemmas yields the BP version of
Theorem \ref{P21}:

\begin{theorem}
\label{P31}If border $\mathcal{B}_{i}$ has received the messages from all
members of $\left\{  \mathcal{H}_{\mathcal{B}_{i}}\backslash\mathcal{B}%
_{i-1,k}\right\}  \cup\mathcal{L}_{\mathcal{B}_{i}}$, then it can send an
upward message $\Lambda_{\mathcal{B}_{i}}\left(  \mathcal{B}_{i-1,k}\right)  $
to its parent $\mathcal{B}_{i-1,k}$ as in Lemma \ref{P32} or \ref{P33}, in
which
\[
\Lambda\left(  \mathcal{B}_{i}\right)  =\prod\limits_{\mathcal{W}%
\in\mathcal{L}_{\mathcal{B}_{i}}}\Lambda_{\mathcal{W}}\left(  \mathcal{B}%
_{i}\right)  .
\]

\end{theorem}

Both the downward and upward passes use the reduced cohort probability tables
$\phi\left(  \mathcal{\cdot}\right)  $, rather than the individual CPTs. So
all $\Phi\left(  \mathcal{\cdot}\right)  $ should be calculated and pre-loaded.

\subsubsection{The Border Evidential Cores}

A border is evidential if one of its component variables is. Let the
\textquotedblleft border evidential core\textquotedblright\ (or
\textquotedblleft border EC\textquotedblright) be the \emph{smallest}
sub-polytree which contains all the evidence variables in its borders. For the
BN $\mathbb{C}$, let us assume $\mathcal{E=}\left\{  B,O,Q\right\}  $. Then
the largest sub-polytree in Figure \ref{F10} containing all the evidence
variables includes path $\left\{  N,M,Q,V\right\}  \rightarrow\left\{
N,M,Q\right\}  \rightarrow\left\{  N,P,Q\right\}  \rightarrow...\rightarrow
\left\{  D,F,O,P\right\}  $ and path $\left\{  B,C\right\}  \rightarrow
\left\{  B,C,G,N,P\right\}  $. However, its border EC\ is smaller, including
only path $\left\{  N,P,Q\right\}  \rightarrow\left\{  N,P\right\}
\rightarrow\left\{  B,C,G,N,P\right\}  \rightarrow\left\{  B,C,G,O,P\right\}
$. For some evidence sets $\mathcal{E}$, the border EC\ is not unique.

A border is \textquotedblleft inside\textquotedblright\ if it is in the border
EC; otherwise it is \textquotedblleft outside;\textquotedblright\ edge
$\mathcal{B}_{i}-\mathcal{B}_{j}$ is \textquotedblleft
inside\textquotedblright\ if both borders are inside; otherwise it is
\textquotedblleft outside;\textquotedblright\ a path is \textquotedblleft
inside\textquotedblright\ if all its component edges are.

\subsubsection{The Boundary Conditions}

As in the polytrees, the calculations of the messages along an inside edge of
a border EC may require the knowledge of the boundary conditions; that is, the
messages to an inside border from its outside neighboring borders. However,
unlike the polytree, our definition of the border ECs allows the outside
borders to be evidential. (For example, borders $\left\{  N,M,Q,V\right\}  $
or $\left\{  B,F,O,P\right\}  $ in Figure \ref{F10} with $\mathcal{E=}\left\{
B,O,Q\right\}  $.) So we need the following BP\ version of Theorem \ref{P10}:

\begin{theorem}
\label{P13}Consider an inside border $\mathcal{B}_{i}$.

\noindent(a) If $\mathcal{B}_{i-1}\in\mathcal{H}_{\mathcal{B}_{i}}$ is
outside, then $\Pi_{\mathcal{B}_{i}}\left(  \mathcal{B}_{i-1}\right)
=\Pr\left\{  \mathcal{B}_{i-1}\right\}  \boldsymbol{I}_{\mathcal{B}_{i-1}}$.

\noindent(b) If $\mathcal{B}_{i+1}\in\mathcal{L}_{\mathcal{B}_{i}}$ is
outside, then $\Lambda_{\mathcal{B}_{i+1}}\left(  \mathcal{B}_{i}\right)
=\boldsymbol{I}_{\mathcal{B}_{i}}$.
\end{theorem}

\begin{proof}
(a) In a border polytree, assume $\mathcal{B}_{i}$ is inside, $\mathcal{B}%
_{i-1}\in\mathcal{H}_{\mathcal{B}_{i}}$ is outside, and the outside parentless
$\mathcal{T}_{\mathcal{B}_{i-1}\rightarrow\mathcal{B}_{i}}$ has an evidence
variable $E\in\widetilde{\mathcal{B}}_{1}$ such that $E\notin\mathcal{B}_{i}$.
By its definition, the border EC must include another evidence border
$\widetilde{\mathcal{B}}_{2}\ni E$. By the running intersection property, the
path connecting $\widetilde{\mathcal{B}}_{1}$ and $\widetilde{\mathcal{B}}%
_{2}$ cannot go through $\mathcal{B}_{i}$. Thus there are two paths connecting
$\widetilde{\mathcal{B}}_{1}$ and $\widetilde{\mathcal{B}}_{2}$, including the
one via $\mathcal{B}_{i}$. Because this is contradictory to the polytree
assumption, we must have $E\in\mathcal{B}_{i}$. By the running intersection
property, $E\in\mathcal{B}_{i-1}$. In other words, $\boldsymbol{I}%
_{\mathcal{T}_{\mathcal{B}_{i-1}\rightarrow\mathcal{B}_{i}}}=\boldsymbol{I}%
_{\mathcal{B}_{i-1}}$. By Definition (\ref{78}),%
\[
\Pi_{\mathcal{B}_{i}}\left(  \mathcal{B}_{i-1}\right)  =\sum_{\mathcal{T}%
_{\mathcal{B}_{i-1}\rightarrow\mathcal{B}_{i}}\backslash\mathcal{B}_{i-1}}%
\Pr\left\{  \mathcal{T}_{\mathcal{B}_{i-1}\rightarrow\mathcal{B}_{i}}\right\}
\boldsymbol{I}_{\mathcal{T}_{\mathcal{B}_{i-1}\rightarrow\mathcal{B}_{i}}}%
=\Pr\left\{  \mathcal{B}_{i-1}\right\}  \boldsymbol{I}_{\mathcal{B}_{i-1}}.
\]

(b) In a border polytree, assume $\mathcal{B}_{i}$ is inside, $\mathcal{B}%
_{i+1}\in\mathcal{L}_{\mathcal{B}_{i}}$ is outside, and the bottom part
$\mathcal{U}_{\mathcal{B}_{i}\rightarrow\mathcal{B}_{i+1}}$ has an evidence
variable $E\in\widetilde{\mathcal{B}}_{3}$ such that $E\notin\mathcal{B}_{i}$.
The border EC must include another evidence border $\widetilde{\mathcal{B}%
}_{4}\ni E$. By the running intersection property, the path connecting
$\widetilde{\mathcal{B}}_{3}$ and $\widetilde{\mathcal{B}}_{4}$ cannot go
through $\mathcal{B}_{i}$. Thus there are two paths connecting $\widetilde
{\mathcal{B}}_{3}$ and $\widetilde{\mathcal{B}}_{4}$, including the one via
$\mathcal{B}_{i}$. As this is a contradiction to the polytree assumption, all
evidence variables in $\mathcal{U}_{\mathcal{B}_{i}\rightarrow\mathcal{B}%
_{i+1}}$ must be in $\mathcal{B}_{i}$. In other words, $\boldsymbol{I}%
_{\mathcal{U}_{\mathcal{B}_{i}\rightarrow\mathcal{B}_{i+1}}}=\boldsymbol{I}%
_{\mathcal{B}_{i}}$. By Definition (\ref{77}),%
\begin{align*}
\Lambda_{\mathcal{B}_{i+1}}\left(  \mathcal{B}_{i}\right)   &  =\sum
_{\mathcal{U}_{\mathcal{B}_{i}\rightarrow\mathcal{B}_{i+1}}}\Pr\left\{
\mathcal{U}_{\mathcal{B}_{i}\rightarrow\mathcal{B}_{i+1}}|\mathcal{B}%
_{i}\right\}  \boldsymbol{I}_{\mathcal{B}_{i}\cup\mathcal{U}_{\mathcal{B}%
_{i}\rightarrow\mathcal{B}_{i+1}}}\\
&  =\sum_{\mathcal{U}_{\mathcal{B}_{i}\rightarrow\mathcal{B}_{i+1}}}%
\Pr\left\{  \mathcal{U}_{\mathcal{B}_{i}\rightarrow\mathcal{B}_{i+1}%
}|\mathcal{B}_{i}\right\}  \boldsymbol{I}_{\mathcal{B}_{i}}=\boldsymbol{I}%
_{\mathcal{B}_{i}}.
\end{align*}

\end{proof}

\subsubsection{The Message Initializations}

Following is the BP\ version of Theorem \ref{P19}:

\begin{theorem}
\label{P46}Consider a border EC.

\noindent(a) If $\mathcal{B}_{R}$ is a root with inside child $\mathcal{B}%
_{C}$, then $\Pi\left(  \mathcal{B}_{R}\right)  =\Pi_{\mathcal{B}_{C}}\left(
\mathcal{B}_{R}\right)  =\Pr\left\{  \mathcal{B}_{R}\right\}  \boldsymbol{I}%
_{\mathcal{B}_{R}}$.

\noindent(b) If $\mathcal{B}_{L}$ is a leaf, then $\Lambda\left(
\mathcal{B}_{L}\right)  =\boldsymbol{I}_{\mathcal{B}_{L}}$ in Lemma \ref{P32}
or \ref{P33}.
\end{theorem}

\begin{proof}
(a) Let $\mathcal{B}_{R}$ be a root of the border EC. As shown in the proof of
Theorem \ref{P13}(a), for all $\mathcal{W}\in\mathcal{H}_{\mathcal{B}_{R}}$,
all evidence variables in $\mathcal{T}_{\mathcal{W}\rightarrow\mathcal{B}_{R}%
}$ are in the inside $\mathcal{B}_{R}$; so are all evidence variables in
$\mathcal{P}_{\mathcal{B}_{R}}=\mathcal{B}_{R}\cup_{\mathcal{W}\in
\mathcal{H}_{\mathcal{B}_{R}}}\mathcal{T}_{\mathcal{W}\rightarrow
\mathcal{B}_{R}}$. Hence $\boldsymbol{I}_{\mathcal{P}_{\mathcal{B}_{R}}%
}=\boldsymbol{I}_{\mathcal{B}_{R}}$. From Definition (\ref{14}),
\[
\Pi\left(  \mathcal{B}_{R}\right)  =\sum_{\mathcal{A}_{_{\mathcal{B}_{R}}}}%
\Pr\left\{  \mathcal{P}_{\mathcal{B}_{R}}\right\}  \boldsymbol{I}%
_{\mathcal{P}_{_{\mathcal{B}_{R}}}}=\sum_{\mathcal{A}_{_{\mathcal{B}_{R}}}}%
\Pr\left\{  \mathcal{A}_{\mathcal{B}_{R}},\mathcal{B}_{R}\right\}
\boldsymbol{I}_{\mathcal{B}_{R}}=\Pr\left\{  \mathcal{B}_{R}\right\}
\boldsymbol{I}_{\mathcal{B}_{R}}.
\]

Also, because $\mathcal{B}_{R}$ has only one inside child $\mathcal{B}_{C}$,
$\Lambda_{\mathcal{W}}\left(  \mathcal{B}_{R}\right)  =\boldsymbol{I}%
_{\mathcal{B}_{R}}$ for all $\mathcal{W}\in\mathcal{L}_{\mathcal{B}_{R}%
}\backslash\mathcal{B}_{C}$. Thus $\Pi\left(  \mathcal{B}_{R}\right)
=\Pi_{\mathcal{B}_{C}}\left(  \mathcal{B}_{R}\right)  $ from Lemma \ref{P7}.

(b) Let $\mathcal{B}_{L}$ be a leaf of the border EC. As shown in the proof of
Theorem \ref{P13}(b), for all $\mathcal{W}\in\mathcal{L}_{\mathcal{B}_{L}}$,
all evidence variables in $\mathcal{U}_{\mathcal{B}_{L}\rightarrow\mathcal{W}%
}$ must be in $\mathcal{B}_{L}$; so are all evidence variables in
$\mathcal{D}_{\mathcal{B}_{L}}=\cup_{\mathcal{W}\in\mathcal{L}_{\mathcal{B}%
_{L}}}\mathcal{U}_{\mathcal{B}_{L}\rightarrow\mathcal{W}}$. Hence
$\boldsymbol{I}_{\mathcal{D}_{\mathcal{B}_{L}}}=\boldsymbol{I}_{\mathcal{B}%
_{L}}$. From Definition (\ref{13}),%
\[
\Lambda\left(  \mathcal{B}_{L}\right)  =\sum_{\mathcal{D}_{\mathcal{B}_{L}}%
}\Pr\left\{  \mathcal{D}_{\mathcal{B}_{L}}|\mathcal{B}_{L}\right\}
\boldsymbol{I}_{\mathcal{B}_{L}\cup\mathcal{D}_{\mathcal{B}_{L}}}%
=\sum_{\mathcal{D}_{\mathcal{B}_{L}}}\Pr\left\{  \mathcal{D}_{\mathcal{B}_{L}%
}|\mathcal{B}_{L}\right\}  \boldsymbol{I}_{\mathcal{B}_{L}}=\boldsymbol{I}%
_{\mathcal{B}_{L}}.
\]

\end{proof}

For the border chain obtained in Section 2, recall that we denote the first
time and the last time an evidence is recruited into $\mathcal{P}$ by $\alpha$
and $\beta$, respectively. We now see that the part of the border chain from
$\alpha$\ to $\beta$ is its border EC, and Theorem \ref{P46} is consistent
with Equations (\ref{47}) and (\ref{32}).

\subsubsection{The Collection Phase}

The inferences in a border polytree are carried out in the same manner as in a
polytree:\ In the collection phase, we start with Theorem \ref{P46} and
propagate the messages inside the border EC to an arbitrarily chosen inside
pivot border.

For the border EC in Figure \ref{F10} with $\mathcal{E=}\left\{
B,O,Q\right\}  $, suppose we pick its leaf $\left\{  B,C,G,O,P\right\}  $ as
the pivot, then the collection phase includes:

\begin{enumerate}
\item By Theorem \ref{P46}(a),
\[
\Pi\left(  N,P,Q\right)  =\Pr\left\{  N,P,Q\right\}  \boldsymbol{I}_{Q}.
\]

\item By Lemma \ref{P34}(b) and Theorem \ref{P30},
\[
\Pi_{\left\{  B,C,G,N,P\right\}  }\left(  N,P\right)  =\Pi\left(  N,P\right)
=\sum_{Q}\Pr\left\{  N,P,Q\right\}  \boldsymbol{I}_{Q}.
\]

\item By Lemma \ref{P35}, with the boundary conditions $\Pi_{\left\{
B,C,G,N,P\right\}  }\left(  B,C\right)  =\Pr\left\{  B,C\right\}
\boldsymbol{I}_{B}$ and $\Pi_{\left\{  B,C,G,N,P\right\}  }\left(  G\right)
=\Pr\left\{  G\right\}  $,
\begin{align*}
&  \Pi\left(  B,C,G,N,P\right) \\
&  =\Pi_{\left\{  B,C,G,N,P\right\}  }\left(  B,C\right)  \Pi_{\left\{
B,C,G,N,P\right\}  }\left(  G\right)  \Pi_{\left\{  B,C,G,N,P\right\}
}\left(  N,P\right) \\
&  =\Pr\left\{  B,C\right\}  \boldsymbol{I}_{B}\Pr\left\{  G\right\}  \sum
_{Q}\Pr\left\{  N,P,Q\right\}  \boldsymbol{I}_{Q}.
\end{align*}

\item By Lemma \ref{P34}(b), at the pivot border,%
\begin{align}
&  \Pi\left(  B,C,G,O,P\right)  =\sum_{N}\Pr\left\{  O|C,G,N\right\}
\boldsymbol{I}_{O}\Pi\left(  B,C,G,N,P\right) \nonumber\\
&  =\sum_{N}\Pr\left\{  O|C,G,N\right\}  \boldsymbol{I}_{O}\Pr\left\{
B,C\right\}  \boldsymbol{I}_{B}\Pr\left\{  G\right\}  \sum_{Q}\Pr\left\{
N,P,Q\right\}  \boldsymbol{I}_{Q}. \label{68}%
\end{align}

\end{enumerate}

With the initial condition $\Lambda\left(  B,C,G,O,P\right)  =\boldsymbol{I}%
_{B}\boldsymbol{I}_{O}$, the pivot border is now informed. By Corollary
\ref{FF}, $\Pr\left\{  P,\left[  B,O,Q\right]  \right\}  $ can be calculated
as%
\begin{align}
&  \sum_{\left\{  B,C,G,O\right\}  }\Pi\left(  B,C,G,O,P\right)
\Lambda\left(  B,C,G,O,P\right) \nonumber\\
&  =\sum_{\left\{  B,C,G,O\right\}  }\sum_{N}\Pr\left\{  O|C,G,N\right\}
\boldsymbol{I}_{O}\Pr\left\{  B,C\right\}  \boldsymbol{I}_{B}\Pr\left\{
G\right\}  \sum_{Q}\Pr\left\{  N,P,Q\right\}  \boldsymbol{I}_{Q}. \label{69}%
\end{align}
It is easy to verify that the RHS is indeed $\Pr\left\{  P,\left[
B,O,Q\right]  \right\}  $.

On the other hand, suppose we pick\ root $\left\{  N,P,Q\right\}  $ as the
pivot border, then the collection phase includes:

\begin{enumerate}
\item By Theorem \ref{P46}(b),%
\[
\Lambda\left(  B,C,G,O,P\right)  =\boldsymbol{I}_{B}\boldsymbol{I}_{O}.
\]

\item By Lemma \ref{P32}(b),%
\begin{align*}
\Lambda\left(  B,C,G,N,P\right)   &  =\sum_{O}\Pr\left\{  O|C,G,N\right\}
\boldsymbol{I}_{O}\Lambda\left(  B,C,G,O,P\right) \\
&  =\sum_{O}\Pr\left\{  O|C,G,N\right\}  \boldsymbol{I}_{O}\boldsymbol{I}_{B}.
\end{align*}

\item By Theorem \ref{P31} and Lemma \ref{P33}, with the boundary conditions
$\Pi_{\left\{  B,C,G,N,P\right\}  }\left(  B,C\right)  =\Pr\left\{
B,C\right\}  \boldsymbol{I}_{B}$ and $\Pi_{\left\{  B,C,G,N,P\right\}
}\left(  G\right)  =\Pr\left\{  G\right\}  $,%
\begin{align*}
&  \Lambda\left(  N,P\right)  =\Lambda_{\left\{  B,C,G,N,P\right\}  }\left(
N,P\right) \\
&  =\sum_{\left\{  B,C,G\right\}  }\Pi_{\left\{  B,C,G,N,P\right\}  }\left(
B,C\right)  \Pi_{\left\{  B,C,G,N,P\right\}  }\left(  G\right)  \Lambda\left(
B,C,G,N,P\right) \\
&  =\sum_{\left\{  B,C,G\right\}  }\Pr\left\{  B,C\right\}  \boldsymbol{I}%
_{B}\Pr\left\{  G\right\}  \sum_{O}\Pr\left\{  O|C,G,N\right\}  \boldsymbol{I}%
_{O}.
\end{align*}

\item By Theorem \ref{P31} and Lemma \ref{P32}(b), at the pivot border,%
\begin{equation}
\Lambda\left(  N,P,Q\right)  =\Lambda\left(  N,P\right)  . \label{59}%
\end{equation}

\end{enumerate}

With the initial condition $\Pi\left(  N,P,Q\right)  =\Pr\left\{
N,P,Q\right\}  \boldsymbol{I}_{Q}$, the pivot border is now informed. By
Corollary \ref{FF}, $\Pr\left\{  P,\left[  B,O,Q\right]  \right\}  $ can be
found by%
\begin{align*}
&  \sum_{\left\{  N,Q\right\}  }\Pi\left(  N,P,Q\right)  \Lambda\left(
N,P,Q\right) \\
&  =\sum_{\left\{  N,Q\right\}  }\Pr\left\{  N,P,Q\right\}  \boldsymbol{I}%
_{Q}\sum_{\left\{  B,C,G\right\}  }\Pr\left\{  B,C\right\}  \boldsymbol{I}%
_{B}\Pr\left\{  G\right\}  \sum_{O}\Pr\left\{  O|C,G,N\right\}  \boldsymbol{I}%
_{O},
\end{align*}
which is the same as Equation (\ref{69}).

\subsubsection{The Distribution Phase}

In the collection phase, the messages converge to the pivot border; so it
helps to know all the evidence variables and construct the border EC before
the message propagations. Introducing a new evidence may require
re-calculating some messages already sent to the pivot border. On the other
hand, as discussed in \S 5.6, the distribution phase starts when the pivot
border becomes the single member of the informed set $\mathcal{J}$.
Introducing a new query border $\mathcal{Q}$, which includes a query variable,
only requires calculating the messages from the corresponding gate in
$\mathcal{J}$ to $\mathcal{Q}$, making $\mathcal{Q}$ informed and thus allow
$\Pr\left\{  \mathcal{Q},\left[  \mathcal{E}\backslash\mathcal{Q}\right]
\right\}  \boldsymbol{I}_{\mathcal{Q}}$ to be calculated as $\Pi\left(
\mathcal{Q}\right)  \Lambda\left(  \mathcal{Q}\right)  $ by Theorem \ref{FB}.
There is no message re-calculation when the query variables are considered one-at-a-time.

For the BN $\mathbb{C}$ with $\mathcal{E=}\left\{  B,O,Q\right\}  $, assume we
wish to obtain the posterior marginal of the outside variable $M$. With
$\Lambda\left(  N,P,Q\right)  $ obtained in Equation (\ref{59}), we can use
Lemma \ref{P32} to send an upward message to border $\left\{  N,M,Q\right\}  $
as%
\[
\Lambda\left(  N,M,Q\right)  =\sum_{P}\Pr\left\{  P|N,M,Q\right\}
\boldsymbol{I}_{Q}\Lambda\left(  N,P,Q\right)
\]

Finally, with $\Pi\left(  N,M,Q\right)  =\Pr\left\{  N,M,Q\right\}
\boldsymbol{I}_{Q}$,%
\[
\Pr\left\{  M,\left[  B,O,Q\right]  \right\}  =\sum_{\left\{  N,Q\right\}
}\Pi\left(  N,M,Q\right)  \Lambda\left(  N,M,Q\right)  =\sum_{\left\{
N,Q\right\}  }\Pr\left\{  N,M,Q\right\}  \boldsymbol{I}_{Q}\Lambda\left(
N,M,Q\right)  .
\]

On the other hand, if we wish to obtain the posterior marginal of the outside
variable $F$, then we can start with $\Pi\left(  B,C,G,O,P\right)  $
calculated in Equation (\ref{68}):

\begin{enumerate}
\item Using Lemma \ref{P34},%
\[
\Pi\left(  B,C,O,P\right)  =\sum_{G}\Pi\left(  B,C,G,O,P\right)  .
\]

\item Using Lemma \ref{P34}, $\Pi\left(  B,F,O,P\right)  =\sum_{C}\Pr\left\{
F|B,C\right\}  \boldsymbol{I}_{B}\Pi\left(  B,C,O,P\right)  .$
\end{enumerate}

Finally, with $\Lambda\left(  B,F,O,P\right)  =\boldsymbol{I}_{B}%
\boldsymbol{I}_{O}$,%
\[
\Pr\left\{  F,\left[  B,O,Q\right]  \right\}  =\sum_{\left\{  B,O,P\right\}
}\Pi\left(  B,F,O,P\right)  \boldsymbol{I}_{B}\boldsymbol{I}_{O}.
\]

If all non-evidence variables are the query variables, the messages may be
sent \textquotedblleft asynchronously,\textquotedblright\ without a goal; that
is, once a border becomes informed, it may send the messages to all of its
neighbors that have not received a message from it. (See D\'{\i}ez \& Mira,
1994.) Especially in this case,

\begin{enumerate}
\item If the number of children of $\mathcal{B}_{i}$ is large, then it may be
advantageous to calculate $\Lambda\left(  \mathcal{B}_{i}\right)
=\prod\limits_{\mathcal{W}\in\mathcal{L}_{\mathcal{B}_{i}}}\Lambda
_{\mathcal{W}}\left(  \mathcal{B}_{i}\right)  $ first, then, for all
$\mathcal{B}_{i+1,j}\in\mathcal{L}_{\mathcal{B}_{i}}$ in Theorem \ref{P30}, we
calculate $\Lambda\left(  \mathcal{B}_{i}\right)  /\Lambda_{\mathcal{B}%
_{i+1,j}}\left(  \mathcal{B}_{i}\right)  $ instead of $\prod
\limits_{\mathcal{W}\in\mathcal{L}_{\mathcal{B}_{i}}\backslash\mathcal{B}%
_{i+1,j}}\Lambda_{\mathcal{W}}\left(  \mathcal{B}_{i}\right)  $;

\item If the number of parents of $\mathcal{B}_{i}$ is large, then it may be
advantageous to calculate $\Pi\left(  \mathcal{H}_{\mathcal{B}_{i}}\right)
=\prod\limits_{\mathcal{W}\in\mathcal{H}_{\mathcal{B}_{i}}}\Pi_{\mathcal{B}%
_{i}}\left(  \mathcal{W}\right)  $ first, then, for all $\mathcal{B}%
_{i-1,k}\in\mathcal{H}_{\mathcal{B}_{i}}$ in Lemma \ref{P33}, we calculate
$\Pi\left(  \mathcal{H}_{\mathcal{B}_{i}}\right)  /\Pi_{\mathcal{B}_{i}%
}\left(  \mathcal{B}_{i-1,k}\right)  $ instead of $\prod\limits_{\mathcal{W}%
\in\mathcal{H}_{\mathcal{B}_{i}}\backslash\mathcal{B}_{i-1,k}}\Pi
_{\mathcal{B}_{i}}\left(  \mathcal{W}\right)  $.
\end{enumerate}

\section{Discussions}

\noindent1.\qquad Our algorithm can handle what are known as the
\textquotedblleft soft evidences.\textquotedblright\ An evidence variable $E$
is soft if $\operatorname*{Va}^{e}\left(  E\right)  $, its set of observed
values, may have more than one members. In other words, the evidence indicator
column $\boldsymbol{I}_{E}$ may have more than one non-zero values. This is
less restrictive than the \textquotedblleft hard evidence\textquotedblright%
\ assumption normally found in other inference algorithms,\ which requires
$\operatorname*{Va}^{e}\left(  E\right)  $ to have only one value. (See
Langevin \& Valtorta, 2008.)

\noindent2.\qquad All junction-tree based inferences share a worst-case
complexity, which is exponential with respect to the largest clique size of
the underlying undirected graph. According to Wu and Butz (2005),
\textquotedblleft Lauritzen and Spiegelhalter... were concerned with the size
of the clique in the junction tree (transformed from the DAG of a BN), and
they realized that their method would not be computational \textit{feasible}
if a large clique is present in the junction tree. The Hugin architecture has
the same concern as the Lauritzen-Spiegelhalter architecture, namely, the size
of the clique in a junction tree. The Shafer-Shenoy architecture... used
hypertree and Markov tree (junction tree) to describe the architecture. In
[9], it was repeatedly emphasized that the efficiency and feasibility of their
architecture depends on the size of the clique in a junction
tree.\textquotedblright\ ([9] referred to Shafer, 1996.) Wu and Butz (2005)
then showed that \textquotedblleft the presence of a node with a large number
of parents can occur in both singly connected and multiply connected BNs.
Therefore, in both singly and multiply connected BNs, the computation for
exact inference will be exponential in the worst case.\textquotedblright

In our algorithm, let us similarly assume that the largest border size
in a border polytree is not too large. This imposes some limitations on the
sizes of all cohort probability tables $\Phi\left(  \mathcal{\cdot}\right)  $,
which is dependent to the numbers of parents and children, and the number of
possible values of each variable $V\in\mathcal{V}$.

\noindent3.\qquad The collection phase collects information about the
evidences in the entire border polytree. It is intuitive that this process
should be reduced to within the sub-polytree border EC: We only need to pass
messages within it, toward its inside pivot border, starting from its
evidential roots and leaves as in Theorem \ref{P46}. In other words, in the
collection phase, the BP is \textquotedblleft pruned\textquotedblright\ to its
border EC. As illustrated above with the BN $\mathbb{C}$ having $\mathcal{E=}%
\left\{  B,O,Q\right\}  $, the collection phase to the pivot border $\left\{
N,P,Q\right\}  $ only requires four message propagations within its border EC.

This message passing reduction is possible in our algorithm because we know
the boundary conditions: Having a \emph{directed} border polytree, we can use
Theorem \ref{P13} because we can determine whether a border's outside neighbor
is its parent or child. Also, it is important that all prior marginals
$\Pr\left\{  \mathcal{B}_{i}\right\}  $ are pre-calculated (only once) and
\emph{pre-loaded} for this theorem. For example, to use Equation (\ref{68}),
besides the CPTs $\Pr\left\{  O|C,G,N\right\}  $ and $\Pr\left\{  G\right\}
$, we need the prior marginals $\Pr\left\{  N,P,Q\right\}  $ and $\Pr\left\{
B,C\right\}  $.

In many applications, the difference between the prior and the posterior
marginals of a variable is more telling than the posterior itself, so
pre-calculating the prior marginals should be done anyhow. Off-line, we
calculate $\Pr\left\{  \mathcal{B}_{i}\right\}  $ as $\Pi\left(
\mathcal{B}_{i}\right)  $ without evidence, in a topological order of the
macro-nodes and of the borders inside each macro-node, using Lemma \ref{P34}
(with the complete cohort probability tables $\Phi\left(  \cdot\right)  $) and
Lemma \ref{P35} (with $\Pi_{\mathcal{B}_{i}}\left(  \mathcal{B}_{i-1,j}%
\right)  =\Pi\left(  \mathcal{B}_{i-1,j}\right)  $ by Theorem \ref{P30}).

In the collection phase, each border in a EC sends only one message toward the
pivot border. Thus the time complexity of the collection phase is linear with
respect to the number of borders within the sub-polytree border EC. With one
additional evidence, this number increases by the number of borders between it
and the pivot border. If $N$ is the sole evidence in the BN\ $\mathbb{C}$,
then the collection phase is not needed, as the pivot border $\left\{
N,P,Q\right\}  $ is automatically informed with $\Lambda\left(  N,P,Q\right)
=\boldsymbol{I}_{N}$ and $\Pi\left(  N,P,Q\right)  =\Pr\left\{  N,P,Q\right\}
\boldsymbol{I}_{N}$, where $\Pr\left\{  N,P,Q\right\}  $ is pre-loaded. Thus
the time complexity in the collection phase is linear with respect to the
number of evidence variables.

\noindent4.\qquad Similarly, in the distribution phase, we pass messages from
the growing informed set $\mathcal{J}$ to a query border $\mathcal{Q}$. As
discussed above, the propagation starts from the informed gate connecting
$\mathcal{J}$ with $\mathcal{Q}$; hence there is no need to re-visit any node
inside $\mathcal{J}$. In other words, in the distribution phase, the BP is
\textquotedblleft pruned\textquotedblright\ to its \textquotedblleft query
core,\textquotedblright\ which is the smallest sub-polytree that contains all
the query borders. The computational complexity in the distribution phase is
linear with respect to the number of borders inside the query core, which in
turn is linear with the number of query variables. Again, the pre-loaded prior
marginals $\Pr\left\{  \mathcal{B}_{i}\right\}  $ and the directed border
polytrees are essential for the boundary conditions.

Consider the case where all non-evidence variables are the query variables. As
far as we know, all junction-tree based inference architectures (including the
LAZY propagation algorithm, Madsen \& Jensen, 1999) require two passes through
the entire network, one in each phase. Our algorithm requires one pass through
the smaller border EC in the collection phase, and one pass through the entire
network in the distribution phase. On the other hand, consider the case in
which the single evidence variable and the single query variable are in the
same border; our algorithm requires no message propagation in both phases.

\noindent5.\qquad In summary, the novel features in this paper are:

\begin{enumerate}
\item[i.] The parentless polytree method (\S 6.1) to partition a BN\ into a
macro-node polytree, by opening the loops in an otherwise growing parentless polytree.

\item[ii.] The border algorithm to construct a directed chain from a BN
(\S 2), or from a macro-node (\S 6.2).
\end{enumerate}

Combining the above two algorithms, we can convert any Bayesian network\ into
a border polytree. The border algorithm then provides the means to propagate
the downward and upward messages in a border polytree, allowing us to
calculate its posterior marginal probabilities (\S 6.3).

Also, the message propagations in the distribution phase is carried out one
query border at a time, within the query core sub-polytree only (\S 5.5).

With the above novel features, \emph{the time complexity of our inferences in
a Bayesian network is linear with respect to the number of its evidence and
query variables, regardless of the number of borders in its corresponding
border polytree, or the number of its variables.}

\section*{References}

\begin{enumerate}
\item Chang, K., and Fung, R. (1989) \textquotedblleft Node Aggregation for
Distributed Inference in Bayesian Networks\textquotedblright\ In
\textit{Proceedings of the 11th International Joint Conference on Artificial
Intelligence}, Detroit, Michigan, 265--270.

\item Cooper, G. F. (1990) \textquotedblleft The computational complexity of
probabilistic inference using Bayesian belief networks\textquotedblright%
\ \textit{Artificial Intelligence,} 42, 393--405.

\item Darwiche, A. (2003) \textquotedblleft A differential approach to
inference in Bayesian networks\textquotedblright\ \textit{Journal of the ACM},
50, 280-305.

\item Dechter, R. (1999) \textquotedblleft Bucket elimination: A unifying
framework for reasoning\textquotedblright\ \textit{Artificial Intelligence},
113, 41--85.

\item D\'{\i}ez, F. (1996) \textquotedblleft Local conditioning in Bayesian
networks\textquotedblright\ \textit{Artificial Intelligence}, 87(1996).

\item D\'{\i}ez, F. and Mira, J. (1994) \textquotedblleft Distributed
inference in Bayesian networks\textquotedblright\ \textit{Cybernetics and
Systems}, 25, 39--61.

\item Guo, H. and Hsu, W. (2002) \textquotedblleft A survey of algorithms for
real-time Bayesian network inference\textquotedblright\ In the joint
AAAI-02/KDD-02/UAI-02 workshop on Real-Time Decision Support and Diagnosis Systems.

\item Kim, J. H. and Pearl J. (1983) \textquotedblleft A computational model
for combined causal and diagnostic reasoning in inference
engines\textquotedblright\ \textit{Proceedings of the 8th International Joint
Conference on Artificial Intelligence}, Karlsruhe, West Germany, 190--193.

\item Koller, D. and Friedman, N. (2009) \textit{Probabilistic Graphical
Models: Principles and Techniques,} Massachusetts: MIT Press.

\item Langevin, S. and Valtorta, M. (2008) \textquotedblleft Performance
evaluation of algorithms for soft evidential update in Bayesian networks:
First results\textquotedblright\ \textit{Scalable Uncertainty Management}, 284--297.

\item Lauritzen, S. L. and Spiegelhalter, D. J. (1988)\ \textquotedblleft
Local computations with probabilities on graphical structures and their
applications to expert systems\textquotedblright\ \textit{Journal of the Royal
Statistical Society}, Series B, 50, 157-224.

\item Lepar, V. and Shenoy, P. (1999). \textquotedblleft A comparison of
Lauritzen-Spiegelhalter, Hugin, and Shenoy-Shafer architectures for computing
marginals of probability distributions\textquotedblright\ in G. F. Cooper \&
S. Moral (eds.), \textit{Uncertainty in Artificial Intelligence}, 14, 328-337,
Morgan Kaufmann, San Francisco, CA.

\item Madsen, A. L. and Jensen, F. V. (1999) \textquotedblleft LAZY
propagation: A junction tree inference algorithm based on lazy
evaluation\textquotedblright\ \textit{Artificial Intelligence}, 113, 203-245.

\item Ng, K. and Levitt, T. S. (1994) \textquotedblleft Incremental Dynamic
Construction of Layered Polytree Networks\textquotedblright%
\ \textit{Proceedings of the Tenth Conference on Uncertainty in Artificial
Intelligence, 440-446.}

\item Pearl, J. (1986a) \textquotedblleft A constraint-propagation approach to
probabilistic reasoning\textquotedblright\ In L. N. Kanal and J. F. Lemmer
(Eds), \textit{Proceedings of the 2nd Conference on Uncertainty in Artificial
Intelligence}, Amsterdam, NorthHolland, 357-369.

\item Pearl, J. (1986b) \textquotedblleft Fusion, propagation and structuring
in belief networks\textquotedblright\ \textit{Artificial Intelligence}, 29, 241-288.

\item Pearl, J. (1987) \textquotedblleft Evidential reasoning using stochastic
simulation of causal models\textquotedblright\ \textit{Artificial
Intelligence}, 32, 245--257.

\item Pearl, J. (1988) \textit{Probabilistic Reasoning in Intelligent Systems:
Networks of Plausible Inference}, Morgan Kaufmann.

\item Peot, M. A., and Shachter, R. D. (1991) \textquotedblleft Fusion and
propagation with multiple observations in belief networks.\textquotedblright%
\ \textit{Artificial Intelligence}, 48, 299--318.

\item Shachter, R. D. (1990) \textquotedblleft Evidence absorption and
propagation through evidence reversals\textquotedblright\ In M. Henrion, R. D.
Shachter, J. F. Lemmer, and L. N. Kanal (Eds.), \textit{Uncertainty in
Artificial Intelligence,} 5, 173-190.

\item Shachter, R. D., D'Ambrosio, B., Del Favero, B. D. (1990)
\textquotedblleft Symbolic probabilistic inference in belief
networks\textquotedblright\ \textit{Proceedings of the 8th National Conference
on Artificial Intelligence}, MIT Press, Boston, 126-131.

\item Shafer, G. (1996) \textit{Probabilistic Expert Systems}, Society for
Industrial and Applied Mathematics.

\item Wu, D. and Butz C. \textquotedblleft On the complexity of probabilistic
inference in singly connected Bayesian networks\textquotedblright%
\ \textit{Rough Sets, Fuzzy Sets, Data Mining, and Granular Computing}, 581-590.

\item Zhang, N. L. and Poole, D. (1996) \textquotedblleft Exploiting causal
independence in Bayesian network inference\textquotedblright\ \textit{Journal
of Artificial Intelligence Research}, 5, 301--328.
\end{enumerate}

\end{document}